\def\templateArxiv{arxiv}
\def\templateIEEE{IEEE}

\def\templateType{arxiv}

\ifx\templateType\templateArxiv
    \documentclass[lettersize, twocolumn, journal, 11pt]{IEEEtran}

\usepackage[table]{xcolor}

\usepackage[english]{babel}
\usepackage[utf8]{inputenc}

\usepackage{amssymb}
\usepackage{multicol}
\usepackage{afterpage}

\usepackage{wrapfig}

\usepackage[table]{xcolor}
\usepackage{multirow}
\usepackage{makecell}
\usepackage{array}
\usepackage{rotating}
\usepackage{booktabs}
\usepackage[skip=3pt]{caption}
\usepackage{stfloats} 

\usepackage{cellspace} 
\usepackage{subcaption}
\usepackage{setspace}
\usepackage{placeins}

\usepackage{paralist}

\usepackage{amssymb}
\usepackage{amsmath}
\usepackage{bm}

\usepackage{expl3}

\usepackage{dirtytalk}
\usepackage{gensymb}

\usepackage{hyperref}
\hypersetup{
    colorlinks,
    linkcolor={red!50!black},
    citecolor={blue!60!black},
    urlcolor={blue!70!black}
}

\usepackage{soul}
\usepackage[final]{pdfcomment}

 \usepackage{microtype}

\usepackage{tikz}

\newcommand\copyrighttext{\footnotesize 
This work has been submitted for publication.
  DOI: \href{https://ieeexplore.ieee.org/document/11162505}{10.1109/ACCESS.2025.3609630}}
\newcommand\copyrightnotice{\begin{tikzpicture}[remember picture,overlay]
\node[anchor=south,yshift=10pt] at (current page.south) {\fbox{\parbox{\dimexpr\textwidth-\fboxsep-\fboxrule\relax}{\copyrighttext}}};
\end{tikzpicture}}

 \else \ifx\templateType\templateIEEE

\let\texyear\year

\documentclass[Path=./ieeeTemplates/]{ieeeaccess}

\let\ieeeaccessyear\year
\let\year\texyear

\usepackage[table]{xcolor}

\usepackage[english]{babel}
\usepackage[utf8]{inputenc}

\usepackage{amssymb}
\usepackage{multicol}
\usepackage{afterpage}

\usepackage{wrapfig}

\usepackage[table]{xcolor}
\usepackage{multirow}
\usepackage{makecell}
\usepackage{array}
\usepackage{rotating}
\usepackage{booktabs}
\usepackage[skip=3pt]{caption}
\usepackage{stfloats} 

\usepackage{cellspace} 
\usepackage{subcaption}
\usepackage{setspace}
\usepackage{placeins}

\usepackage{paralist}

\usepackage{amssymb}
\usepackage{amsmath}
\usepackage{bm}

\usepackage{expl3}

\usepackage{dirtytalk}
\usepackage{gensymb}

\usepackage{hyperref}
\hypersetup{
    colorlinks,
    linkcolor={red!50!black},
    citecolor={blue!60!black},
    urlcolor={blue!70!black}
}

\usepackage{soul}
\usepackage[final]{pdfcomment}

\let\year\ieeeaccessyear
\definecolor{accessblue}{cmyk}{1, 0.3, 0, 0.2}
\definecolor{greycolor}{cmyk}{0,0,0,.8}

\graphicspath{{./ieeeTemplates}{./pics/authors}} \fi\fi

\graphicspath{{.}{./images}{./images/dist}{./images/graph_examples/dist}{./images/lit}{./images/noise}{./images/exDist}}

\begin{document}

\ifx\templateType\templateArxiv
    
\title{Bayes Error Rate Estimation in Difficult Situations
}

\author{\IEEEauthorblockN{Lesley Wheat\IEEEauthorrefmark{1}\IEEEauthorrefmark{2}\IEEEauthorrefmark{3}, Martin v. Mohrenschildt\IEEEauthorrefmark{1} and Saeid Habibi\IEEEauthorrefmark{2}}

\IEEEauthorblockA{\IEEEauthorrefmark{1}Department of Computing and Software, McMaster University, Hamilton, Canada}

\IEEEauthorblockA{\IEEEauthorrefmark{2}Center for Mechatronics and Hybrid Technologies, McMaster University, Hamilton, Canada}

\IEEEauthorblockA{\IEEEauthorrefmark{3}Email: wheatd@mcmaster.ca}

}

\twocolumn[
\begin{@twocolumnfalse}
\maketitle
\end{@twocolumnfalse}
\copyrightnotice

\begin{abstract}
~{The Bayes Error Rate (BER) is the fundamental limit on the achievable generalizable classification accuracy of any machine learning model due to inherent uncertainty within the data.
BER estimators offer insight into the difficulty of any classification problem and set expectations for optimal classification performance.
In order to be useful, the estimators must also be accurate with a limited number of samples on multivariate problems with unknown class distributions.
To determine which estimators meet the minimum requirements for \say{usefulness}, an in-depth examination of their accuracy is conducted using Monte Carlo simulations with synthetic data in order to obtain their confidence bounds for binary classification.
To examine the usability of the estimators for real-world applications, new  
non-linear multi-modal test scenarios are introduced.
In each scenario, 2500 Monte Carlo simulations per scenario are run over a wide range of BER values.
In a comparison of k-Nearest Neighbor (kNN), Generalized Henze-Penrose (GHP) divergence and Kernel Density Estimation (KDE) techniques, results show that kNN is overwhelmingly the more accurate non-parametric estimator.
In order to reach the target of an under 5\% range for the 95\% confidence bounds, the minimum number of required samples per class is 1000.
As more features are added, more samples are needed, so that 2500 samples per class are required at only 4 features.
Other estimators do become more accurate than kNN as more features are added, but continuously fail to meet the target range.
 }
\end{abstract}


\bigskip
]

 \else \ifx\templateType\templateIEEE

\history{Date of submission July 11, 2024.}
\doi{10.1109/ACCESS.2024.DOI}

\title{Testing Bayes Error Rate Estimators in Difficult Situations using Monte Carlo Simulations}

\author{\uppercase{Lesley Wheat}\authorrefmark{1}\authorrefmark{2},
\uppercase{Martin v. Mohrenschildt} \authorrefmark{1}, \uppercase{ and Saeid Habibi} \authorrefmark{2} \IEEEmembership{Member, IEEE}}
\address[1]{Department of Computing and Software, McMaster University, Hamilton, ON L8S4L7 Canada}
\address[2]{Center for Mechatronics and Hybrid Technologies (CMHT), Department of Mechanical Engineering, McMaster University}
\tfootnote{
This work was supported in part by the Canada Research Chairs (CRC) Program under Project CRC-2020-0127 and in part by the Natural Sciences and Engineering Research Council of Canada (NSERC) Ford-Mitacs Alliance under Project ALLRP-590906-23.
 }

\markboth
{Wheat \headeretal: Testing Bayes Error Rate Estimators in Difficult Situations using Monte Carlo Simulations}
{Wheat \headeretal: Testing Bayes Error Rate Estimators in Difficult Situations using Monte Carlo Simulations}

\corresp{Corresponding author: Lesley Wheat (e-mail: wheatd@mcmaster.ca).}

\begin{abstract}
The Bayes Error Rate (BER) is the fundamental limit on the achievable generalizable classification accuracy of any machine learning model due to inherent uncertainty within the data.
BER estimators offer insight into the difficulty of any classification problem and set expectations for optimal classification performance.
In order to be useful, the estimators must also be accurate with a limited number of samples on multivariate problems with unknown class distributions.
To determine which estimators meet the minimum requirements for \say{usefulness}, an in-depth examination of their accuracy is conducted using Monte Carlo simulations with synthetic data in order to obtain their confidence bounds for binary classification.
To examine the usability of the estimators for real-world applications, new  
non-linear multi-modal test scenarios are introduced.
In each scenario, 2500 Monte Carlo simulations per scenario are run over a wide range of BER values.
In a comparison of k-Nearest Neighbor (kNN), Generalized Henze-Penrose (GHP) divergence and Kernel Density Estimation (KDE) techniques, results show that kNN is overwhelmingly the more accurate non-parametric estimator.
In order to reach the target of an under 5\% range for the 95\% confidence bounds, the minimum number of required samples per class is 1000.
As more features are added, more samples are needed, so that 2500 samples per class are required at only 4 features.
Other estimators do become more accurate than kNN as more features are added, but continuously fail to meet the target range.
 \end{abstract}


\titlepgskip=-15pt

\maketitle \else \ifx\templateType\templateElsevier
    \begin{frontmatter}

\title{Bayes Error Rate Estimation in Difficult Situations}

\affiliation[inst1]{organization={Department of Computing and Software, McMaster University},addressline={1280 Main St W}, 
            city={Hamilton},
            postcode={L8S4L7}, 
            state={ON},
            country={Canada}}

\affiliation[inst2]{organization={Center for Mechatronics and Hybrid Technologies (CMHT),
Department of Mechanical Engineering, 
McMaster University},addressline={1280 Main St W}, 
            city={Hamilton},
            postcode={L8S4L7}, 
            state={ON},
            country={Canada}}

\author[inst1,inst2]{Lesley Wheat}
\author[inst1]{Martin v. Mohrenschildt}
\author[inst2]{Saeid Habibi}

\begin{abstract}

\noindent
Abstract

\end{abstract}

\begin{highlights}
\item A
\item B
\item C
\end{highlights}


\end{frontmatter} \fi\fi\fi

\section{Introduction}

While perfect classification is ideal, for many real-life problems, it is often not possible.
While errors may arise due to the design of the classifier itself, the data itself has fundamental limits due to overlapping class densities, a property of the dataset.
Such overlap may be caused by mislabeled samples, measurement noise, or other errors within the dataset that can not necessarily be corrected by the classifier.
The result is the classification stage starts off with an imperfect dataset, which itself puts a limit on the best accuracy that can be achieved on the problem.

As the ideal classifier is not known for the large majority of current open problems, this can lead to an impossible search to design a classifier with a minimum accuracy higher than what is possible on a particular problem.
In such cases, it can be incredibly difficult to recognize when the classifier design is not the issue, and there are fundamental errors within the data that must be addressed.
By estimating the performance limitations of the data, realistic expectations can be set for classification accuracy.

\begin{figure}[h]
\centering
\includegraphics[width=0.98\linewidth]{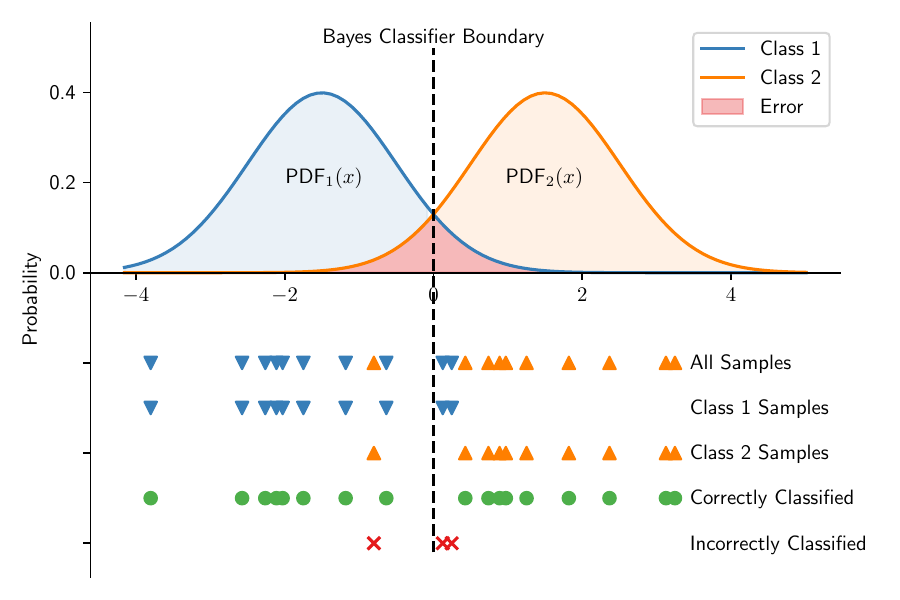}
\caption{
Example of the Bayes Classifier and BER from probability distribution functions (PDFs) on uni-variate Gaussian distributions from two classes. }
\label{fig:ber_ex}
\end{figure}

The Bayes Error Rate (BER) is the fundamental performance limit for statistical pattern recognition problems and provides the lower bound on the generalizable error rate that can be achieved on a specific problem \cite{tumer_estimating_1996} \pdfmarkupcomment[color=yellow]{(Figure {\ref{fig:ber_ex}})}{(RVR2.7) Figure has been changed.}.
The BER is the classification rate of the Bayes Classifier \cite{sekeh_learning_2020}, which can only be used when the class probability distributions are known, which is rare in real-world problems.
However, there are ways to estimate the BER without being given information of the underlying class distributions, through the use of non-parametric methods.

In contrast to previous research, the focus will be placed on the \say{usefulness} of BER estimators for open \textit{engineering} problems. Emphasis will be placed on the following criteria:

\noindent
\begin{minipage}{0.95\linewidth}
\smallskip
\begin{itemize}
    \item Overall estimator accuracy and confidence bounds.
    \item Required number of samples.
    \item Maximum number of features.
    \item Distribution/problem robustness.
\end{itemize}
\smallskip
\end{minipage}

\noindent To this end, new testing scenarios will be introduced based on open problems to get a better picture of the expected estimator performance under different use cases, described in Section \ref{sec:dist}.
The code used to create these scenarios is made publicly available so that they can be reused for future research.

On these scenarios, a selection of popular estimation techniques are applied including: k-Nearest Neighbor (kNN) \cite{berisha_empirically_2016}, Generalized Henze-Penrose (GHP) divergence \cite{sekeh_learning_2020} and Gaussian Kernel Density Estimation (GKDE) \cite{renggli_evaluating_2021}.
Additionally, a new estimation technique, Comparative Localized Kernel Density Estimation (CLAKDE) is introduced to address some of the shortcomings of GKDE, \pdfmarkupcomment[color=yellow]{which can serve as a stepping stone for further research}{(RVR2.2) Clarified.}.
The 95\% confidence intervals for the best estimators are reported in Section \ref{sec:testResults}.

The results show significant changes in accuracy depending on the test scenario and the number of features.
Ultimately, the number of required samples in many scenarios is much higher than preferred.
In Section \ref{sec:dis}, several problems with the current methods are examined to explain why the estimators largely fail to meet the goals set out in Section \ref{sec:estimators} and areas for improvement are highlighted.

\noindent
\begin{minipage}{0.95\linewidth}
\smallskip
The contributions of this paper are:
\begin{itemize}
    \item \pdfmarkupcomment[color=yellow]{To compare the performance of different estimators over a selection of synthesized datasets.}{Edited.}
    \item \pdfmarkupcomment[color=yellow]{Provide error bounds and guidelines for minimum sample sizes based on the numbers of features.}{Edited.}
    \item To introduce a new BER estimator based on kernel density estimation \pdfmarkupcomment[color=yellow]{to improve estimation with larger numbers of features}{(RVR2.2) Added clarification.}.
\end{itemize}
\smallskip
\end{minipage} 
\section{Background}

BER estimation has a long history \cite{fukunaga_bayes_1987}, but is still an ongoing area of research which has recently had renewed interest.
The availability of data and computational resources has increased the feasibility of BER estimation and sparked an interest on applying estimators to real-world problems \cite{theisen_evaluating_2021, renggli_evaluating_2021, sekeh_learning_2020}.
However, there are still ongoing challenges with these estimation techniques, particularly with respect to measuring high-dimensional datasets with unknown distributions, non-parametric problems.

To address these challenges, a wide variety of techniques have been applied in the past, including those focused on but not limited to: graphs \cite{berisha_empirically_2016}, local structure \cite{fukunaga_bayes_1987}, global structure \cite{renggli_evaluating_2021},  and ensemble methods \cite{tumer_estimating_1996}.
Many methods have a basis in non-parametric two-sample tests, due to the similar nature of the problem, while others take an optimal classification approach.
However, it is expected that all estimators suffer from performance deterioration as the number of features in the dataset increases, but there are only a limited number of works which have done a comparison on numbers of features \cite{noshad_learning_2019, berisha_empirically_2016}.
The theory behind the individual estimators being evaluated will be covered in Section \ref{sec:estimators}.

\renewcommand{\arraystretch}{1.2}
\begin{table}[t]
\centering
\begin{tabular}{
lll
} 
\hline

Source & \makecell[l]{Distribution \\ Type} & Parameters \\ \hline

\multirow{18}{*}{\cite{chen_evaluating_2023}} & 
\makecell[l]{
Multi-class \\
Gaussian \\
Distribution
}& 
\makecell[l]{
Classes: 4 \\
Samples: [400, 400000]\\
Spacing: 1.2\\
Variance: 0.3 }
\\ \cline{3-3}

& &
\makecell[l]{
Classes: 4 \\
Samples: 4000 \\
Spacing: [0, 2.4]\\
Variance: 0.3 } \\ \cline{2-3}

& \makecell[l]{
Binary \\
Gaussian \\
Distribution
}& 
\makecell[l]{
Variance: 0.3 \\
Samples: [400, 400000]\\
Spacing: 1.2\\
Variance: 0.3} \\ \cline{3-3}

& &
\makecell[l]{
Variance: 0.3 \\
Spacing: [0, 2.4] } \\ \cline{2-3}

& Binary L1 &
Samples: 1500 \\ \cline{2-3}

& Binary L2 &
Samples: 1500 \\ \cline{2-3}

& Binary Moon &
Samples: 1500 \\ \cline{2-3}

& Binary Anis &
Samples: 1500 \\ \cline{2-3}

& Binary Squa &
Samples: 9771 \\ \cline{2-3}

& Binary Tria &
Samples: 9771 \\ \cline{2-3}

& Binary Star &
Samples: 6955 \\ \cline{2-3}

& Binary Chan &
Samples: 5621 \\ \hline

\end{tabular}

\caption{Synthetic Dataset Details from Existing Literature. Table 1 of 3.}
\label{table:litRev_p1}
\ifx\templateType\templateArxiv \vspace{-10mm} \fi
\end{table}

\begin{table}[ht]
\centering
\begin{tabular}{ lll
} 
\hline

Source & \makecell[l]{Distribution\\Type} & Parameters \\ \hline

\multirow{24}{1cm}{\cite{noshad_learning_2019}} & 
\makecell[l]{Binary \\ Gaussian \\ Distributions} &
\makecell[l]{
Samples: $[<100, 3000]$\\
Spacing: 5\\
Features: 10
}
\\ \cline{2-3}

& \makecell[l]{Multi-class \\ Concentric \\ Distributions} &
\makecell[l]{
Classes: 4\\
Features: 20\\
Samples: $[<100, 5000]$
}
\\ \cline{2-3}

& \multirow{2}{1.5cm}{\makecell[l]{Multi-class \\ Rayleigh \\ Distributions}} &
\makecell[l]{
Classes: 3\\
Samples: $[<100, 1000]$ \\
Features: 30 \\
$\alpha$: 0.7, 1.0, 1.3
}
\\ \cline{3-3}

& &
\makecell[l]{
Classes: 3\\
Samples: $[<100, >3000]$\\
Features: 10\\
$\alpha$: 2, 4, 6
}
\\ \cline{2-3}

& \multirow{2}{1.5cm}{Binary Isotropic Normal Distributions} &
\makecell[l]{
Samples: $[<100, 1600]$\\
Spacing: 2 \\
Features: 4
}
\\ \cline{3-3}

& &
\makecell[l]{
Samples: $[<100, 1600]$\\
Spacing: 2 \\
Features: 100
}
\\ \cline{3-3}

& & 
\makecell[l]{
Samples: $[<100, >3000]$\\
Spacing: 5 \\
Features: 10}
\\ \cline{3-3}

& &
\makecell[l]{
Samples: $[<100, >3000]$\\
Spacing: 5 \\
Features: 100
}
\\ \cline{2-3}

& \makecell[l]{Multi-class \\Isotropic \\Normal \\Distributions} &
\makecell[l]{
Classes: 4\\
Samples: $[<100, >3000]$\\
Spacing: 5\\
Features: 100
}
\\ \cline{2-3}

& \makecell[l]{Multi-class\\Beta\\Distributions} &
\makecell[l]{
Classes: 3\\
Features: 50\\
Samples: $[<100, 1600]$\\
}
\\ \hline

 \end{tabular}

\caption{Synthetic Dataset Details from Existing Literature. Table 2 of 3.}
\label{table:litRev_p2}
\ifx\templateType\templateArxiv \vspace{-4mm} \fi

\end{table}

\begin{table}[ht]
\centering
\begin{tabular}{
lll
} 
\hline

Source & \makecell[l]{Distribution \\ Type} & Parameters \\ \hline

\multirow{8}{*}{\cite{sekeh_learning_2020}} & 
\multirow{8}{2cm}{Multi-class Gaussian Distribution}&
\makecell[l]{
Features: 2\\
Classes: 4\\
Variance: 0.3\\
Spacing: [0.00, 2.00] 
}
\\ \cline{3-3}

& &
\makecell[l]{
Features: 2\\
Classes: [2, 20] \\
Spacing: 1 
}
\\ \cline{3-3}

& &\makecell[l]{
Features: [2, 5, 10, 20, 30]\\
Classes: 4\\
Samples: $[<100, 10000]$\\
Variance: 0.1\\
Spacing: 0.7\\
Monte Carlo simulations: 100
}
\\ \hline

\multirow{7}{1cm}{\cite{jeong_demystifying_2024}} &
 \multirow{7}{2cm}{Multi-class Gaussian Distribution} &
\makecell[l]{
Classes: 4\\
Features: 2 \\
Samples: 2000 \\
Spacing: [0.25, 2.00]
} \\ \cline{3-3}

& &
 \makecell[l]{
Classes: 4\\
Features: 2 \\
Samples: [40, 4000]\\
 Number of simulations\\per sample size: 50
 } \\ \cline{3-3}

& &
 \makecell[l]{
Classes: 4\\
Features: 2 \\
Samples: [40, 120000] 
 }\\ \hline

\cite{theisen_evaluating_2021} &
\makecell[l]{Binary\\Gaussian\\Distributions} &
\makecell[l]{
Features: 784\\
$\tau$: $[0.25, >2.5]$\\
Spacing: Random\\
Monte Carlo runs: 100
}
\\ \hline
\end{tabular}

\caption{Synthetic Dataset Details from Existing Literature. Table 3 of 3.}
\label{table:litRev_p3}
\ifx\templateType\templateArxiv \vspace{-4mm} \fi

\end{table}
 
\subsection{Testing Scenarios} \label{sec:lit_dists}

In existing literature, a wide variety of distributions have been previously used to test the performance of BER estimators.
Tables \ref{table:litRev_p1}, \ref{table:litRev_p2} and \ref{table:litRev_p3} showcase the variety of the testing scenarios.
Further information on the specific Gaussian distributions used can be found in Table \ref{table:litRev_ex} (Section \ref{sec:AppA}).
From these tables, one issue that can be observed with many of these scenarios is the lack of testing across different target BER values (eg. changes in variance or spacing).
This is an issue because the results of \cite{theisen_evaluating_2021, renggli_evaluating_2021, sekeh_learning_2020} all show that estimator accuracy can change when the distribution parameters are altered.

Overall, Gaussian distributions (binary and multi-class) are the most popular, even when non-parametric estimators are being evaluated.
There are advantages to this, most importantly the algorithm available for calculating the Bayes Error from the distribution parameters \cite{fukunaga_error_1990}.
Using Monte Carlo approximation to determine the BER \cite{sekeh_learning_2020} has become viable in more recent years due to advances in computation speed, and therefore, new literature tends to have more variety in the types of testing scenarios utilized.
However, this inconsistency in test cases makes comparison even more difficult, a natural tradeoff to the increasing number of test cases.

\begin{figure}[!ht]
\centering
  \begin{minipage}{0.98\linewidth}
\subcaptionbox{Example One}
      {\includegraphics[width=0.48\linewidth]{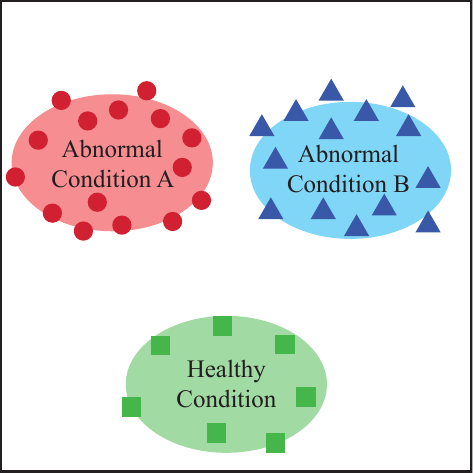}}
      \hfill
    \subcaptionbox{Example Two}
      {\includegraphics[width=0.48\linewidth]{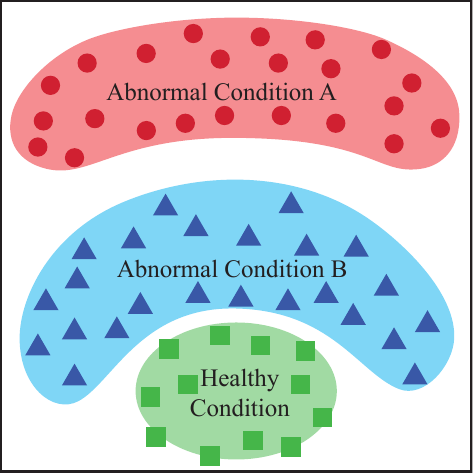}}
      \hfill
    \caption{Examples cases of structures seen in research data.}
    \label{fig:ex_dist_fdd}
  \end{minipage}\quad
  \ifx\templateType\templateArxiv \vspace{-4mm} \fi
\end{figure}

Still, none of these previous works appear to address some of the distributions seen in current engineering problems.
In this case, the area of specific concern is Fault Detection and Diagnosis (FDD), which is a field that attempts to detect abnormal conditions within mechanical systems based on sensor data \cite{wheat_impact_2024}.
Two simplified examples of distributions that have been seen in real data are displayed in Figure \ref{fig:ex_dist_fdd}.

Figure \ref{fig:ex_dist_fdd}.A shows a single cluster, a simple case that may be Gaussian (or close enough to one that it may be treated as such). 
Figure \ref{fig:ex_dist_fdd}.B shows a healthy condition in a single cluster, but abnormal conditions that spread out and begin to curve.
This may occur when there are changes in an underlying variable or through the use of analysis methods.
For example, semi-circular distributions can be caused by changes in energy within vibration data when examined in the frequency domain (when projected into two dimensions).

To think about it a different way, the circular distribution is simply a Gaussian distribution, stretched out and twisted in space.
Effectively, the points distributed along the curve can be seen as situated on a manifold, a topological space that locally resembles Euclidean space.
Therefore, this manifold could be {\say{unfolded}}, transforming the distribution back into a Gaussian, and simplifying the problem.

Unfortunately, it can be very difficult to identify these structures inside datasets.
Figure {\ref{fig:ex_dist_fdd}}.B is a simple example in two dimensions, but it is important to remember that these manifolds can exist in higher dimensions, where they can be much more difficult to recognize and unfold.
Without the knowledge of what these structures are, or even that they exist within the data, there is no chance of being able to transform the data into a more manageable state.

While attempting to find this transformation is an approach that can be taken, which has it's own set of difficulties, and it is considered to be out of scope.
Rather, a {\say{black box}} solution, that works for both Figures {\ref{fig:ex_dist_fdd}}.A and {\ref{fig:ex_dist_fdd}}.B, would be a more preferable approach.

However, the behavior and accuracy of existing BER estimators is unknown in this case, as it is not a typical testing scenario.
The most relevant existing test case to such a scenario is the concentric distribution from \cite{noshad_learning_2019} (Table \ref{table:litRev_p2}), however, it is only evaluated against one BER scenario.
Given that no existing literature was found to address this particular issue, new testing distributions needed to be created (described in Section \ref{sec:dist}).

\FloatBarrier

\begin{figure}[b]
\centering
  \begin{minipage}{0.98\linewidth}
\centering
    \ifx\templateType\templateArxiv
        \subcaptionbox{Noisy Labels}
      {\includegraphics[width=0.48\linewidth]{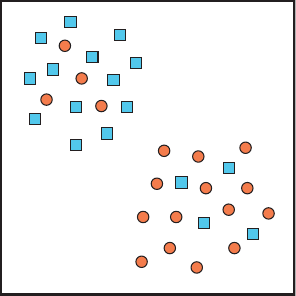}}
    \else
      \subcaptionbox{Noisy Labels}
      {\includegraphics[width=0.45\linewidth]{Artboard_1_noise.pdf}}
    \fi
      \hfill
    \ifx\templateType\templateArxiv
    \subcaptionbox{Noisy Edges}
      {\includegraphics[width=0.48\linewidth]{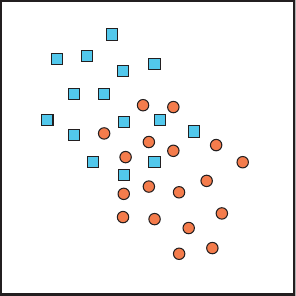}}
    \else
    \subcaptionbox{Noisy Edges}
      {\includegraphics[width=0.45\linewidth]{Artboard_2_noise.pdf}}
    \fi
      \hfill
    \caption{Examples of noisy edges versus noisy labels with two clusters of different classes.}
    \label{fig:noise_ex}
  \end{minipage}\quad
\end{figure} \section{Simulation Data} \label{sec:dist}

In order to test the estimators, synthetic data generated from known probability densities is used.
This allows for an arbitrary number of samples to be generated as needed, the use of the Bayes Classifier, and the ability to establish a ground truth for the BER.
Based on the issues described in Section \ref{sec:lit_dists}, new class distributions were created to test the performance of the estimators.
Additional distributions were added to observe how the estimators may behave in three different scenarios to check for any differences in performance, since the underlying class distribution is considered to be unknown.
The criteria for selecting the additional class distributions was based on the following:

\noindent
\begin{minipage}{0.95\linewidth}
\ifx\templateType\templateArxiv\smallskip\fi
\smallskip
\begin{itemize}
    \item The probability distribution function (PDF) for each class distribution is known.
    \item The BER is determined by parameters that control the level of overlap of the two distributions.
    \item The distribution can be expanded to a large number of features.
    \item Each additional dimension must contribute to the overall separability of the problem. For example, new features can not simply be random values.
    \item Should be a case in which a non-parametric estimator would be required.
\end{itemize}
\smallskip
\end{minipage}

\ifx\templateType\templateArxiv
\afterpage{\clearpage} \ifx\templateType\templateArxiv
\begin{figure*}[p]
\else
\begin{figure*}[htb]
\fi
  \begin{minipage}{\linewidth}
\subcaptionbox{GvG}
      {\includegraphics[width=0.24\linewidth]{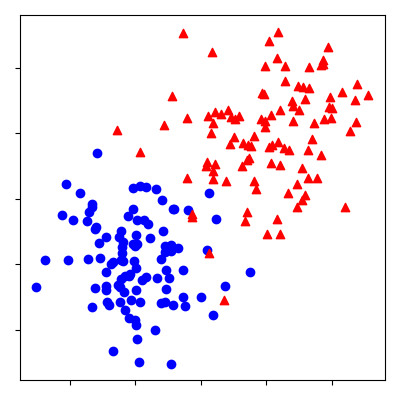}}
      \hfill
    \subcaptionbox{TvT}
      {\includegraphics[width=0.24\linewidth]{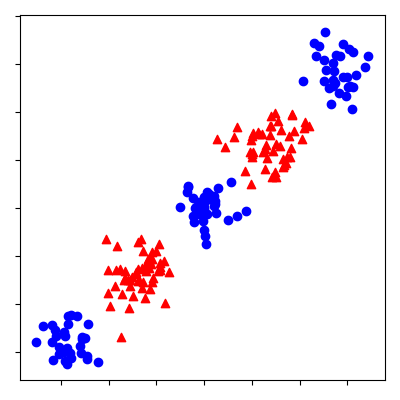}}
      \hfill
    \subcaptionbox{TvS}
      {\includegraphics[width=0.24\linewidth]{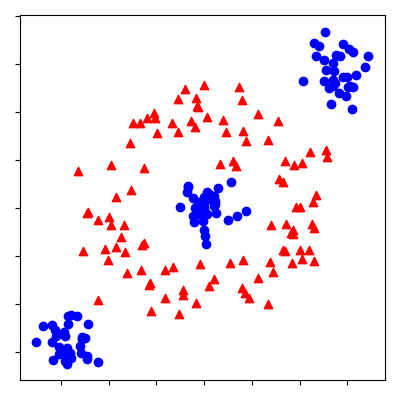}}
      \hfill
    \subcaptionbox{SvS}
      {\includegraphics[width=0.24\linewidth]{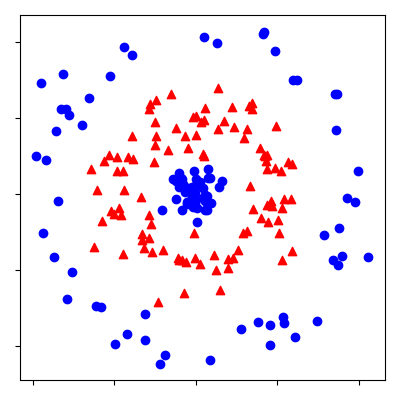}}
      \hfill
    \caption{Examples of distribution types in 2D.}
    \label{fig:dist_2d}
  \end{minipage}\quad
  \bigskip
  \begin{minipage}{\linewidth}
\subcaptionbox{GvG}
      {\includegraphics[width=0.24\linewidth]{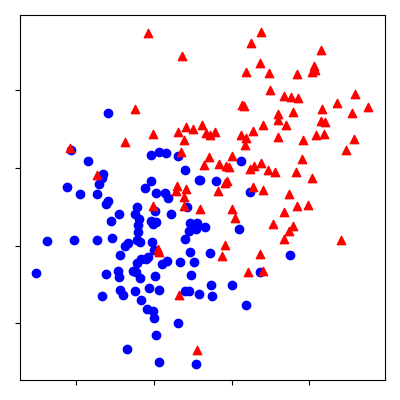}}
      \hfill
    \subcaptionbox{TvT}
      {\includegraphics[width=0.24\linewidth]{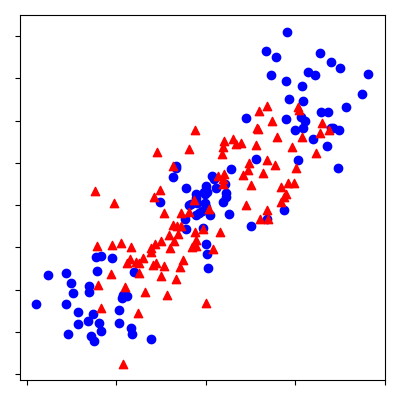}}
      \hfill
    \subcaptionbox{TvS}
      {\includegraphics[width=0.24\linewidth]{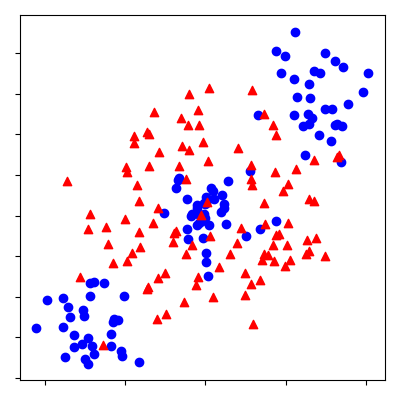}}
      \hfill
    \subcaptionbox{SvS}
      {\includegraphics[width=0.24\linewidth]{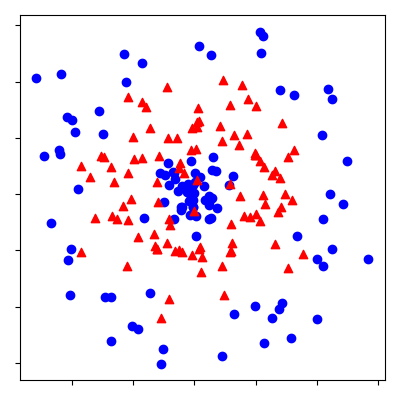}}
      \hfill
    \caption{Examples of distribution types in 2D with increased distribution overlap.}
    \label{fig:dist_2d_noisy}
  \end{minipage}\quad
  \bigskip
  \begin{minipage}{\linewidth}
\subcaptionbox{GvG}
      {\includegraphics[width=0.24\linewidth]{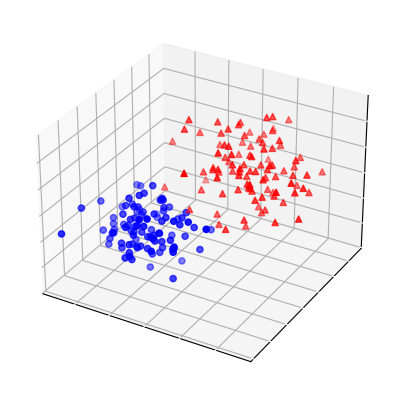}}
      \hfill
    \subcaptionbox{TvT}
      {\includegraphics[width=0.24\linewidth]{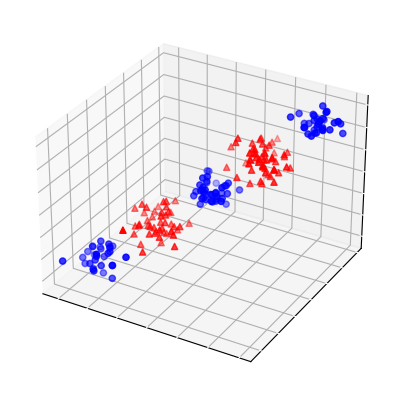}}
      \hfill
    \subcaptionbox{TvS}
      {\includegraphics[width=0.24\linewidth]{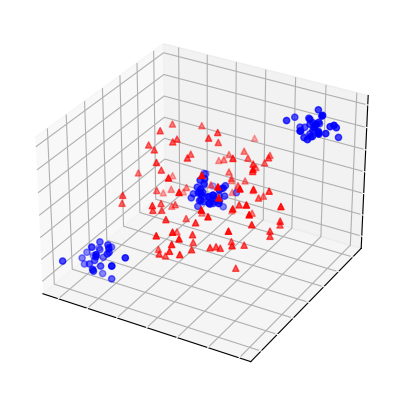}}
      \hfill
    \subcaptionbox{SvS}
      {\includegraphics[width=0.24\linewidth]{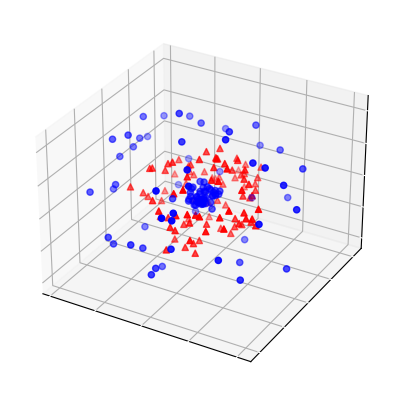}}
      \hfill
    \caption{Examples of distribution types in 3D.}
    \label{fig:dist_3d}
  \end{minipage}\quad
\end{figure*}

 \else \ifx\templateType\templateArxiv
\begin{figure*}[p]
\else
\begin{figure*}[htb]
\fi
  \begin{minipage}{\linewidth}
\subcaptionbox{GvG}
      {\includegraphics[width=0.24\linewidth]{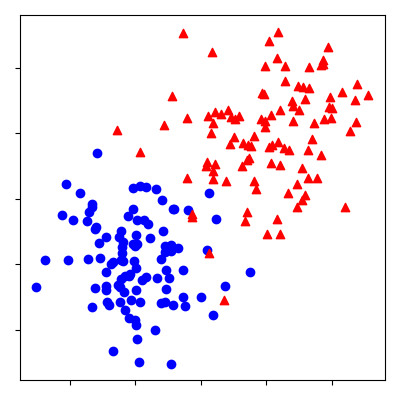}}
      \hfill
    \subcaptionbox{TvT}
      {\includegraphics[width=0.24\linewidth]{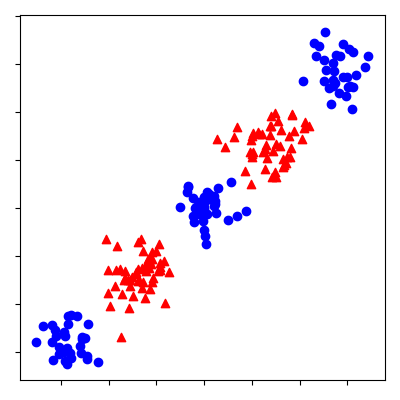}}
      \hfill
    \subcaptionbox{TvS}
      {\includegraphics[width=0.24\linewidth]{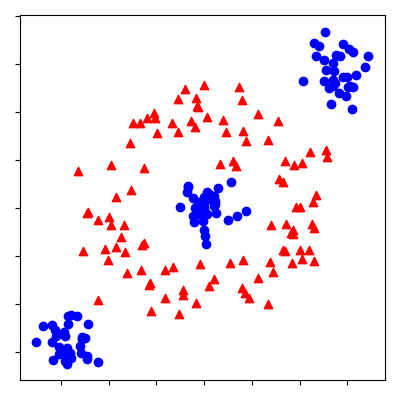}}
      \hfill
    \subcaptionbox{SvS}
      {\includegraphics[width=0.24\linewidth]{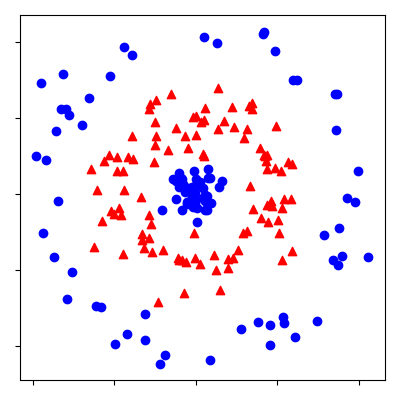}}
      \hfill
    \caption{Examples of distribution types in 2D.}
    \label{fig:dist_2d}
  \end{minipage}\quad
  \bigskip
  \begin{minipage}{\linewidth}
\subcaptionbox{GvG}
      {\includegraphics[width=0.24\linewidth]{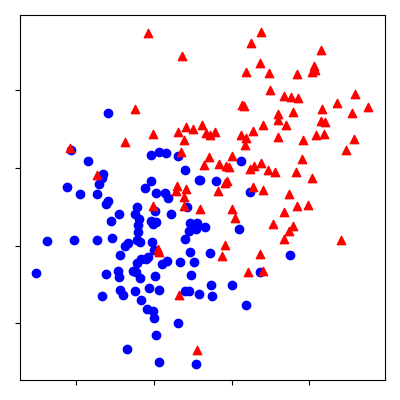}}
      \hfill
    \subcaptionbox{TvT}
      {\includegraphics[width=0.24\linewidth]{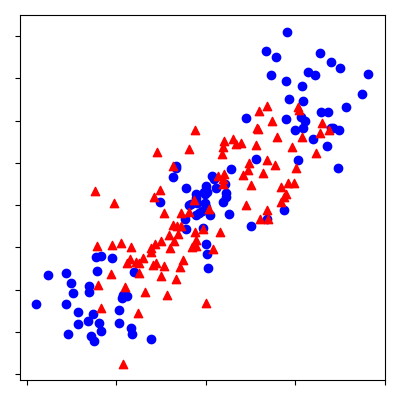}}
      \hfill
    \subcaptionbox{TvS}
      {\includegraphics[width=0.24\linewidth]{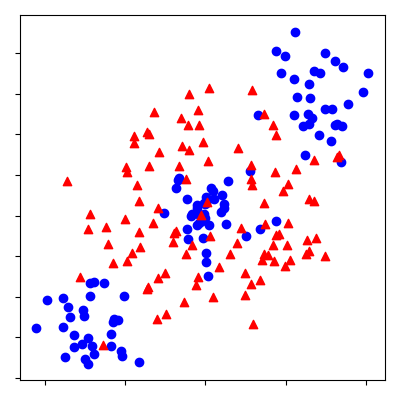}}
      \hfill
    \subcaptionbox{SvS}
      {\includegraphics[width=0.24\linewidth]{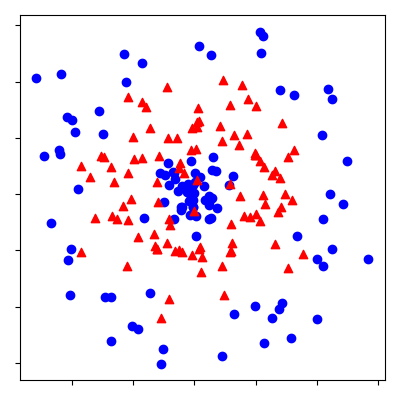}}
      \hfill
    \caption{Examples of distribution types in 2D with increased distribution overlap.}
    \label{fig:dist_2d_noisy}
  \end{minipage}\quad
  \bigskip
  \begin{minipage}{\linewidth}
\subcaptionbox{GvG}
      {\includegraphics[width=0.24\linewidth]{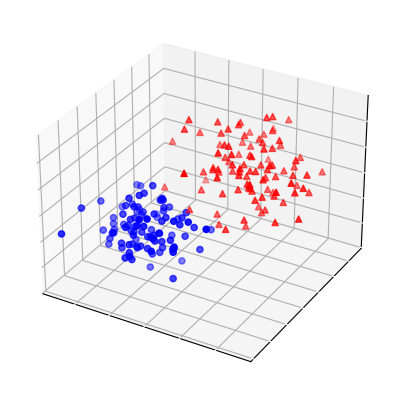}}
      \hfill
    \subcaptionbox{TvT}
      {\includegraphics[width=0.24\linewidth]{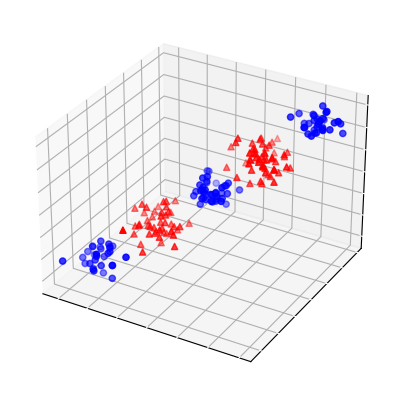}}
      \hfill
    \subcaptionbox{TvS}
      {\includegraphics[width=0.24\linewidth]{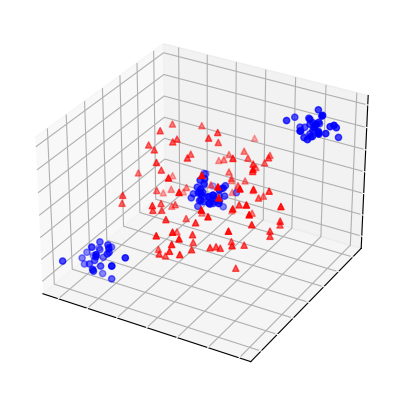}}
      \hfill
    \subcaptionbox{SvS}
      {\includegraphics[width=0.24\linewidth]{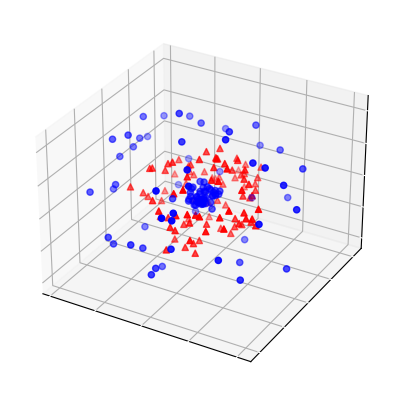}}
      \hfill
    \caption{Examples of distribution types in 3D.}
    \label{fig:dist_3d}
  \end{minipage}\quad
\end{figure*}

 \fi

Additionally, the structure of \say{noisy edges} over \say{noisy labels} \cite{renggli_evaluating_2021} (see Figure \ref{fig:noise_ex}) is preferred, due to the similarity to the distributions of interest (see Figure \ref{fig:ex_dist_fdd}).
In our applications, we are less concerned with label noise and more concerned with mixture of classes at the edges, which can be caused by measurement noise and other confounding factors.
While the label noise, in theory, produces the same change in BER, the overall structure of the data is different and this may impact the behavior of the estimators.
Therefore, the parameters of the distributions (offset, noise, etc.) must be configurable in order to alter the BER of the scenario.

With that in mind, four combinations of distributions of two classes were used for testing: simple clusters (GvG), multiple clusters (TvT), a sphere with clusters (TvS), and multiple spheres (SvS).
Examples of these distributions are shown in Figures \ref{fig:dist_2d}, \ref{fig:dist_2d_noisy} and \ref{fig:dist_3d}.

GvG is the most basic test case (see Figure \ref{fig:dist_2d}.A), where each class is drawn from a multivariate Gaussian distribution.
This is a common case used in literature (see Section \ref{sec:lit_dists}) and is included for comparison purposes.
The class PDFs are given by:
\begin{align}
p_A(x) = N(x | \mu=\vec{0}, \textstyle \sum = \sigma^2_A I) \\
p_B(x) = N(x | \mu= \mu_B \vec{1}, \textstyle \sum = \sigma^2_B I)
\end{align}

\noindent With an offset in every dimension and $\sigma^2_A \neq \sigma^2_B$, every feature helps contribute to the overall class separability.

To build on the GvG distribution, TvT (Figure \ref{fig:dist_2d}.B) is used to introduce a problem that cannot be solved with a linear classifier.
To this end, each class is turned into a Gaussian mixture and inter-spaced to increase complexity.
Class A (blue) is composed of three Gaussian distributions and Class B (red) is made up of two Gaussian distributions (located halfway between the central and outer distributions of class A).
Each sample has an equal probability of being drawn from all Gaussian distributions within the class.

\subsection{Spherical Distributions}

In order to create a suitable test for the curved distribution described in Section \ref{sec:lit_dists}, a Gaussian mixture is constructed with the centers distributed evenly over the surface of a sphere.
With enough points over the surface, and an equal chance of drawing from each Gaussian, an approximation of the target distribution can be created.
This distribution is similar in shape to the multiclass concentric distributions used in \cite{noshad_learning_2019}, but the implementation is different.

Using a Gaussian mixture approximation has multiple implementation benefits.
No matter the placement of the centers, the overall distribution is still interdependently and identically distributed (i.i.d.).
The radius and covariance of the Gaussians can be altered, to change the overlap of the class distributions.
This method can be expanded to more features, based on the construction of hyperspheres in higher dimensional space, so that the overall structure stays the same.

In TvS (Figure \ref{fig:dist_2d}.C), the class A (blue) is the same as class A from TvT, a three Gaussian mixture, and class B (red) is an approximate spherical distribution.
The radius used for class B is half the spacing used for class A.

SvS (Figure \ref{fig:dist_2d}.D) has a similar appearance to a \say{bull's eye} and represents the most difficult, or complex, of the four cases.
Class A (blue) is composed of a single Gaussian at the center and a spherical distribution, with a even chance of samples being drawn from the central Gaussian or the sphere.
Class B (red) is a spherical distribution with half the radius of class A.

\section{Estimators} \label{sec:estimators}

To clarify the requirements, the objective is to find estimators that meet the following criteria:

\begin{enumerate}[{R}.1]
\item Consistent performance over different types of problems. \label{req:robustToDist}
\item Do not require user-defined parameters. \label{req:param}
\item Show robust performance across increasing numbers of features. \label{req:features}
\item Require low numbers of samples for accurate estimation (ideal number of samples per class: $<1000$, allowable: $<2000$). \label{req:samples}
\item The 95\% confidence range of the estimator should be under 5\%. \label{req:errorBounds}
\item Produce deterministic results (always output the same result given the same data). \label{req:deterministic}
\end{enumerate}

Based on the factors listed above, a number of estimators were selected for testing, which will be covered in further detail in the following sections:

\begin{minipage}{0.95\linewidth}
\smallskip
\begin{itemize}
    \item kNN: k-Nearest Neighbour Classifier
    \item GHP: Generalized Henze-Penrose divergence
    \item GKDE: Gaussian Kernel Density Estimation
    \item CLAKDE (new): Comparative Localized Adaptive Kernel Density Estimation
    \item GC (new): GHP and CLAKDE combination
\end{itemize}
\smallskip
\end{minipage}

\pdfmarkupcomment[color=yellow]{
It is important to note that the objectives in this case do differ from previous work in several respects.
Consequently, the requirements criteria is tailored to the problem and the list of estimators is not exhaustive.
Also, the final evaluation is performed on confidence bounds rather than simply overall error, as the final goal is usability, and simply outperforming other estimators is not good enough.
}{Edited for clarity.}

\pdfmarkupcomment[color=yellow]{Another major difference, due to the focus on smaller datasets (R.{\ref{req:samples}}), is that computation complexity is not as much of a concern.
This is because at smaller sample sizes, overall computation time is heavily determined by time per sample, rather than the rate of growth per additional sample.
In this case, some increase in computation time is also seen as an allowable tradeoff if it ultimately results in a better estimation.
}{(RVR2.4) Added further discussion on computation complexity.}

\subsection{Divergence Based Estimators}

Divergence measures are another strategy to estimate the BER, as these measures are based on the dissimilarity between two distribution.
There are several methods that fall into this category, for the sake of simplicity, only Generalized Henze-Penrose (GHP) divergence \cite{sekeh_learning_2020} will be included, as the state-of-the-art in this area.

There are several contributing factors to this choice.
On multi-class Gaussian distributions, GHP has been shown to be a tighter bound than Jensen-Shannon (JS) divergence or pairwise (PW).
It may be more resilient to the curse of dimensionality, but has not been previously tested in more complex scenarios.
It has the inherent advantage of requiring no user-selected parameters.
Popularity is another factor, as this estimator has already been used on several benchmark datasets \cite{renggli_evaluating_2021}.

Although GHP is for bounding the BER, the goal here is to estimate the true BER. 
With this in mind, three values will be used as separate estimators with accuracy measured in relation to the true BER:

\noindent
\begin{minipage}{0.95\linewidth}
\smallskip
\begin{itemize}
    \item $\mbox{GHP}_H$: GHP upper bound.
    \item $\mbox{GHP}_L$: GHP lower bound.
    \item $\mbox{GHP}_M$: Midpoint of $\mbox{GHP}_H$ and $\mbox{GHP}_L$.
\end{itemize}
\smallskip
\end{minipage}

When the bounds are tight, it is expected that all these values will be close to the true BER.
It is to be expected that $\mbox{GHP}_H$ overestimates the BER while $\mbox{GHP}_L$ underestimates it, and this is expected to be observed in the bounds on estimator error (that the error is not symmetrical).

A graph-based approach, GHP involves the construction of a minimum spanning tree (MST) over the dataset \cite{sekeh_learning_2020}.
The same implementation in \cite{renggli_evaluating_2021} is used without modification.
Given how the datasets are collected (random number generation), it is likely that there are no equal (or extremely closely) weighted edges, thus the MST will be unique.
Thus, the method is also expected to be deterministic, generating the same result on the same dataset every time.

\subsection{k Nearest Neighbor Based Estimators} \label{sec:knn}

This section takes a k Nearest Neighbours (kNN) approach  to BER estimation and covers the following estimators:

\noindent
\begin{minipage}{0.9\linewidth}
\ifx\templateType\templateArxiv\smallskip\fi
\smallskip
\begin{itemize}
    \item $\mbox{kNN}_H$: Lowest upper bound.
    \item $\mbox{kNN}_L$: Lower bound of $\mbox{kNN}_H$.
    \item $\mbox{kNN}_M$: Midpoint of $\mbox{kNN}_H$ and $\mbox{kNN}_L$.
\end{itemize}
\smallskip
\end{minipage}

kNN-based techniques have long been used in BER estimation \cite{fukunaga_bayes_1987, devijver_multiclass_1985}.
Here, a leave-one-out (LOO) approach is used for train-test splitting.
This implementation makes use of the implementation by \cite{renggli_evaluating_2021}, for kNN-LOO upper and lower bounds, for consistency.
In this case, only the two-class case is used.
The kNN upper bound ($\mbox{kNN}_H(k)$) for any $k$ is the kNN-LOO error.
The lower bound ($\mbox{kNN}_L(k)$) \cite{renggli_evaluating_2021}, based on \cite{cover_nearest_1967} and \cite{devroye_asymptotic_1981}, is given by:

\begin{equation} \label{eq:knnL}
\mbox{kNN}_L(k) = 
\begin{cases}
\mbox{kNN}_H(k)/(1+\sqrt{1/k}), & k > 2, \\
\mbox{kNN}_H(2)/2, & k = 2, \\
\frac{1- \sqrt{\max (0, 1-2 \mbox{kNN}_H(1)} ) }{2}, & k = 1.
\end{cases}
\end{equation}

\pdfmarkupcomment[color=yellow]{Note that Equation {\ref{eq:knnL}} only holds as the number of samples approach infinity.
As the number of samples in this experiment is much lower than infinity, $\mbox{kNN}_L$ is not expected to hold and is simply treated as another estimator.
}{(RVR1.4) Added explanation to why BER may be lower than lower bounds.}

It is well known that the value of the parameter $k$ is key to the performance of the kNN classifier.
The optimal value of $k$ depends on the probability distributions of the classes, the number of samples available and the distance metric used \cite{samworth_optimal_2012, fukunaga_bayes_1987}.
For the distance measure, the Euclidean distance squared ($L_2^2$) is used.
Given that for this problem, the distribution is not known and, therefore, the optimal value of $k$ can not be computed, the method of \cite{renggli_evaluating_2021} will be modified slightly.
To make it more robust, the minimum kNN-LOO error for a range is taken:

\begin{equation}
k_0 = \arg\min_{k \in k_{range}} \mbox{kNN}_H(k).
\end{equation}

Then, the upper bound is taken as $\mbox{kNN}_H = \mbox{kNN}_H(k_0)$ with $\mbox{kNN}_L = \mbox{kNN}_L(k_0)$.
The midpoint between these two values ($\mbox{kNN}_H$ and $\mbox{kNN}_L$) is $\mbox{kNN}_M$. 
\subsection{Kernel Density Estimation}

Kernel Density Estimation (KDE) is a non-parametric method to estimate the probability density function (PDF) of a variable using samples from a population by applying kernel smoothing \cite{wand_kernel_1995}.
This section covers:

\noindent
\begin{minipage}{0.95\linewidth}
\ifx\templateType\templateArxiv\smallskip\fi
\smallskip
\begin{itemize}
    \item GKDE: Established KDE method
    \item CLAKDE: Comparative approach with Localized Adaptive KDE
    \item GC: Mean of GHP and CLAKDE
\end{itemize}
\smallskip
\end{minipage}

The GKDE estimator (referred to as "Gaussian KDE" or "KDE" in \cite{renggli_evaluating_2021}) is based upon using KDE to estimate the PDF for each individual class, using those respective samples.
Then the Bayes formula can be applied to estimate the BER, as described in the equations below.

Take a dataset $D$ of i.i.d. pairs $\{ (\mathbf{x}_i, y_i)\}^N_{i=1}$, where $\mathbf{x}_i$ is an observation of the random vector $\mathbf{X} \in R^d$ and $y_i$ is an observation of the random variable $Y \in \{1, 2, ..., m\}$. Here, $\mathbf{x}_i$ is the feature and $y_i$ is the associated label. For $k = 1, ..., m$, the prior label probabilities are represented by $p_k = P_c(Y=k)$ with $\sum^m_{k=1} p_k =1$ and the conditional class probability is $P(\mathbf{x} |Y=k)$.

\noindent
\begin{minipage}{1\linewidth}
\smallskip
\hspace{\parindent} \pdfmarkupcomment[color=yellow]{Based on this, the Bayes error rate can be rewritten as {\cite{sekeh_learning_2020}}:
}{(RVR1.3) Edited equation.}

\begin{equation}
\mbox{BER} = 1 - \int \max_{k \in \{1, 2, ..., m\}} \{P(\mathbf{x} | Y=k) \} d\mathbf{x}
\label{eq:kde_ber_est}
\end{equation}
\smallskip
\end{minipage}

For the dataset $D$, a subset of samples having label $k$: $D^{(k)} = \{ (\mathbf{x}_i, y_i)\}^N_{i=1, y_i=k}$ exist with a cardinality of $n_k = \sum^n_{i=1} \mbox{I}(y_i = k)$ so that $n = \sum_{i=1}^m n_i$.
Then the estimator from \cite{renggli_evaluating_2021} can be written as:

\ifx\templateType\templateArxiv\small\fi
\begin{equation}
\widehat{\mbox{BER}}_H = 1 - \sum_{i=1}^{n} 
\max_{k \in \{1, 2, ..., m\}} \left[ \frac{L_H(\mathbf{x}_i ; D^{(k)})} {\sum_{j=0}^n L_H(\mathbf{x}_j ; D^{(k)})} \frac{n_k} {n} \right] 
\end{equation}
\ifx\templateType\templateArxiv\normalsize\fi

\noindent where $L_H(\mathbf{x}; D)$ is a density estimate with given bandwidth parameter $H$ for a dataset $D$.

In the case of GKDE \cite{renggli_evaluating_2021}, the density estimate is obtained by using multivariate kernel density estimation.
From subset $D^k$ the features vectors form the associated subset $\mathbf{X}^{(k)} = \{\mathbf{x}^{(k)}_1,  \mathbf{x}^{(k)}_2, ..., \mathbf{x}^{(k)}_{n_k}\}$, which contains all the points  with label $k$.
The kernel density estimation function is given as \cite{wand_kernel_1995}:

\ifx\templateType\templateArxiv\small\fi
\begin{equation}
L_H(\mathbf{x}; D^{(k)}) = \hat{f}_H(\mathbf{x}; \mathbf{X}^{(k)}) =  n_k^{-1} \sum^{n_k}_{i=1} K_H ( \mathbf{x}-\mathbf{x}^{(k)}_i)
\end{equation}
\ifx\templateType\templateArxiv\normalsize\fi

\noindent where $K_H$ is a kernel function $K$ with bandwidth matrix $H$ for multivariate density estimation.
For GKDE \cite{renggli_evaluating_2021}, this equation is simplified for a single bandwidth parameter, $h$ \cite{wand_kernel_1995}:

\begin{equation}
\hat{f}_H(\mathbf{x}; \mathbf{X}^{(k)}) =  n_k^{-1} h^{-d} \sum^{n_k}_{i=1} K ( (\mathbf{x}-\mathbf{x}^{(k)}_i)/h)
\end{equation}

\noindent with the Gaussian kernel:

\begin{equation}
K (x_h ) = (2 \pi)^{-d/2} \exp(-\frac{1}{2} x_h^T x_h)
\end{equation}

\noindent as per implementation details given in \cite{pedregosa_scikit-learn_2011}.

\ifx\templateType\templateIEEE\else
\begin{figure}[htb]
\centering
\ifx\templateType\templateArxiv
\includegraphics[width=0.75\linewidth]{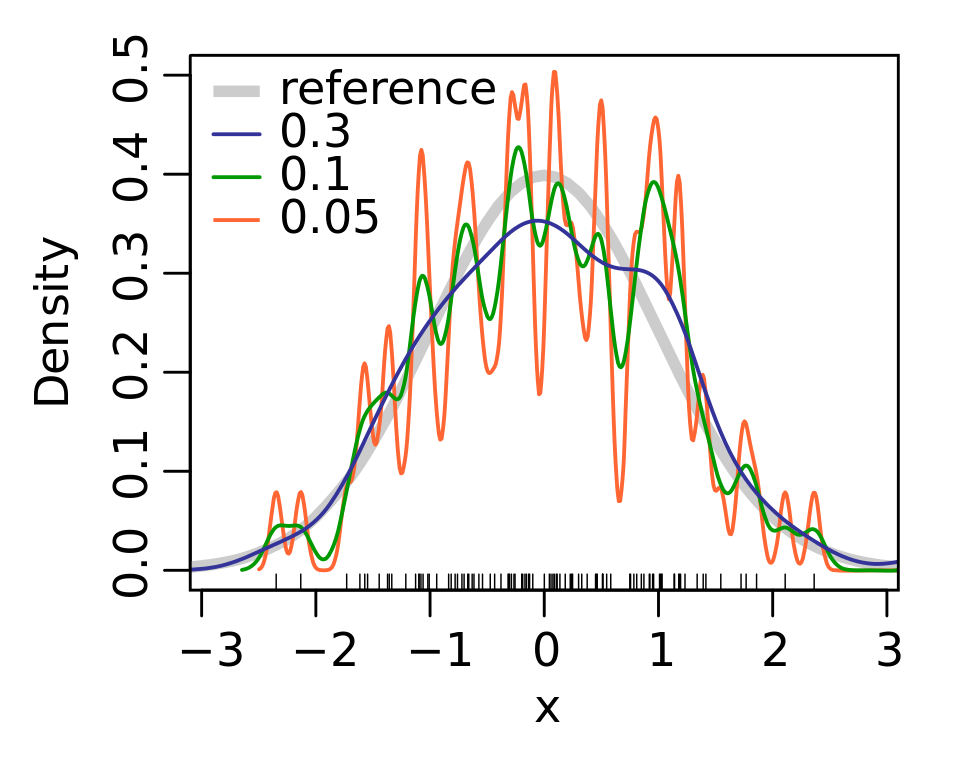}
\else
\includegraphics[width=0.6\linewidth]{Kernel_density.png}
\fi
\caption{Several uni-variate KDE functions using different smoothing bandwidths generated from 100 samples randomly drawn from a Gaussian distribution \cite{Toews_kernel_2007}.}
\label{fig:kde_ex}
\ifx\templateType\templateArxiv \vspace{-4mm} \fi
\end{figure} \fi

\ifx\templateType\templateIEEE
The
\else
As illustrated in Figure \ref{fig:kde_ex}, the
\fi
bandwidth parameter $h$ can have a very large impact on the generated estimation, which is not desirable (R.{\ref{req:param}}).
While there are known selection guidelines for approximating Gaussian distributions, that assumption does not hold (R.{\ref{req:robustToDist}}).
However, based on those formulas \cite{wand_kernel_1995}, it can be expected that the ideal bandwidth would depend on the number of samples and number of features.

\pdfmarkupcomment[color=yellow]{
For further information, we can look at kNN as, even though KDE and kNN are presented here as different types of methods, they are actually quite similar.
In KDE, each class is effectively a weighted kNN where all points of one class are used with a Gaussian kernel for the distance.
Given this relationship, and given how optimal kNN parameters are based on the distributions {\cite{fukunaga_bayes_1987}}, it is likely that the optimal parameters are dependent on the type of distribution as well.
For consistency, all the values from {\cite{renggli_evaluating_2021}} will be used and are listed in Table {\ref{table:estParam}}.
}{(RVR1.2) Explictly stated connection between kNN and KDE.}

\pdfmarkupcomment[color=yellow]{While KDEs do eventually converge to the PDF, they}{Edited.} are known to be highly susceptible to the curse of dimensionality, requiring larger numbers of samples every time a feature is added to maintain the accuracy of the estimation \cite{scott_feasibility_1991}.
This is an issue for R.{\ref{req:features}}, \pdfmarkupcomment[color=yellow]{and an issue that will be attempted to address by taking a different, comparative approach}{Edited.}.

\subsubsection{Comparative Density Estimation}

In this section, a new binary BER estimation technique is introduced.
This proposed technique is also based on density estimation, but with a different approach.
Instead of scoring the individual samples, this method can be seen as comparing the differences of the probability distributions.
It seeks to improve upon GKDE, with the following goals:

\noindent
\begin{minipage}{0.95\linewidth}
\smallskip
\begin{itemize}
\item Stabilize performance over different types of distributions.
(R.{\ref{req:robustToDist}})
\item Remove the need for user-defined parameters.
(R.{\ref{req:param}})
\item Stabilize accuracy with respect to the number of features. (R.{\ref{req:features}})
\end{itemize}
\smallskip
\end{minipage}

\noindent In addition, to compare the similarity of two distributions, the following properties are desirable:

\noindent
\begin{minipage}{0.95\linewidth}
\ifx\templateType\templateArxiv\small\fi
\smallskip
\begin{itemize}
\item Self similarity: $\mbox{CD}(f, f) = 1$
\item Self identification: $\mbox{CD}(f, f) = 1$\pdfmarkupcomment[color=yellow]{if and only if}{(RVR1.3) Edited.} $f=g$
\item Range: $\mbox{CD}(f, g) \in [0, 1]$
\item Symmetry: $\mbox{CD}(f, g) = \mbox{CD}(g, f)$
\end{itemize}
\ifx\templateType\templateArxiv\normalsize\fi
\smallskip
\end{minipage}

One of the issues of GKDE in higher dimensions is that it tends to underestimate the error.
In Equation \ref{eq:kde_ber_est}, the estimate is based on the maximum likelihood and sum of likelihoods.
The estimated probability that the points come from their own class distributions becomes very large, while the estimated probability those points come from the other class becomes small.
This could be due to the changing distance calculation, or the shape of the KDE.

\ifx\templateType\templateIEEE
When using KDE at smaller bandwidths the estimated probability around the points used to create the estimate is overestimated, a similar effect tends to happen in higher dimensions.
\else
As seen in Figure \ref{fig:kde_ex}, at smaller bandwidths the estimated probability around the points used to create the estimate is overestimated, a similar effect tends to happen in higher dimensions.
\fi
\pdfmarkupcomment[color=yellow]{
Instead of attempting to adjust the bandwidth for a more accurate estimation, it is possible to remove the point to be scored from the KDE, a LOO approach (similar to the approach for kNN).
This creates a more even playing field between the scores for each class, reducing the tendency to underestimate the error at higher numbers of features (R.{\ref{req:features}}).
}{(RVR1.2) Added more explanation to reasoning behind why CLAKDE should be more robust to dimensionality.}

Given two sets of $d$-variate points representing two different classes: $A = (a_1, a_2, ... a_n)$ with $a_n=(a_{n1},a_{n2},...,a_{nd})$ and $B = (b_1, b_2, ... b_m)$ with $b_m=(b_{m1},b_{m2},...,b_{md})$, these points can be scored against their own classes:

\begin{align}
g_{AA} = \sum^{n_A}_{i=1} \frac{\hat{f}(\mathbf{x}^{(A)}_i ; \{x : x \in \mathbf{X}^{(A)}, x \neq \mathbf{x}^{(A)}_i \})}{n_A} \\
g_{BB} = \sum^{n_B}_{i=1} \frac{\hat{f}(\mathbf{x}^{(B)}_i ; \{x : x \in \mathbf{X}^{(B)}, x \neq \mathbf{x}^{(B)}_i \})}{n_B}
\end{align}

Where $\hat{f}(x; X)$ is a density estimator of $x$ given data $X$.
The points can also be scored against their opposing classes:

\begin{align}
g_{BA} = \sum^{n_B}_{i=1} \frac{\hat{f}(\mathbf{x}^{(B)}_i ; \mathbf{X}^{(A)})}{n_B} \\
g_{AB} = \sum^{n_A}_{i=1} \frac{\hat{f}(\mathbf{x}^{(A)}_i ; \mathbf{X}^{(B)})}{n_A}
\end{align}

\noindent Thus, a measure can be constructed from the comparison:

\begin{equation}
    J = \frac{g_{AB}+g_{BA}}{g_{AA}+g_{BB}}
\end{equation}

\noindent Which can also be written as:
\begin{equation} \label{eq:jLog}
    J = \log (g_{AB}+g_{BA}) - \log (g_{AA}+g_{BB})
\end{equation}

Where $J$ is the estimated similarity of the probability distributions.
As a similarity of 1 indicates the same distributions and results in an error of 0.5, the estimator becomes: $\widehat{\mbox{BER}}= J/2$.
Note the assumption that the probability of drawing from either class is the same.

This strategy can be thought of as, if the chance of drawing the points from their own class is the same as drawing them from the other class, then the probability distributions are similar.
Likewise, if the chance the drawn points come from their own classes and not the other, then the probability distributions are not very similar, and the BER ought to be low.

It should be noted that this method is more difficult to implement than GKDE, as the density estimates are no longer the same for all points.
\pdfmarkupcomment[color=yellow]{However, the score can now be computed with logarithms (Equation {\ref{eq:jLog}}).
This offers some robustness to floating point errors, which can become an issue when dealing with Gaussian distributions in higher dimensions due to the very small probability values (R.{\ref{req:features}}).
}{(RVR1.2) Added further explanation to the theory behind CLAKDE.}

For CLAKDE, this comparative method is used with Locally Adaptive Kernel Density Estimation (LAKDE) {\cite{olsen_think_2024}} as the density estimator.
The advantage of this kernel optimization method is two-fold: the user does not need to supply any parameters (R.{\ref{req:param}}) and the density estimation should perform better over a wide range of distributions (R.{\ref{req:robustToDist}}).
Note that the density estimate optimized for class $j$: $\mbox{LAKDE}_j(\mathbf{x}, \mathbf{X})$, for point $\mathbf{x}$ and training points $\mathbf{X}$, can be optimized using all data for that particular class $\mathbf{X}^{(j)}$ to save computation time, as the kernel is associated with the class distribution itself. 
However, the density estimate of the class with itself must still exclude the point being evaluated:

\ifx\templateType\templateArxiv\small\fi
\begin{equation}
    g_{jj} = \sum^{n_j}_{i=1} \frac{\mbox{LAKDE}_j (\mathbf{x}^{(j)}_i ; \{x : x \in \mathbf{X}^{(j)}, x \neq \mathbf{x}^{(j)}_i \})}{n_j} \\
\end{equation}
\ifx\templateType\templateArxiv\normalsize\fi

\subsubsection{GC}

\pdfmarkupcomment[color=yellow]{
As GHP and CLAKDE are the two estimators designed for large numbers of features, but with different approaches, there is the possibility that the combination of both of them may lead to a better result than each one individually.
So, in addition to running each separately, we also include GC, so for any fixed dataset: $\mbox{GC} = (\mbox{GHP}_L + \mbox{CLAKDE})/2$.
Here, the mean is taken for simplicity, as we have no additional information about how they should be weighted.
}{(RVR1.2) Added section to introduce GC.
(RVR2.6) Clarified the combination of GHP and CLAKDE is used because they are supposed to be the best estimators for high-dimensional problems.
}

\FloatBarrier

\ifx\templateType\templateIEEE
\renewcommand{\arraystretch}{1.1}
\begin{table}[htb]
\centering
\begin{tabular}{ccScSl}

Name & & \makecell[c]{Distribution\\Type} & Parameters \\
\hline

\multirow{4}{*}{GvG}
& Class A
& \makecell[l]{Multivariate\\Gaussian}
& \makecell[l]{$\mu_A = 0$, \\ $\sigma^2_A > 0.01$} \\
\cline{2-4}
& Class B
& \makecell[l]{Multivariate\\Gaussian}
& \makecell[l]{ $\sigma^2_B > 0.01$,
$\mu \geq 0$, \\
$\mu_B = \vec1 \sqrt{\mu^2/d}$} \\
\hline

\multirow{6}{*}{TvT}
& Class A
& \makecell[l]{Three-mode\\Gaussian\\Mixture}
& \makecell[l]{$\sigma^2_A > 0.01$, $\mu > 0$, \\
$\mu_A = \sqrt{\mu^2/d}$\\
$\mu_1 = \vec0$, \\
$\mu_2 = \vec1 \mu_A$, \\
$\mu_3 = -\vec1 \mu_A$} \\
\cline{2-4}
& Class B
& \makecell[l]{Two-mode\\Gaussian\\Mixture}
& \makecell[l]{
$\mu_B = \sqrt{(\mu/2)^2/d}$\\
$\mu_1 = \mu_B \vec1$, \\
$\mu_2 = -\mu_B \vec1$, \\
$\sigma^2_B > 0.01$} \\
\hline

\multirow{5}{*}{TvS}
& Class A
& \makecell[l]{Three-mode\\Gaussian\\Mixture}
& \makecell[l]{$\sigma^2_A > 0.01$, $\mu > 0$, \\
$\mu_A = \sqrt{\mu^2/d}$,\\
$\mu_1 = \vec0$, \\
$\mu_2 = \mu_A \vec1$, \\
$\mu_3 = -\mu_A \vec1$} \\
\cline{2-4}
& Class B
& \makecell[l]{Gaussian\\Mixture of\\Hypersphere}
& \makecell[l]{ $r_B = \mu / 2$, \\
$\sigma^2_B > 0.1$} \\
\hline

\multirow{4}{*}{SvS}
& Class A
& \makecell[l]{Gaussian\\Mixture of\\Hypersphere\\with Center}
& \makecell[l]{$n_{HS} = 190 \pm 2$,\\$r_A > 0$, \\ $\sigma^2_A > 0.1$} \\
\cline{2-4}
& Class B
& \makecell[l]{Gaussian\\Mixture of\\Hypersphere}
& \makecell[l]{$n_{HS} = 200 \pm 2$,\\$r_B = r_A/2$, \\
$\sigma^2_B > 0.1$} \\
\hline

\end{tabular}

\caption{Distribution types used for simulations and associated allowed parameters. $d$ is the number of features.}
\label{table:dparam}
\ifx\templateType\templateArxiv \vspace{-4mm} \fi

\end{table} \begin{table}[ht]
\centering
\begin{tabular}{cSc} 
\hline

Estimator & Parameters\\ \hline
\makecell[c]{$\mbox{kNN}_L$, \\ $\mbox{kNN}_M$, \\ $\mbox{kNN}_H$} & $k_{range} = \{ x | 1 \leq x \leq 199, 2 \nmid x \}$ \\ \hline
GKDE & $h = \{ 0.0025, 0.05, 0.1, 0.25, 0.5 \} $ \\ \hline
CLAKDE & None \\ \hline
\makecell[c]{$\mbox{GHP}_L$, \\ $\mbox{GHP}_M$,\\$\mbox{GHP}_H$} & None \\ \hline
NB & None \\ \hline
GC & None \\ \hline
\end{tabular}

\caption{Estimator Parameters}
\label{table:estParam}

\ifx\templateType\templateArxiv \vspace{-4mm} \fi

\end{table} \fi

\section{Experiment}

The estimators are tested through a series of simulations on different class distributions (R.{\ref{req:robustToDist}}) over a range of BER values.
Each distribution type is composed of two classes with different probability distributions and different parameters.
Overview of the simulations:

\noindent
\begin{minipage}{0.95\linewidth}
\smallskip
\begin{enumerate}
\item Select a distribution type for both classes.
\item Select distribution parameters with associated ground truth BER.
\item Generate the requested number of points from the distributions and run the estimators.
\end{enumerate}
\smallskip
\end{minipage}

\noindent
Afterwards, the results are compiled and the error is calculated for each estimator.

\ifx\templateType\templateArxiv
\renewcommand{\arraystretch}{1.1}
\begin{table}[htb]
\centering
\begin{tabular}{ccScSl}

Name & & \makecell[c]{Distribution\\Type} & Parameters \\
\hline

\multirow{4}{*}{GvG}
& Class A
& \makecell[l]{Multivariate\\Gaussian}
& \makecell[l]{$\mu_A = 0$, \\ $\sigma^2_A > 0.01$} \\
\cline{2-4}
& Class B
& \makecell[l]{Multivariate\\Gaussian}
& \makecell[l]{ $\sigma^2_B > 0.01$,
$\mu \geq 0$, \\
$\mu_B = \vec1 \sqrt{\mu^2/d}$} \\
\hline

\multirow{6}{*}{TvT}
& Class A
& \makecell[l]{Three-mode\\Gaussian\\Mixture}
& \makecell[l]{$\sigma^2_A > 0.01$, $\mu > 0$, \\
$\mu_A = \sqrt{\mu^2/d}$\\
$\mu_1 = \vec0$, \\
$\mu_2 = \vec1 \mu_A$, \\
$\mu_3 = -\vec1 \mu_A$} \\
\cline{2-4}
& Class B
& \makecell[l]{Two-mode\\Gaussian\\Mixture}
& \makecell[l]{
$\mu_B = \sqrt{(\mu/2)^2/d}$\\
$\mu_1 = \mu_B \vec1$, \\
$\mu_2 = -\mu_B \vec1$, \\
$\sigma^2_B > 0.01$} \\
\hline

\multirow{5}{*}{TvS}
& Class A
& \makecell[l]{Three-mode\\Gaussian\\Mixture}
& \makecell[l]{$\sigma^2_A > 0.01$, $\mu > 0$, \\
$\mu_A = \sqrt{\mu^2/d}$,\\
$\mu_1 = \vec0$, \\
$\mu_2 = \mu_A \vec1$, \\
$\mu_3 = -\mu_A \vec1$} \\
\cline{2-4}
& Class B
& \makecell[l]{Gaussian\\Mixture of\\Hypersphere}
& \makecell[l]{ $r_B = \mu / 2$, \\
$\sigma^2_B > 0.1$} \\
\hline

\multirow{4}{*}{SvS}
& Class A
& \makecell[l]{Gaussian\\Mixture of\\Hypersphere\\with Center}
& \makecell[l]{$n_{HS} = 190 \pm 2$,\\$r_A > 0$, \\ $\sigma^2_A > 0.1$} \\
\cline{2-4}
& Class B
& \makecell[l]{Gaussian\\Mixture of\\Hypersphere}
& \makecell[l]{$n_{HS} = 200 \pm 2$,\\$r_B = r_A/2$, \\
$\sigma^2_B > 0.1$} \\
\hline

\end{tabular}

\caption{Distribution types used for simulations and associated allowed parameters. $d$ is the number of features.}
\label{table:dparam}
\ifx\templateType\templateArxiv \vspace{-4mm} \fi

\end{table} \begin{table}[ht]
\centering
\begin{tabular}{cSc} 
\hline

Estimator & Parameters\\ \hline
\makecell[c]{$\mbox{kNN}_L$, \\ $\mbox{kNN}_M$, \\ $\mbox{kNN}_H$} & $k_{range} = \{ x | 1 \leq x \leq 199, 2 \nmid x \}$ \\ \hline
GKDE & $h = \{ 0.0025, 0.05, 0.1, 0.25, 0.5 \} $ \\ \hline
CLAKDE & None \\ \hline
\makecell[c]{$\mbox{GHP}_L$, \\ $\mbox{GHP}_M$,\\$\mbox{GHP}_H$} & None \\ \hline
NB & None \\ \hline
GC & None \\ \hline
\end{tabular}

\caption{Estimator Parameters}
\label{table:estParam}

\ifx\templateType\templateArxiv \vspace{-4mm} \fi

\end{table} \fi

\subsection{Ground Truth Bayes Error Rate} \label{sec:gt}

The values used as the ground truth of the BER are generated via Monte Carlo simulation.
For each set of distribution parameters, a large number of samples 1024000 in total, calculated from 1000 batches of 1024) are used for the high precision estimation that represents the true value of the BER.
This is the same method used in \cite{sekeh_learning_2020}, and a similar version was used in \cite{chen_evaluating_2023}.

Each distribution type has several controllable parameters (listed in Table \ref{table:dparam}) that can be changed to control the level of mixture between the two classes, thus controlling the BER.
In order to have a wide range of BER values to use for the simulations, the parameters are iteratively sampled until a minimum gap of 1\% is reached between the nearest BER values.
The implementation is configured to focus on changes in offset ($\mu$) rather than variance.
As when the variance is very large, compared to the offset, the structure of the distributions start to disappear, an unwanted effect.
Therefore, the ground truth of BER values are generated primarily on differences in offset.
Once the list of BER values and parameters is generated, it can then be sampled uniformly to generate a number of distribution parameters that evenly cover the requested BER range.

\subsection{Application of Estimators}

Each simulation is conducted by generating samples from a set of distribution parameters, then evaluating all estimators on the same data.
The same number of points is generated for each class.
There is no pre-processing, such as feature-by-feature max-min scaling, as this is dependent on the samples obtained, and uneven scaling can change the underlying BER.
Additional parameters for each estimator are covered in Table \ref{table:estParam}.
In addition to the previously discussed estimators (see Section {\ref{sec:estimators}}), the Naive Bayes Classification Training Error (NB) is also applied as an estimator for verification and comparison purposes.

Each set of distribution parameters may be run multiple times with different seed values in order to produce the number of requested simulations from a smaller set of BER values.
The list of the parameters are selected to be uniform over the requested range of BER values, so that the results are not skewed towards specific areas of the BER range.
This must be done because the accuracy of the estimators can vary based on the true BER, this will be further explained in Section \ref{sec:dis}.
In this case, it is assumed that no prior information is known about the BER, so the full range must be taken into consideration.

\ifx\templateType\templateIEEE\begin{figure*}[b]
  \begin{minipage}{\linewidth}
\subcaptionbox{Estimator: $\mbox{KNN}_L$. MSE: 11.6. 95\% error bounds: (-5.0, 9.6).\\
    $R^2$ (Linear fit): 0.95. $R^2$ (LEOSS fit): 0.98.}
      {\includegraphics[width=0.49\linewidth]{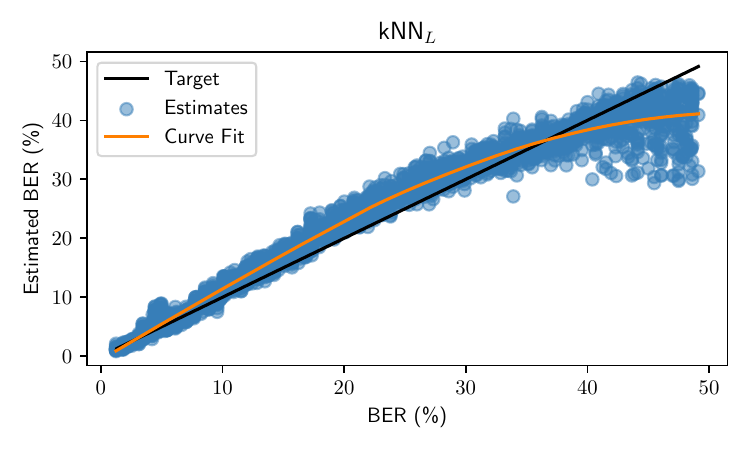}}
      \hfill
    \subcaptionbox{Estimator: $\mbox{KNN}_H$. MSE: 27.6. 95\% error bounds: (-10.4, 0.3).\\
    $R^2$ (Linear fit): 0.96. $R^2$ (LEOSS fit): 0.98.}
      {\includegraphics[width=0.49\linewidth]{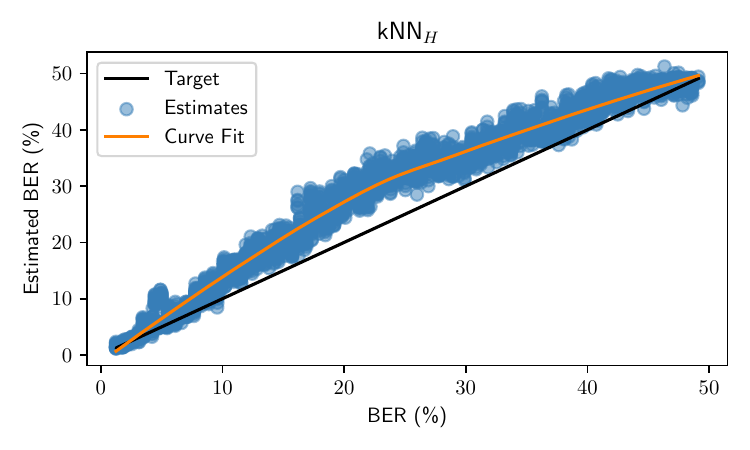}}
    \caption{Comparison of KNN estimators. Distribution Type: TvT. Features: 20. Samples per Class: 1000.}
    \label{fig:ex-KNN}
  \end{minipage}\quad
  \ifx\templateType\templateArxiv \vspace{-4mm} \fi
\end{figure*}

\begin{figure*}[b]
  \begin{minipage}{\linewidth}
\subcaptionbox{Estimator: $\mbox{GHP}_L$. MSE: 13.3. 95\% error bounds: (-6.7, 4.2).\\
    $R^2$ (Linear fit): 0.97. $R^2$ (LEOSS fit): 0.97.}
      {\includegraphics[width=0.49\linewidth]{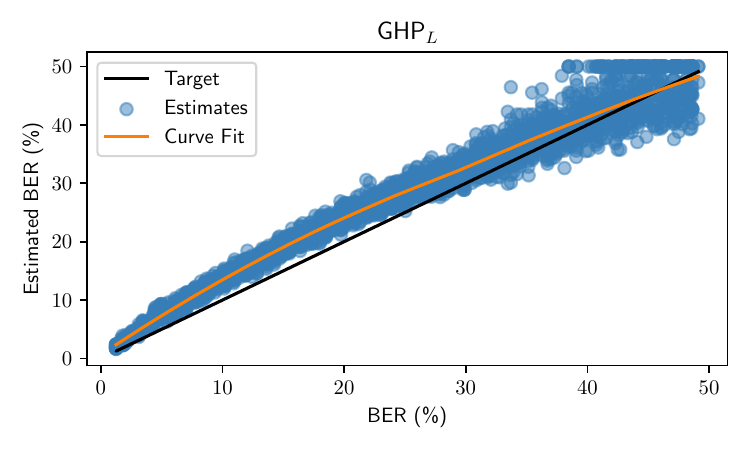}}
      \hfill
    \subcaptionbox{Estimator: $\mbox{GHP}_H$. MSE: 151.2. 95\% error bounds: (-17.9, -1.9).\\
    $R^2$ (Linear fit): 0.89. $R^2$ (LEOSS fit): 0.99.}
      {\includegraphics[width=0.49\linewidth]{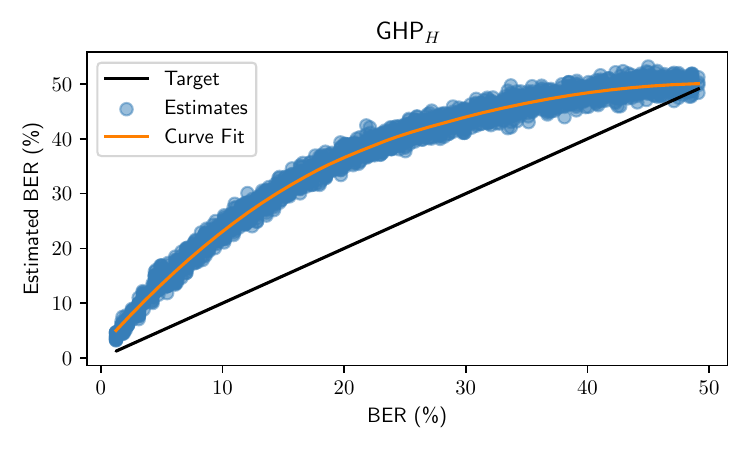}}
      \hfill
    \caption{Comparison of GHP estimators. Distribution Type: TvT. Features: 20. Samples per Class: 1000.}
    \label{fig:ex-GHP}
  \end{minipage}\quad
  \ifx\templateType\templateArxiv \vspace{-4mm} \fi
\end{figure*} \fi
\ifx\templateType\templateArxiv\vspace{-3mm}\fi
\subsection{Error Calculation}

While other literature often uses the BER as a fraction for error calculation, in this case percentage is preferred to evaluate against (R.{\ref{req:errorBounds}}).
Therefore, the estimator error is calculated as: 
$E = 100\% * \mbox{BER} - 100\% * \widehat{\mbox{BER}} $ and the MSE is $\frac{1}{n} \sum^n_{i=1} E_i^2 $.
Note that this means the results in this case can appear much larger in comparison to similar works in this area due to this difference, but percentages are better suited reporting method for the application.
Likewise, the 95\% error bounds (2.5\% to 97.5\%) are calculated based on percentiles (R.{\ref{req:errorBounds}}).

\ifx\templateType\templateArxiv\vspace{-3mm}\fi
\subsection{Implementation}

The program used to conduct the simulations is provided at
\href{https://github.com/LesleyWheat/bayesErrorRate-estimatorTester}{github.com/LesleyWheat/bayesErrorRate-estimatorTester}.
For ease of installation, it contains the required components from: \cite{renggli_evaluating_2021} and \cite{olsen_think_2024}.
It also directly relies upon the following libraries: \cite{pedregosa_scikit-learn_2011}, \cite{virtanen_scipy_2020}, \cite{harris_array_2020}, \cite{hunter_matplotlib_2007}, \cite{paszke_pytorch_2019}, \cite{the_mpmath_development_team_mpmath_2023}, \cite{seabold_statsmodels_2010} and \cite{mckinney_data_2010}.

\pdfmarkupcomment[color=yellow]{
For comparability, the original code for the existing estimators was kept as similar as possible to the original implementation.
Subsequently, estimators that did not use GPU-acceleration were not converted, so some estimators utilize GPUs and others do not.
When tested, it was determined that differences in floating point computation were negligible and should not impact the final results.
}{(RVR2.4) Added additional implementation details.}

\pdfmarkupcomment[color=yellow]{
While computation time is not part of the requirements (Section {\ref{sec:estimators}}), each simulation (including all estimators) does complete in a reasonable time on the system (less than 200 seconds).
To achieve this, GPU-acceleration is effectively a requirement for some estimators (notably CLAKDE).
This is only important to insure the total time for all the simulations stays in a realistic timeframe, but may be relaxed in target applications.
}{(RVR2.4) Added details on computation time.}

\ifx\templateType\templateIEEE
\begin{figure}[tb]
\centering
\includegraphics[width=0.9\linewidth]{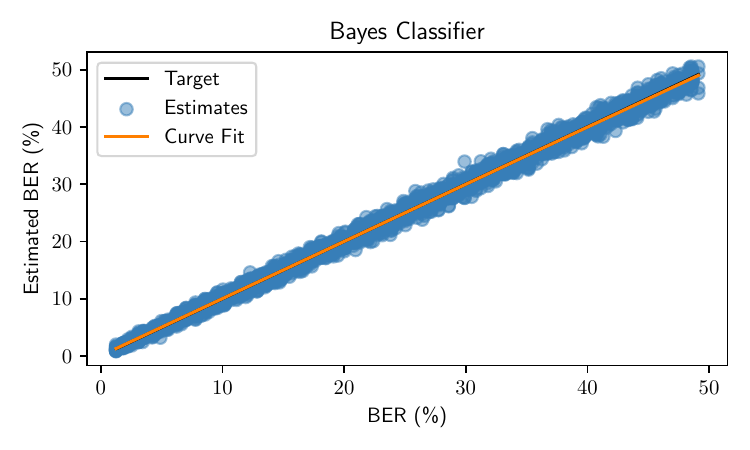}

\caption{Example of simulation results from the Bayes Classifier.
Distribution Type: TvT. \\
Features: 20. Samples per Class: 1000.\\
95\% Error Bounds: (-1.7, 1.9). MSE: 0.8. }
\label{fig:bc-error}
\ifx\templateType\templateArxiv \vspace{-4mm} \fi
\end{figure} \else
\begin{figure}[tb]
\centering
\includegraphics[width=0.9\linewidth]{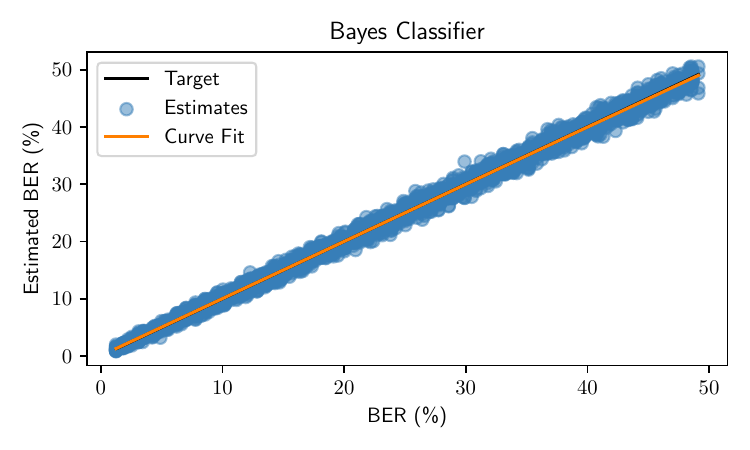}

\caption{Example of simulation results from the Bayes Classifier.
Distribution Type: TvT. \\
Features: 20. Samples per Class: 1000.\\
95\% Error Bounds: (-1.7, 1.9). MSE: 0.8. }
\label{fig:bc-error}
\ifx\templateType\templateArxiv \vspace{-4mm} \fi
\end{figure} \begin{figure*}[b]
  \begin{minipage}{\linewidth}
\subcaptionbox{Estimator: $\mbox{KNN}_L$. MSE: 11.6. 95\% error bounds: (-5.0, 9.6).\\
    $R^2$ (Linear fit): 0.95. $R^2$ (LEOSS fit): 0.98.}
      {\includegraphics[width=0.49\linewidth]{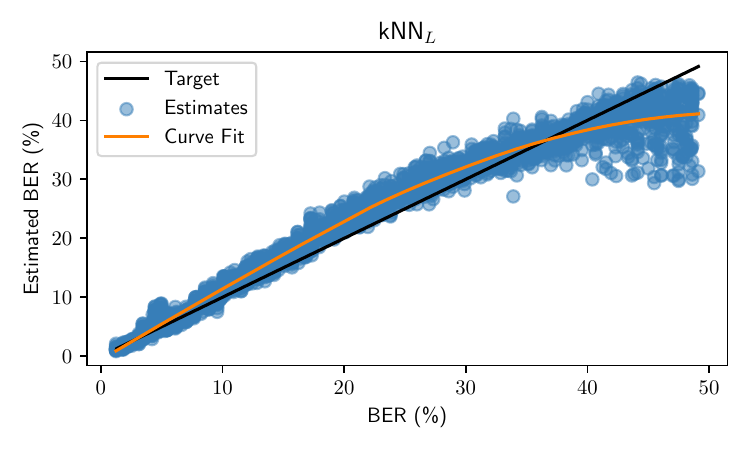}}
      \hfill
    \subcaptionbox{Estimator: $\mbox{KNN}_H$. MSE: 27.6. 95\% error bounds: (-10.4, 0.3).\\
    $R^2$ (Linear fit): 0.96. $R^2$ (LEOSS fit): 0.98.}
      {\includegraphics[width=0.49\linewidth]{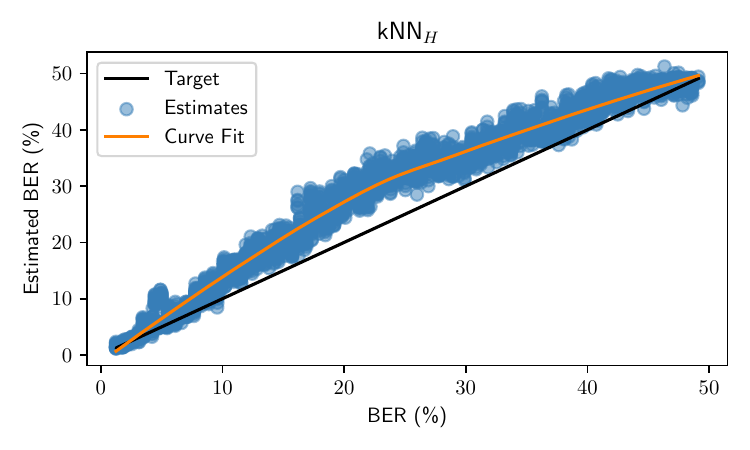}}
    \caption{Comparison of KNN estimators. Distribution Type: TvT. Features: 20. Samples per Class: 1000.}
    \label{fig:ex-KNN}
  \end{minipage}\quad
  \ifx\templateType\templateArxiv \vspace{-4mm} \fi
\end{figure*}

\begin{figure*}[b]
  \begin{minipage}{\linewidth}
\subcaptionbox{Estimator: $\mbox{GHP}_L$. MSE: 13.3. 95\% error bounds: (-6.7, 4.2).\\
    $R^2$ (Linear fit): 0.97. $R^2$ (LEOSS fit): 0.97.}
      {\includegraphics[width=0.49\linewidth]{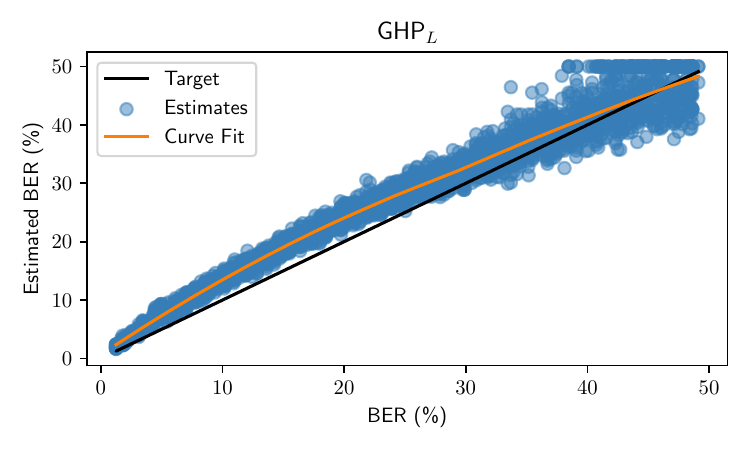}}
      \hfill
    \subcaptionbox{Estimator: $\mbox{GHP}_H$. MSE: 151.2. 95\% error bounds: (-17.9, -1.9).\\
    $R^2$ (Linear fit): 0.89. $R^2$ (LEOSS fit): 0.99.}
      {\includegraphics[width=0.49\linewidth]{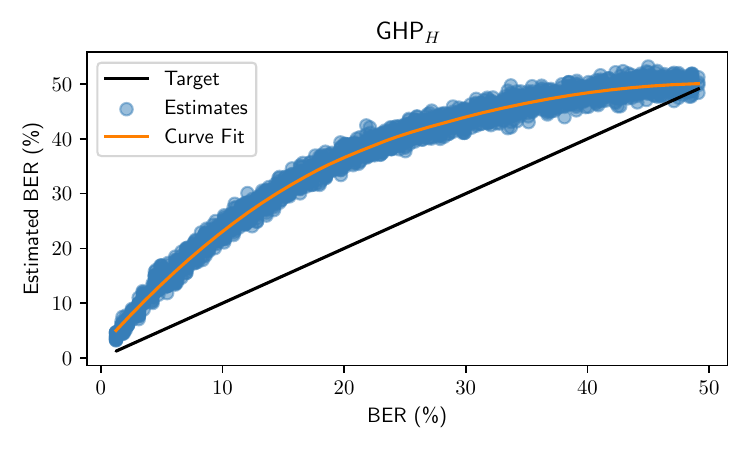}}
      \hfill
    \caption{Comparison of GHP estimators. Distribution Type: TvT. Features: 20. Samples per Class: 1000.}
    \label{fig:ex-GHP}
  \end{minipage}\quad
  \ifx\templateType\templateArxiv \vspace{-4mm} \fi
\end{figure*} \fi

\section{Results}

This section will cover the simulation results and observed behavior of the estimators.
The error of each set of simulations (specific test case, number of features and number of samples) is calculated from the output of over 2500 Monte Carlo runs.
Figures {\ref{fig:ex-KNN}}, {\ref{fig:ex-GHP}} and {\ref{fig:bc-error}} are examples of the results of a single estimator on one set of simulations.
In these figures, the blue dots are simulation results and the black line represents the ground truth, the ideal estimator.
The orange line is the result of locally estimated scatterplot smoothing (LEOSS) curve fitting, to highlight the underlying relationship between the true and predicted values.

All error bounds are expected to be to within $\pm 0.2 \%$, including the tables included in this section (Tables {\ref{table:gvg-best-bounds}}, {\ref{table:tvt-best-bounds}}, {\ref{table:tvs-best-bounds}} and {\ref{table:svs-best-bounds}}) and Appendix {\ref{sec:AppB}}.
First value is always 2.5\% error confidence and second value is 97.5\% error confidence.

\renewcommand{\arraystretch}{1}
\ifx\templateType\templateIEEE
\begin{table*}[b]
\else
\begin{table*}[h]
\fi
\centering

\begin{minipage}{1\linewidth}
    \centering
    \subcaptionbox{All estimators.}{\begin{tabular*}{\linewidth}{@{\extracolsep{\fill}}cccccccc}\\
&&\multicolumn{6}{c}{Number of Features}\\\cline{3-8}
&&2&3&4&10&20&30\\ \cline{2-8}
\parbox[t]{3mm}{\multirow{10}{*}{\rotatebox[origin=c]{90}{Number of Samples Per Class}}}&\multirow{2}{*}{500}&\cellcolor{gray!25}$\mbox{NB}$&$\mbox{NB}$&\cellcolor{gray!25}$\mbox{kNN}_{H}$&$\mbox{NB}$&\cellcolor{gray!25}$\mbox{NB}$&$\mbox{NB}$\\
&&\cellcolor{gray!25}(-2.4, 2.8)&(-2.2, 2.9)&\cellcolor{gray!25}(-2.6, 2.7)&(-1.8, 4.3)&\cellcolor{gray!25}(-1.4, 6.3)&(-1.0, 8.1)\\ \cline{2-8}
&\multirow{2}{*}{1000}&$\bm{\mbox{NB}}$&\cellcolor{gray!25}$\bm{\mbox{NB}}$&$\bm{\mbox{NB}}$&\cellcolor{gray!25}$\bm{\mbox{NB}}$&$\mbox{NB}$&\cellcolor{gray!25}$\mbox{NB}$\\
&&\textbf{(-1.7, 2.1)}&\cellcolor{gray!25}\textbf{(-1.7, 2.0)}&\textbf{(-1.6, 2.1)}&\cellcolor{gray!25}\textbf{(-1.5, 2.8)}&(-1.2, 4.1)&\cellcolor{gray!25}(-1.0, 5.1)\\ \cline{2-8}
&\multirow{2}{*}{1500}&\cellcolor{gray!25}$\bm{\mbox{NB}}$&$\bm{\mbox{NB}}$&\cellcolor{gray!25}$\bm{\mbox{NB}}$&$\bm{\mbox{NB}}$&\cellcolor{gray!25}$\bm{\mbox{NB}}$&$\bm{\mbox{NB}}$\\
&&\cellcolor{gray!25}\textbf{(-1.4, 1.5)}&\textbf{(-1.4, 1.7)}&\cellcolor{gray!25}\textbf{(-1.4, 1.7)}&\textbf{(-1.2, 2.2)}&\cellcolor{gray!25}\textbf{(-1.0, 3.0)}&\textbf{(-0.8, 3.9)}\\ \cline{2-8}
&\multirow{2}{*}{2000}&$\bm{\mbox{NB}}$&\cellcolor{gray!25}$\bm{\mbox{NB}}$&$\bm{\mbox{NB}}$&\cellcolor{gray!25}$\bm{\mbox{NB}}$&$\bm{\mbox{NB}}$&\cellcolor{gray!25}$\bm{\mbox{NB}}$\\
&&\textbf{(-1.2, 1.3)}&\cellcolor{gray!25}\textbf{(-1.3, 1.4)}&\textbf{(-1.2, 1.4)}&\cellcolor{gray!25}\textbf{(-1.1, 1.8)}&\textbf{(-0.9, 2.4)}&\cellcolor{gray!25}\textbf{(-0.7, 3.3)}\\ \cline{2-8}
&\multirow{2}{*}{2500}&\cellcolor{gray!25}$\bm{\mbox{NB}}$&$\bm{\mbox{NB}}$&\cellcolor{gray!25}$\bm{\mbox{NB}}$&$\bm{\mbox{NB}}$&\cellcolor{gray!25}$\bm{\mbox{NB}}$&$\bm{\mbox{NB}}$\\
&&\cellcolor{gray!25}\textbf{(-1.1, 1.2)}&\textbf{(-1.1, 1.3)}&\cellcolor{gray!25}\textbf{(-1.1, 1.3)}&\textbf{(-1.0, 1.5)}&\cellcolor{gray!25}\textbf{(-0.8, 2.2)}&\textbf{(-0.7, 2.6)}\\ \cline{2-8}
\end{tabular*} }
\end{minipage}

\begin{minipage}{1\linewidth}
    \centering
    \subcaptionbox{All estimators not including Naive Bayes.}{\begin{tabular*}{\linewidth}{@{\extracolsep{\fill}}cccccccc}\\
&&\multicolumn{6}{c}{Number of Features}\\\cline{3-8}
&&2&3&4&10&20&30\\ \cline{2-8}
\parbox[t]{3mm}{\multirow{10}{*}{\rotatebox[origin=c]{90}{Number of Samples Per Class}}}&\multirow{2}{*}{500}&\cellcolor{gray!25}$\mbox{kNN}_{H}$&$\mbox{kNN}_{H}$&\cellcolor{gray!25}$\mbox{kNN}_{H}$&$\mbox{kNN}_{M}$&\cellcolor{gray!25}$\mbox{kNN}_{M}$&$\mbox{kNN}_{L}$\\
&&\cellcolor{gray!25}(-2.1, 3.2)&(-2.2, 3.0)&\cellcolor{gray!25}(-2.6, 2.7)&(-3.1, 5.0)&\cellcolor{gray!25}(-8.3, 5.2)&(-8.3, 11.0)\\ \cline{2-8}
&\multirow{2}{*}{1000}&$\bm{\mbox{kNN}_{H}}$&\cellcolor{gray!25}$\bm{\mbox{kNN}_{H}}$&$\bm{\mbox{kNN}_{H}}$&\cellcolor{gray!25}$\mbox{kNN}_{M}$&$\mbox{kNN}_{M}$&\cellcolor{gray!25}$\mbox{kNN}_{L}$\\
&&\textbf{(-1.6, 2.2)}&\cellcolor{gray!25}\textbf{(-1.8, 1.9)}&\textbf{(-2.0, 1.9)}&\cellcolor{gray!25}(-2.8, 3.7)&(-7.7, 3.9)&\cellcolor{gray!25}(-7.9, 9.0)\\ \cline{2-8}
&\multirow{2}{*}{1500}&\cellcolor{gray!25}$\bm{\mbox{kNN}_{H}}$&$\bm{\mbox{kNN}_{H}}$&\cellcolor{gray!25}$\bm{\mbox{kNN}_{H}}$&$\mbox{kNN}_{M}$&\cellcolor{gray!25}$\mbox{kNN}_{L}$&$\mbox{kNN}_{L}$\\
&&\cellcolor{gray!25}\textbf{(-1.6, 1.6)}&\textbf{(-1.5, 1.6)}&\cellcolor{gray!25}\textbf{(-1.8, 1.4)}&(-2.6, 3.3)&\cellcolor{gray!25}(-4.0, 6.8)&(-7.4, 8.0)\\ \cline{2-8}
&\multirow{2}{*}{2000}&$\bm{\mbox{kNN}_{H}}$&\cellcolor{gray!25}$\bm{\mbox{kNN}_{H}}$&$\bm{\mbox{kNN}_{H}}$&\cellcolor{gray!25}$\mbox{kNN}_{M}$&$\mbox{GHP}_{L}$&\cellcolor{gray!25}$\mbox{kNN}_{L}$\\
&&\textbf{(-1.4, 1.4)}&\cellcolor{gray!25}\textbf{(-1.6, 1.2)}&\textbf{(-1.6, 1.2)}&\cellcolor{gray!25}(-2.3, 2.8)&(-5.4, 2.9)&\cellcolor{gray!25}(-7.1, 7.3)\\ \cline{2-8}
&\multirow{2}{*}{2500}&\cellcolor{gray!25}$\bm{\mbox{kNN}_{H}}$&$\bm{\mbox{kNN}_{H}}$&\cellcolor{gray!25}$\bm{\mbox{kNN}_{H}}$&$\bm{\mbox{kNN}_{M}}$&\cellcolor{gray!25}$\mbox{GHP}_{L}$&$\mbox{kNN}_{L}$\\
&&\cellcolor{gray!25}\textbf{(-1.3, 1.2)}&\textbf{(-1.4, 1.1)}&\cellcolor{gray!25}\textbf{(-1.6, 1.1)}&\textbf{(-2.3, 2.6)}&\cellcolor{gray!25}(-5.0, 3.0)&(-7.2, 7.5)\\ \cline{2-8}
\end{tabular*} }
\end{minipage}

\caption{Type: GvG.
Best Estimator by lowest MSE and respective bounds.
Ranges under 5\% are in bold.}
\label{table:gvg-best-bounds}
\ifx\templateType\templateArxiv \vspace{-4mm} \fi

\end{table*}
 \subsection{Bayesian Classifier} \label{sec:bayesClassifier}

To clarify further, the expected BER is considered to be the BER for the two distributions, not the error rate of the Bayes Classifier on the set of samples given.
Therefore, the Bayes Classifier can also be run on the simulation samples as an estimator, an example of which is shown in Figure \ref{fig:bc-error}.
The number of samples drawn for simulations is much lower than the number used to calculate the ground truth BER, effectively each simulation acts as a smaller Monte Carlo simulation, and has a higher overall variance because of this.
Therefore, it should be noted that the deviation of the Bayes classier error from the BER is dependent purely on the number of samples used, and the number of features or distribution parameters has no impact.

This is important because it highlights that any particular set of samples may, by chance, be particularly \say{easy} or \say{difficult} to classify.
As such, the Bayes Classifier error represents a non-zero lower bound for the other estimators and gives a theoretical minimum number of points required to reach the error bound for any problem.
However, the number of samples required for the estimators is often much larger than this value, and therefore that will be the focus going forwards.

\renewcommand{\arraystretch}{1}
\ifx\templateType\templateArxiv
\begin{table*}[p]
\else
\begin{table*}[htb]
\fi

\centering

\ifx\templateType\templateArxiv
    \small
\fi

\begin{minipage}{0.98\linewidth}
\centering
\begin{tabular*}{\linewidth}{@{\extracolsep{\fill}}ccccccccc}\\
&&\multicolumn{7}{c}{Number of Features}\\\cline{3-9}
&&2&4&6&8&10&12&20\\ \cline{2-9}
\parbox[t]{3mm}{\multirow{10}{*}{\rotatebox[origin=c]{90}{Number of Samples Per Class}}}&\multirow{2}{*}{500}&\cellcolor{gray!25}$\mbox{kNN}_{H}$&$\mbox{kNN}_{H}$&\cellcolor{gray!25}$\mbox{kNN}_{M}$&$\mbox{kNN}_{M}$&\cellcolor{gray!25}$\mbox{kNN}_{M}$&$\mbox{GHP}_{L}$&\cellcolor{gray!25}$\mbox{kNN}_{L}$\\
&&\cellcolor{gray!25}(-2.5, 2.8)&(-3.6, 2.3)&\cellcolor{gray!25}(-2.6, 5.4)&(-3.2, 4.9)&\cellcolor{gray!25}(-4.9, 5.6)&(-7.0, 6.0)&\cellcolor{gray!25}(-6.0, 10.5)\\ \cline{2-9}
&\multirow{2}{*}{1000}&$\bm{\mbox{kNN}_{H}}$&\cellcolor{gray!25}$\bm{\mbox{kNN}_{H}}$&$\mbox{kNN}_{M}$&\cellcolor{gray!25}$\mbox{kNN}_{M}$&$\mbox{kNN}_{M}$&\cellcolor{gray!25}$\mbox{GHP}_{L}$&$\mbox{kNN}_{L}$\\
&&\textbf{(-1.8, 1.9)}&\cellcolor{gray!25}\textbf{(-2.3, 1.5)}&(-1.6, 3.8)&\cellcolor{gray!25}(-2.3, 4.0)&(-3.9, 4.1)&\cellcolor{gray!25}(-5.5, 4.8)&(-5.0, 9.6)\\ \cline{2-9}
&\multirow{2}{*}{1500}&\cellcolor{gray!25}$\bm{\mbox{kNN}_{H}}$&$\bm{\mbox{kNN}_{H}}$&\cellcolor{gray!25}$\bm{\mbox{kNN}_{H}}$&$\bm{\mbox{kNN}_{M}}$&\cellcolor{gray!25}$\mbox{kNN}_{M}$&$\mbox{GHP}_{L}$&\cellcolor{gray!25}$\mbox{kNN}_{L}$\\
&&\cellcolor{gray!25}\textbf{(-1.5, 1.6)}&\textbf{(-2.1, 1.2)}&\cellcolor{gray!25}\textbf{(-3.1, 0.8)}&\textbf{(-1.9, 2.9)}&\cellcolor{gray!25}(-3.3, 3.2)&(-4.8, 3.8)&\cellcolor{gray!25}(-4.4, 7.9)\\ \cline{2-9}
&\multirow{2}{*}{2000}&$\bm{\mbox{kNN}_{H}}$&\cellcolor{gray!25}$\bm{\mbox{kNN}_{H}}$&$\bm{\mbox{kNN}_{M}}$&\cellcolor{gray!25}$\bm{\mbox{kNN}_{M}}$&$\mbox{kNN}_{M}$&\cellcolor{gray!25}$\mbox{GHP}_{L}$&$\mbox{GHP}_{L}$\\
&&\textbf{(-1.4, 1.3)}&\cellcolor{gray!25}\textbf{(-1.9, 1.1)}&\textbf{(-1.0, 2.8)}&\cellcolor{gray!25}\textbf{(-1.6, 2.5)}&(-2.9, 3.0)&\cellcolor{gray!25}(-4.1, 3.7)&(-5.3, 3.1)\\ \cline{2-9}
&\multirow{2}{*}{2500}&\cellcolor{gray!25}$\bm{\mbox{kNN}_{H}}$&$\bm{\mbox{kNN}_{H}}$&\cellcolor{gray!25}$\bm{\mbox{kNN}_{H}}$&$\bm{\mbox{kNN}_{M}}$&\cellcolor{gray!25}$\mbox{kNN}_{M}$&$\mbox{GHP}_{L}$&\cellcolor{gray!25}$\mbox{kNN}_{L}$\\
&&\cellcolor{gray!25}\textbf{(-1.3, 1.2)}&\textbf{(-1.8, 0.9)}&\cellcolor{gray!25}\textbf{(-2.5, 0.7)}&\textbf{(-1.4, 2.4)}&\cellcolor{gray!25}(-2.7, 2.6)&(-3.8, 3.4)&\cellcolor{gray!25}(-3.9, 7.1)\\ \cline{2-9}
\end{tabular*} \caption{Type: TvT. Best Estimator by lowest MSE and respective bounds.
Ranges under 5\% are in bold.}
\label{table:tvt-best-bounds}
\end{minipage}

\begin{minipage}{0.98\linewidth}

\centering
\begin{tabular*}{\linewidth}{@{\extracolsep{\fill}}ccccccccc}\\
&&\multicolumn{7}{c}{Number of Features}\\\cline{3-9}
&&2&4&6&8&10&12&14\\ \cline{2-9}
\parbox[t]{3mm}{\multirow{10}{*}{\rotatebox[origin=c]{90}{Number of Samples Per Class}}}&\multirow{2}{*}{500}&\cellcolor{gray!25}$\mbox{kNN}_{H}$&$\mbox{kNN}_{H}$&\cellcolor{gray!25}$\mbox{kNN}_{H}$&$\mbox{kNN}_{M}$&\cellcolor{gray!25}$\mbox{GC}$&$\mbox{GC}$&\cellcolor{gray!25}$\mbox{GC}$\\
&&\cellcolor{gray!25}(-2.3, 3.1)&(-3.3, 2.3)&\cellcolor{gray!25}(-4.3, 2.1)&(-5.2, 5.6)&\cellcolor{gray!25}(-8.2, 3.6)&(-8.7, 4.7)&\cellcolor{gray!25}(-9.3, 5.7)\\ \cline{2-9}
&\multirow{2}{*}{1000}&$\bm{\mbox{kNN}_{H}}$&\cellcolor{gray!25}$\bm{\mbox{kNN}_{H}}$&$\bm{\mbox{kNN}_{H}}$&\cellcolor{gray!25}$\mbox{kNN}_{M}$&$\mbox{GHP}_{L}$&\cellcolor{gray!25}$\mbox{GHP}_{L}$&$\mbox{GC}$\\
&&\textbf{(-1.8, 2.1)}&\cellcolor{gray!25}\textbf{(-2.4, 1.6)}&\textbf{(-3.0, 1.3)}&\cellcolor{gray!25}(-3.8, 4.5)&(-5.3, 4.7)&\cellcolor{gray!25}(-5.5, 4.6)&(-7.9, 3.0)\\ \cline{2-9}
&\multirow{2}{*}{1500}&\cellcolor{gray!25}$\bm{\mbox{kNN}_{H}}$&$\bm{\mbox{kNN}_{H}}$&\cellcolor{gray!25}$\bm{\mbox{kNN}_{H}}$&$\mbox{kNN}_{M}$&\cellcolor{gray!25}$\mbox{GHP}_{L}$&$\mbox{GHP}_{L}$&\cellcolor{gray!25}$\mbox{GHP}_{L}$\\
&&\cellcolor{gray!25}\textbf{(-1.5, 1.7)}&\textbf{(-2.0, 1.3)}&\cellcolor{gray!25}\textbf{(-2.7, 1.0)}&(-3.1, 3.3)&\cellcolor{gray!25}(-4.0, 3.7)&(-4.9, 3.7)&\cellcolor{gray!25}(-5.9, 3.8)\\ \cline{2-9}
&\multirow{2}{*}{2000}&$\bm{\mbox{kNN}_{H}}$&\cellcolor{gray!25}$\bm{\mbox{kNN}_{H}}$&$\bm{\mbox{kNN}_{H}}$&\cellcolor{gray!25}$\mbox{kNN}_{M}$&$\mbox{GHP}_{L}$&\cellcolor{gray!25}$\mbox{GHP}_{L}$&$\mbox{GHP}_{L}$\\
&&\textbf{(-1.4, 1.4)}&\cellcolor{gray!25}\textbf{(-1.9, 1.1)}&\textbf{(-2.4, 0.9)}&\cellcolor{gray!25}(-2.9, 3.1)&(-3.9, 3.4)&\cellcolor{gray!25}(-4.4, 3.5)&(-5.1, 3.2)\\ \cline{2-9}
&\multirow{2}{*}{2500}&\cellcolor{gray!25}$\bm{\mbox{kNN}_{H}}$&$\bm{\mbox{kNN}_{H}}$&\cellcolor{gray!25}$\bm{\mbox{kNN}_{H}}$&$\mbox{kNN}_{M}$&\cellcolor{gray!25}$\mbox{GHP}_{L}$&$\mbox{GHP}_{L}$&\cellcolor{gray!25}$\mbox{GHP}_{L}$\\
&&\cellcolor{gray!25}\textbf{(-1.4, 1.2)}&\textbf{(-1.7, 1.0)}&\cellcolor{gray!25}\textbf{(-2.2, 0.8)}&(-2.5, 2.6)&\cellcolor{gray!25}(-3.0, 3.3)&(-4.1, 3.2)&\cellcolor{gray!25}(-4.9, 2.8)\\ \cline{2-9}
\end{tabular*} 
\caption{Type: TvS. Best Estimator by lowest MSE and respective bounds.
Ranges under 5\% are in bold.}
\label{table:tvs-best-bounds}
\end{minipage}

\begin{minipage}{0.98\linewidth}

\centering

\begin{tabular*}{\linewidth}{@{\extracolsep{\fill}}ccccccccc}\\
&&\multicolumn{7}{c}{Number of Features}\\\cline{3-9}
&&2&3&4&6&8&10&12\\ \cline{2-9}
\parbox[t]{3mm}{\multirow{10}{*}{\rotatebox[origin=c]{90}{Number of Samples Per Class}}}&\multirow{2}{*}{500}&\cellcolor{gray!25}$\mbox{kNN}_{H}$&$\mbox{kNN}_{M}$&\cellcolor{gray!25}$\mbox{GHP}_{L}$&$\mbox{kNN}_{L}$&\cellcolor{gray!25}$\mbox{kNN}_{L}$&$\mbox{kNN}_{L}$&\cellcolor{gray!25}$\mbox{kNN}_{L}$\\
&&\cellcolor{gray!25}(-3.7, 2.3)&(-3.6, 5.9)&\cellcolor{gray!25}(-6.4, 6.5)&(-6.4, 11.7)&\cellcolor{gray!25}(-9.1, 10.8)&(-10.9, 10.2)&\cellcolor{gray!25}(-12.8, 10.8)\\ \cline{2-9}
&\multirow{2}{*}{1000}&$\bm{\mbox{kNN}_{H}}$&\cellcolor{gray!25}$\mbox{kNN}_{M}$&$\mbox{GHP}_{L}$&\cellcolor{gray!25}$\mbox{kNN}_{L}$&$\mbox{kNN}_{L}$&\cellcolor{gray!25}$\mbox{kNN}_{L}$&$\mbox{kNN}_{L}$\\
&&\textbf{(-2.4, 1.6)}&\cellcolor{gray!25}(-1.9, 4.5)&(-4.3, 4.4)&\cellcolor{gray!25}(-4.4, 9.7)&(-6.5, 8.9)&\cellcolor{gray!25}(-8.2, 8.5)&(-9.6, 9.2)\\ \cline{2-9}
&\multirow{2}{*}{1500}&\cellcolor{gray!25}$\bm{\mbox{kNN}_{H}}$&$\mbox{kNN}_{M}$&\cellcolor{gray!25}$\mbox{kNN}_{M}$&$\mbox{GHP}_{L}$&\cellcolor{gray!25}$\mbox{kNN}_{L}$&$\mbox{kNN}_{L}$&\cellcolor{gray!25}$\mbox{kNN}_{L}$\\
&&\cellcolor{gray!25}\textbf{(-1.9, 1.3)}&(-1.3, 3.8)&\cellcolor{gray!25}(-2.8, 3.9)&(-5.0, 3.8)&\cellcolor{gray!25}(-5.4, 9.5)&(-7.1, 8.0)&\cellcolor{gray!25}(-8.8, 7.8)\\ \cline{2-9}
&\multirow{2}{*}{2000}&$\bm{\mbox{kNN}_{H}}$&\cellcolor{gray!25}$\bm{\mbox{kNN}_{M}}$&$\mbox{kNN}_{M}$&\cellcolor{gray!25}$\mbox{GHP}_{L}$&$\mbox{kNN}_{L}$&\cellcolor{gray!25}$\mbox{kNN}_{L}$&$\mbox{kNN}_{L}$\\
&&\textbf{(-1.7, 1.2)}&\cellcolor{gray!25}\textbf{(-0.9, 3.2)}&(-2.3, 3.6)&\cellcolor{gray!25}(-4.2, 3.8)&(-4.8, 8.2)&\cellcolor{gray!25}(-6.3, 7.0)&(-7.8, 7.7)\\ \cline{2-9}
&\multirow{2}{*}{2500}&\cellcolor{gray!25}$\bm{\mbox{kNN}_{H}}$&$\bm{\mbox{kNN}_{M}}$&\cellcolor{gray!25}$\bm{\mbox{kNN}_{M}}$&$\mbox{GHP}_{L}$&\cellcolor{gray!25}$\mbox{kNN}_{L}$&$\mbox{kNN}_{L}$&\cellcolor{gray!25}$\mbox{kNN}_{L}$\\
&&\cellcolor{gray!25}\textbf{(-1.6, 1.0)}&\textbf{(-0.7, 3.0)}&\cellcolor{gray!25}\textbf{(-1.8, 3.0)}&(-3.6, 3.3)&\cellcolor{gray!25}(-4.2, 7.0)&(-5.8, 6.3)&\cellcolor{gray!25}(-7.4, 7.3)\\ \cline{2-9}
\end{tabular*} 
\caption{Type: SvS. Best Estimator by lowest MSE and respective bounds.
Ranges under 5\% are in bold.}
\label{table:svs-best-bounds}

\end{minipage}

\end{table*}

\subsection{Non-Linearity} \label{sec:nonlinear}

All of the tested estimators show evidence of non-linear relationship with respect to the ground truth in at least one simulation scenario.
Figures \ref{fig:ex-KNN}, \ref{fig:ex-GHP}, \ref{fig:GKDE} and  \ref{fig:CLAKDE} show how the behavior of the estimator can change with the BER, with some estimators being more linear than others, and how the behavior differs between estimators.
In general, the non-linearity of the relationship tends to increase with the number of features, however, it should be noted that even with two features, there are such cases.

Most evident in Figure \ref{fig:ex-GHP}, both estimator bias and variance can change over the ground truth BER range.
The error of the estimators tends to be quite skewed in such cases.
This skew can be observed in the unbalance of the estimator bounds in Tables \ref{table:gvg-best-bounds}, \ref{table:tvt-best-bounds}, \ref{table:tvs-best-bounds} and \ref{table:svs-best-bounds}.
As seen in Figure \ref{fig:ex-KNN}, with $\mbox{kNN}_L$, significant deviation begins to occur at large BER values.
Note that reducing the BER range from (1\% - 49\%) to, for example, (1\% - 20\%) would reduce the measured error of $\mbox{kNN}_L$.
In these simulations, the entire BER range is weighted equally, which does mean these results do deviate from results previously reported in literature.
 \ifx\templateType\templateIEEE\begin{figure*}[b]
\centering
  \begin{minipage}{\linewidth}
\subcaptionbox{Estimator: GKDE. Number of Features: 2. \\$R^2$ (LEOSS fit): 1.00. $R^2$ (Linear fit): 0.99. MSE: 3.0. \\ Silverman's Factor is used as the bandwidth \cite{wand_kernel_1995}.}
      {\includegraphics[width=0.49\linewidth]{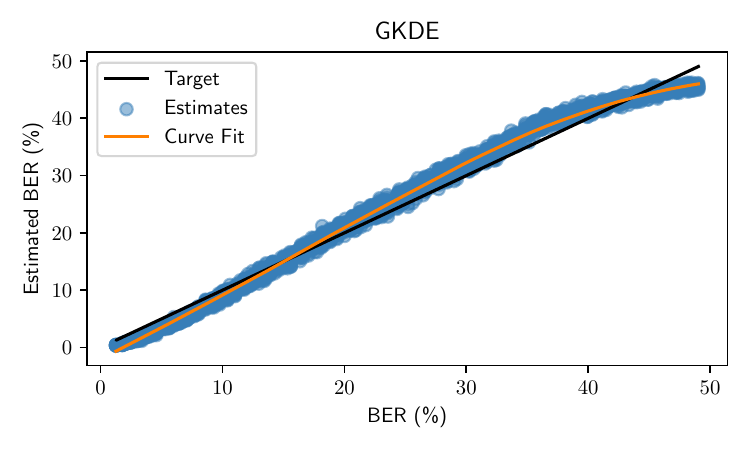}}
      \hfill
    \subcaptionbox{Estimator: GKDE. Number of Features: 10. \\$R^2$ (LEOSS fit): 0.35. $R^2$ (Linear fit): 0.33. MSE: 827.5. \\ Silverman's Factor is used as the bandwidth \cite{wand_kernel_1995}.}
      {\includegraphics[width=0.49\linewidth]{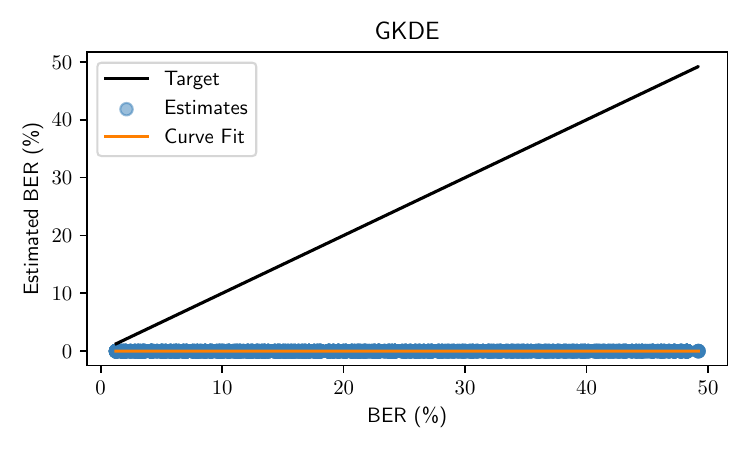}}
      \hfill
    \caption{GKDE example results. Distribution Type: TvT. Samples per Class: 2500.}
    \label{fig:GKDE}
  \end{minipage}\quad
  \ifx\templateType\templateArxiv \vspace{-4mm} \fi
  
  \begin{minipage}{\linewidth}
\subcaptionbox{Estimator: CLAKDE. Number of Features: 2. \\$R^2$ (LEOSS fit): 1.00. $R^2$ (Linear fit): 0.97. MSE: 33.2.}
      {\includegraphics[width=0.49\linewidth]{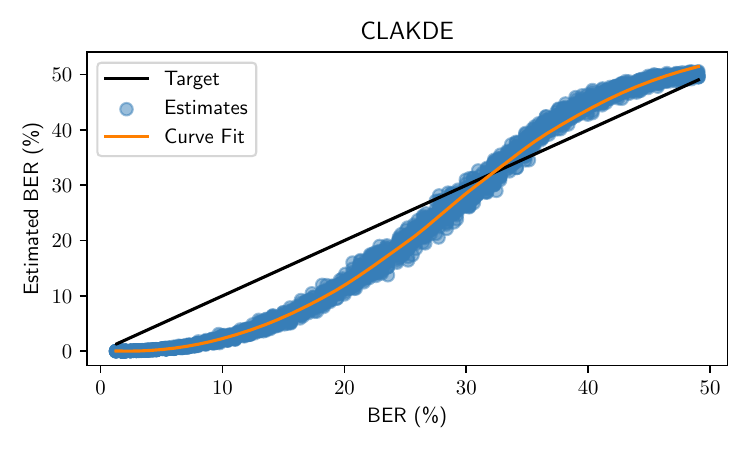}}
      \hfill
    \subcaptionbox{Estimator: CLAKDE. Number of Features: 10. \\$R^2$ (LEOSS fit): 0.99. $R^2$ (Linear fit): 0.97. MSE: 28.5.}
      {\includegraphics[width=0.49\linewidth]{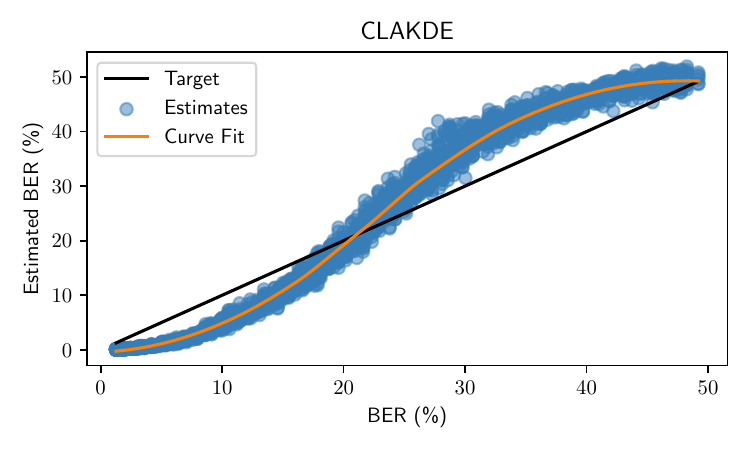}}
      \hfill
    \caption{CLAKDE example results. Distribution Type: TvT. Samples per Class: 2500.
    }
    \label{fig:CLAKDE}
  \end{minipage}\quad
  \ifx\templateType\templateArxiv \vspace{-4mm} \fi

\end{figure*} \fi

\subsection{Test Scenarios} \label{sec:testResults}

The Tables {\ref{table:gvg-best-bounds}}, {\ref{table:tvt-best-bounds}}, {\ref{table:tvs-best-bounds}} and {\ref{table:svs-best-bounds}} report the best estimator (determined by lowest MSE) for each set of simulations.
In some cases, the MSE of multiple estimators are very close, so the selected best estimator can change simply based on how many simulations are run.
The appendix contains additional details, specifically separate tables for the GHP and kNN estimator results, for convience of referencing.

Table \ref{table:gvg-best-bounds}.A shows the best estimator in almost every case is the Naive Bayes classifier error.
There is a minimum requirement of 1000 samples per class to reach the 5\% boundary range (R.{\ref{req:errorBounds}}).
That number of samples is sufficient up to 10 features.
While both GvG and Naive Bayes are included only for comparison purposes, these results do give a more realistic baseline, so the impact of not knowing the distribution can be observed when the Naive Bayes classifier is excluded (Table \ref{table:gvg-best-bounds}.B).

In Table \ref{table:gvg-best-bounds}.B, there is a visible drop in accuracy as the number of features increases and the variations of kNN become the best estimators in all cases.
In this case, 1000 samples per class are only sufficient for up to 4 features.
At 10 features, a minimum of 2500 samples per class is required.

In the testing scenario TvT (see Table \ref{table:tvt-best-bounds}), Naive Bayes does not show up in any case, as expected for that type of problem (see Section \ref{sec:dist}).
Overall, the results between Table \ref{table:gvg-best-bounds}.B and Table \ref{table:tvt-best-bounds} are similar in terms of the best estimators.
kNN is the best estimator in most cases, although GHP does show up as higher numbers of features for TvT.
TvT is more difficult than GvG, and this is reflected in the error bounds, most obvious at the larger bounds for TvT at the most features.

For the spherical testing scenarios (Tables \ref{table:tvs-best-bounds} and \ref{table:svs-best-bounds}), the estimators generally show reduced performance in comparison to the other scenarios (R.{\ref{req:robustToDist}}).
Notably, the dropoff in accuracy is much more abrupt in comparison to TvT and GvG.
While TvT has estimators that meet the criteria at a maximum of 8 features, TvS only goes to 6 features and SvS goes to 4 features.
Overall, additional features make the problem much more difficult, which is sensible, as the complexity of the problem is increasing (R.{\ref{req:features}}).
From Table \ref{table:svs-best-bounds}, it is clear that SvS is the most difficult scenario out of the four.

In TvS (Table \ref{table:tvs-best-bounds}), GC begins to appear as the best estimator at higher numbers of features with less samples.
While this represents an improvement over all existing estimators tested, it only appears in TvS and it is not enough improvement to reach the target 5\% range.

$\mbox{GHP}_L$ continues to appear sporadically at higher numbers of features.
Interestingly, $\mbox{GHP}_L$ is the only GHP estimator that ever appears in these tables, and it often overestimates the BER.
Although $\mbox{GHP}_L$ does often show up as the best estimator, there is only one case where it meets the requirements (see Section \ref{sec:estimators}) as shown in Table \ref{table:tvt-best-bounds} (TvT, 10 features, 2500 samples per class).
It faces a similar problem to GC, the improvements over the other estimators not being sufficient to meet the requirements, which leads to kNN being the best estimator overall.

\ifx\templateType\templateArxiv\FloatBarrier\fi
\ifx\templateType\templateArxiv\begin{figure}[ht]
  \begin{minipage}{\linewidth}
\subcaptionbox{Estimator: CLAKDE. Number of Features: 2. \\$R^2$ (LEOSS fit): 1.00. $R^2$ (Linear fit): 0.97. MSE: 33.2.}
      {\includegraphics[width=0.98\linewidth]{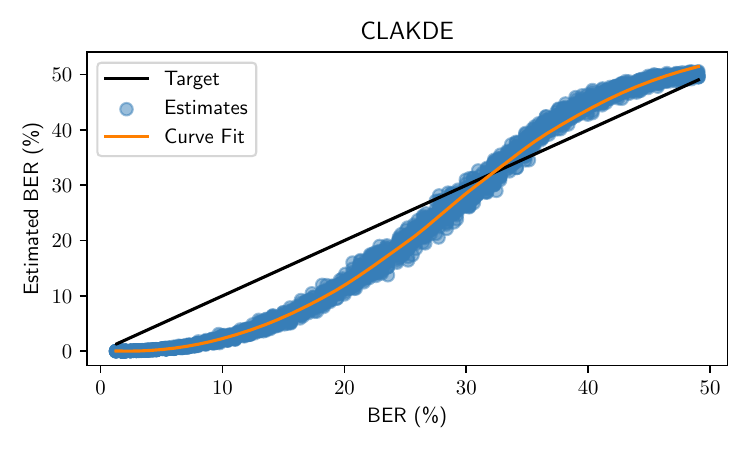}}
      \hfill
    \subcaptionbox{Estimator: CLAKDE. Number of Features: 10. \\$R^2$ (LEOSS fit): 0.99. $R^2$ (Linear fit): 0.97. MSE: 28.5.}
      {\includegraphics[width=0.98\linewidth]{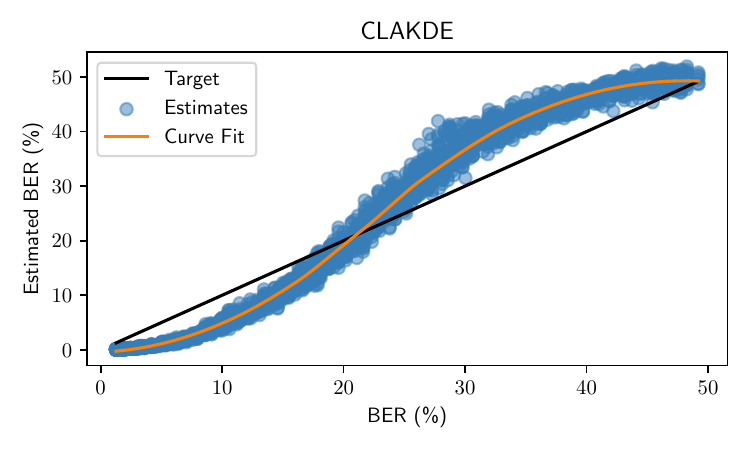}}
      \hfill
    \caption{CLAKDE example results.\\
    Distribution Type: TvT. Samples per Class: 2500.
    }
    \label{fig:CLAKDE}
  \end{minipage}\quad
  \ifx\templateType\templateArxiv \vspace{-4mm} \fi
\end{figure}

\begin{figure}[ht]
  \begin{minipage}{\linewidth}
\subcaptionbox{Estimator: GKDE. Number of Features: 2. \\$R^2$ (LEOSS fit): 1.00. $R^2$ (Linear fit): 0.99. MSE: 3.0.}
      {\includegraphics[width=0.98\linewidth]{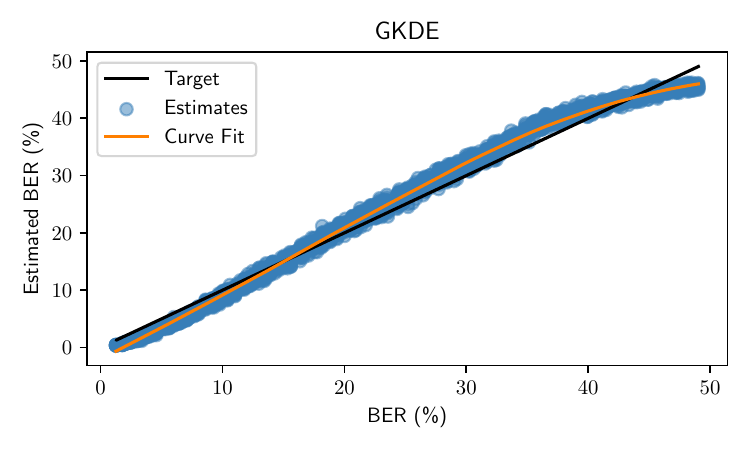}}
      \hfill
    \subcaptionbox{Estimator: GKDE. Number of Features: 10. \\$R^2$ (LEOSS fit): 0.35. $R^2$ (Linear fit): 0.33. MSE: 827.5.}
      {\includegraphics[width=0.98\linewidth]{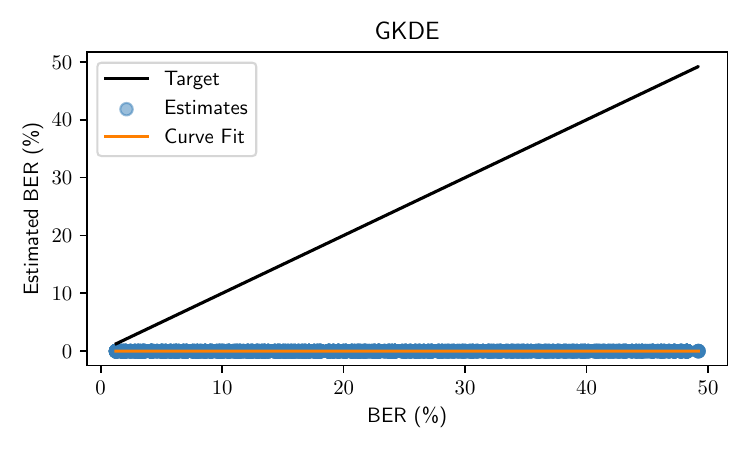}}
      \hfill
    \caption{GKDE example results. Silverman's Factor is used as the bandwidth \cite{wand_kernel_1995}.\\
    Distribution Type: TvT. Samples per Class: 2500.}
    \label{fig:GKDE}
  \end{minipage}\quad
  \ifx\templateType\templateArxiv \vspace{-4mm} \fi
\end{figure} \fi

\subsection{CLAKDE vs. GKDE Comparison}

While CLAKDE is not the best estimator for any of the scenarios, it does show significant improvements over GKDE.
Table \ref{table:gvg-gkde-mse} shows how the MSE of the GKDE estimators increases extremely quickly as more features are added, even in the GvG case.
In contrast, CLAKDE is much more robust when it comes to the number of features (Table \ref{table:gvg-clakde-mse}) (R.{\ref{req:features}}).
GKDE is still a better estimator when using only 2 features, or when a very large number of samples are available, but these are not part of the target application (see Section \ref{sec:estimators}).

As shown in Figure \ref{fig:GKDE}.B, GKDE preforms very poorly in higher dimensional spaces because it tends to underestimate the BER.
CLAKDE successfully solves this particular issue.
However, Figure \ref{fig:CLAKDE}.A also shows the nonlinear relationship of CLAKDE between the estimated and true BER results, which prevents the improvement of the results when more samples are added to the dataset.
Through MSE alone (Table \ref{table:gvg-clakde-mse}), it is difficult to see if the error is coming from a bias (Figure \ref{fig:CLAKDE}.A) or variance (Figure \ref{fig:CLAKDE}.B), which tends to increase with the number of features.

\renewcommand{\arraystretch}{1}
\begin{table*}[t]
\centering

\begin{tabular*}{\linewidth}{@{\extracolsep{\fill}}cccccccc}\\
&&\multicolumn{6}{c}{Number of Features}\\\cline{3-8}
&&2&4&6&8&10&12\\ \cline{2-8}
\parbox[t]{3mm}{\multirow{10}{*}{\rotatebox[origin=c]{90}{Number of Samples Per Class}}}&\multirow{2}{*}{500}&\cellcolor{gray!25}$\mbox{GKDE}_{0.5}$&$\mbox{GKDE}_{0.5}$&\cellcolor{gray!25}$\mbox{GKDE}_{0.5}$&$\mbox{GKDE}_{0.0025}$&\cellcolor{gray!25}$\mbox{GKDE}_{0.0025}$&$\mbox{GKDE}_{0.0025}$\\
&&\cellcolor{gray!25}3.7&397.2&\cellcolor{gray!25}805.1&796.1&\cellcolor{gray!25}810.9&804.5\\ \cline{2-8}
&\multirow{2}{*}{1000}&$\mbox{GKDE}_{0.5}$&\cellcolor{gray!25}$\mbox{GKDE}_{0.5}$&$\mbox{GKDE}_{0.5}$&\cellcolor{gray!25}$\mbox{GKDE}_{0.0025}$&$\mbox{GKDE}_{0.0025}$&\cellcolor{gray!25}$\mbox{GKDE}_{0.0025}$\\
&&5.1&\cellcolor{gray!25}225.4&788.5&\cellcolor{gray!25}787.4&799.6&\cellcolor{gray!25}792.4\\ \cline{2-8}
&\multirow{2}{*}{1500}&\cellcolor{gray!25}$\mbox{GKDE}_{0.25}$&$\mbox{GKDE}_{0.5}$&\cellcolor{gray!25}$\mbox{GKDE}_{0.5}$&$\mbox{GKDE}_{0.0025}$&\cellcolor{gray!25}$\mbox{GKDE}_{0.0025}$&$\mbox{GKDE}_{0.0025}$\\
&&\cellcolor{gray!25}2.9&139.5&\cellcolor{gray!25}772.4&782.4&\cellcolor{gray!25}792.2&785.0\\ \cline{2-8}
&\multirow{2}{*}{2000}&$\mbox{GKDE}_{0.25}$&\cellcolor{gray!25}$\mbox{GKDE}_{0.5}$&$\mbox{GKDE}_{0.5}$&\cellcolor{gray!25}$\mbox{GKDE}_{0.0025}$&$\mbox{GKDE}_{0.0025}$&\cellcolor{gray!25}$\mbox{GKDE}_{0.0025}$\\
&&2.5&\cellcolor{gray!25}91.7&757.0&\cellcolor{gray!25}777.9&786.7&\cellcolor{gray!25}779.5\\ \cline{2-8}
&\multirow{2}{*}{2500}&\cellcolor{gray!25}$\mbox{GKDE}_{0.25}$&$\mbox{GKDE}_{0.5}$&\cellcolor{gray!25}$\mbox{GKDE}_{0.5}$&$\mbox{GKDE}_{0.0025}$&\cellcolor{gray!25}$\mbox{GKDE}_{0.0025}$&$\mbox{GKDE}_{0.0025}$\\
&&\cellcolor{gray!25}2.6&62.8&\cellcolor{gray!25}741.6&775.1&\cellcolor{gray!25}782.6&777.0\\ \cline{2-8}
\end{tabular*} 
\caption{Type: TvT. Best GKDE Estimator by MSE.}
\label{table:gvg-gkde-mse}
\ifx\templateType\templateArxiv \vspace{-4mm} \fi
\end{table*}

\renewcommand{\arraystretch}{1.2}
\begin{table}[t]
\centering \small

\begin{tabular}{@{\extracolsep{\fill}}cccccccc}\\
&&\multicolumn{6}{c}{Number of Features}\\\cline{3-8}
&&2&4&6&8&10&12\\ \cline{2-8}
\parbox[t]{8mm}{\multirow{5}{*}{\rotatebox[origin=c]{90}{\makecell{Number of Samples\\Per Class}}}}&500&\cellcolor{gray!25}21.5&24.0&\cellcolor{gray!25}30.3&41.9&\cellcolor{gray!25}46.0&62.5\\ \cline{2-8}
&1000&25.3&\cellcolor{gray!25}22.3&25.0&\cellcolor{gray!25}31.9&35.1&\cellcolor{gray!25}38.9\\ \cline{2-8}
&1500&\cellcolor{gray!25}28.7&22.5&\cellcolor{gray!25}23.9&29.6&\cellcolor{gray!25}32.0&35.4\\ \cline{2-8}
&2000&31.9&\cellcolor{gray!25}22.6&23.4&\cellcolor{gray!25}28.8&31.8&\cellcolor{gray!25}33.1\\ \cline{2-8}
&2500&\cellcolor{gray!25}33.0&23.4&\cellcolor{gray!25}23.6&28.3&\cellcolor{gray!25}28.6&36.6\\ \cline{2-8}
\end{tabular} 
\caption{Type: TvT. CLAKDE MSE.}
\label{table:gvg-clakde-mse}
\end{table}

\section{Analysis} \label{sec:dis}

\pdfmarkupcomment[color=yellow]{This section will further discuss the behavior and usability of the estimators against the requirements.}{(RVR2.2) (RVR2.3) Reorganized section and broke into different subsections.}

\subsection{Non-Linearity}
There are two major factors which influence the error bounds of our measures: variance and bias.
In Section \ref{sec:nonlinear}, it is shown that some estimators veer away from the true BER value, following a separate curve, increasing their bias.
In this way, a measure can become more precise with more samples, but not more accurate, leading to the error to converge to a non-zero value.
Thus, the error should not be assumed to be Gaussian, and there is an obvious skew in many of the reported error bounds.
This skew is a good hint that there is a nonlinear relationship between the true value and the estimate.

The tendency of an estimator to converge to a function that does not follow the ideal BER is not desirable behavior.
Some estimators have this problem, notably CLAKDE follows a sigmoid-like curve.
In effect, this could likely be added to the list of desirable properties for an estimators, that the error is consistent over the BER range.

\pdfmarkupcomment[color=yellow]{
Unfortunately, this non-linear relationship may not be easily corrected based on empirical results.
The most significant issue is that the estimator curves themselves are not consistent, their shape also changes with the number of samples, number of features and scenario type (as seen in Figure {\ref{fig:CLAKDE}}).
Thus, different corrective models would be required for each scenario, violating {R.\ref{req:robustToDist}}.
Any correction would have to be integrated into the method itself, and thus be required to be just as adaptive, leading back to the original problem.
}{(RVR2.3) Added discussion as to why empirical correction is not  a desirable solution.}

While this \pdfmarkupcomment[color=yellow]{non-linear}{Edited for clarity.} behavior is not often discussed, it does also appear in the results of \cite{sekeh_learning_2020} and \cite{jeong_demystifying_2024}.
Researchers should generally be aware of this behavior when they are picking a limited number of test cases with specific BER values for testing (such as \cite{fukunaga_bayes_1987}, \cite{noshad_learning_2019} and \cite{ishida_is_2023}).

A large BER range (1\% - 49\%) was used to create these simulations, however, the upper end of the BER range may not be relevant to every application, or the accuracy of the measure may be less of a concern.
While the almost entire BER range was used in this case, it may not be necessary to do so.
Consideration should be given to what range of values are of interest or could be expected in specific applications.

As a reminder, a linear selection of parameters for these testing distributions do not result in a linear BER range (which can also be seen in 
{\cite{sekeh_learning_2020}}).
Caution should be taken to avoid skewing the error towards a specific area of the BER range when it is not desired.
In this experiment, resampling was preformed to ensure an even distribution of simulations over the BER range of interest (see Section {\ref{sec:gt}}).

\subsection{Best Estimator}

In GvG (Table \ref{table:gvg-best-bounds}.B) and TvT, the estimator that meets the target 5\% range for the error bounds most often is $\mbox{kNN}_H$.
In all test cases, $\mbox{kNN}_H$ is always best at low numbers of features and, as the numbers of features increases, the best estimator changes.
$\mbox{kNN}_M$ tends to appear in the mid-range (6-10 features) which tends to transition to $\mbox{kNN}_L$ as the number of features increases.

$\mbox{GHP}_L$ continues to appear sporadically at higher numbers of features.
Interestingly, $\mbox{GHP}_L$ is the only GHP estimator that ever appears in these tables, and it often overestimates the BER.
Although $\mbox{GHP}_L$ does often show up as the best estimator, there is only one case where it meets the requirements (Table \ref{table:tvt-best-bounds}: TvT, 10 features, 2500 samples per class).

In TvS (Table \ref{table:tvs-best-bounds}), GC begins to appear as the best estimator at higher numbers of features with less samples.
While this represents an improvement over all existing estimators tested, it only appears in TvS and it is not enough improvement to reach the target 5\% range.
$\mbox{GHP}_L$ faces a similar problem to GC, the improvements over the other estimators not being sufficient to meet the requirements.

\ifx\templateType\templateIEEE\else
\begin{figure}[ht]
\centering
\ifx\templateType\templateArxiv
\includegraphics[width=0.7\linewidth]{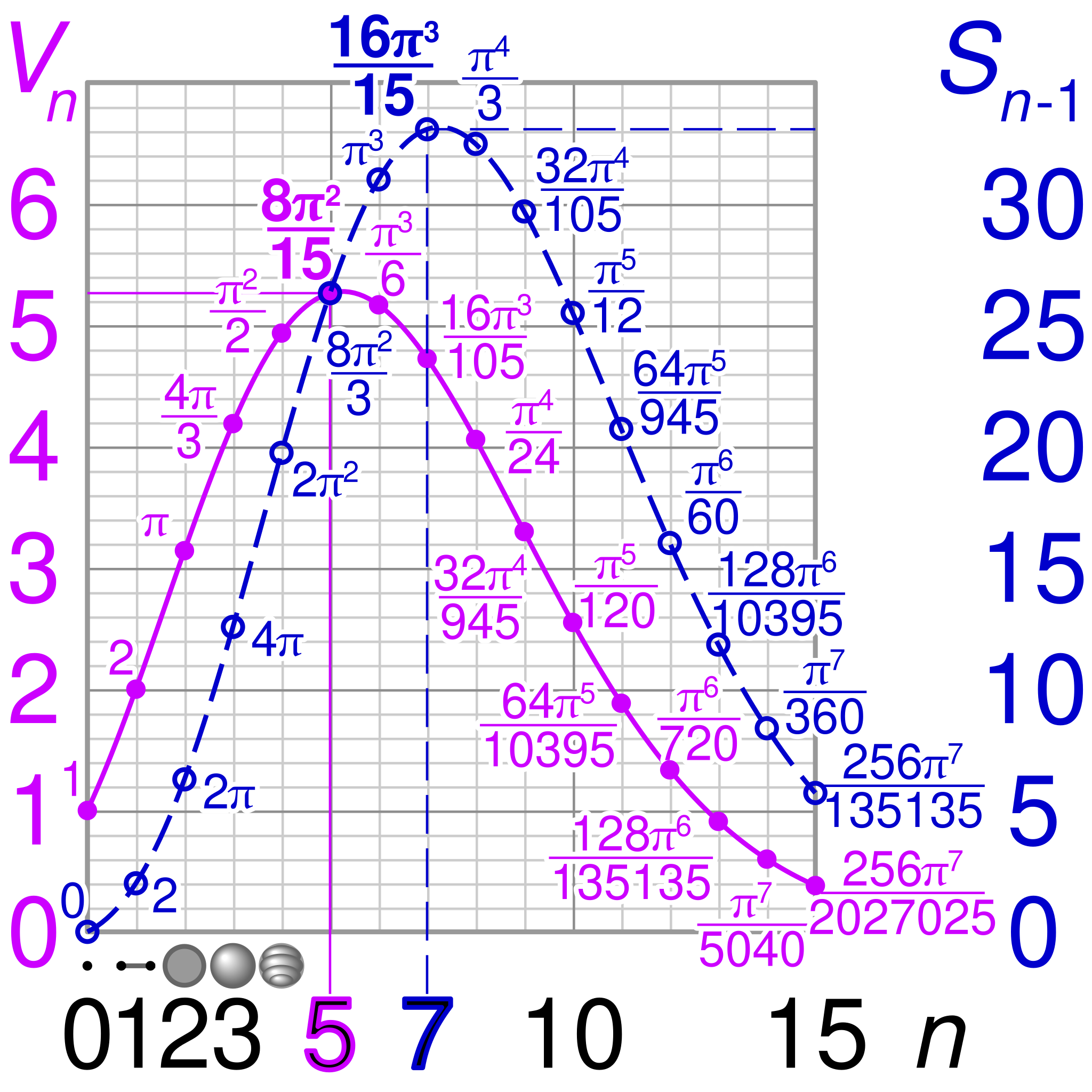}
\else
\includegraphics[width=0.6\linewidth]{Hypersphere_volume_and_surface_area_graphs.png}
\fi
\caption{Volume and surface area of unit n-sphere \cite{lee_hypersphere_2018}.}
\label{fig:nshpere}
\end{figure} \fi

From these results, it is clear that there is no BER estimator that is the best in every scenario.
The decrease in accuracy as the number of features increases is not unexpected, due to the curse of dimensionality.
As the maximum volume of a hypersphere occurs at 5 dimensions and the maximum surface area occurs at 7
\ifx\templateType\templateIEEE
dimensions,
\else
dimensions (see Figure \ref{fig:nshpere}),
\fi
these may indicate possible transition points.

However, the transitions are not the same for every test scenario.
In TvT, the accuracy decrease of the 5\% range takes place from 4 features to 10 features.
TvS experiences a complete drop between 6-8 features.
SvS drops between 2-4 features.
As such, the complexity of the distributions may have a larger influence than the curse of dimensionality.

\subsubsection{k Nearest Neighbour}
\pdfmarkupcomment[color=yellow]{Overall, the various kNN estimators are the top performers.
Further results specific to only kNN are included in Appendix {\ref{sec:AppB}}, which do show changes in performance from scenario to scenario.
Generally, $\mbox{kNN}_H$ is the best with lower numbers of features, but as more features are added, $\mbox{kNN}_M$ then $\mbox{kNN}_L$ become better.
This behaviour should be taken into account during estimator selection ($\mbox{kNN}_H$, $\mbox{kNN}_M$ or $\mbox{kNN}_L$).
}{(RVR2.1) Created section. Added additional details about kNN behavior.}

\pdfmarkupcomment[color=yellow]{
There is still the limitation that a parameter must be selected for the upper limit for $k_{range}$, which controls the number of possible $k$ values for the kNN, which is contrary to R.{\ref{req:param}}.
However, it is possible to set the limit as the number of samples.
Thus, all possible values of $k$ are tested.
However, this option can also significantly increase the computation time without improving performance (as the optimal $k$ values may not lie in the expanded range).
}{(RVR2.1) Added discussion as to reasons why kNN is the best performer.}

\pdfmarkupcomment[color=yellow]{
The additional step of selecting the best (lowest error) value of $k$ from a large range (see Section {\ref{sec:knn}}) was a key step for minimizing the impact of the choice of hyperparameter on the estimator performance, but also made it a more consistent estimator.
This is because allowing variation in $k$ enabled kNN to better adapt to the different simulations, if $k$ had been fixed, the estimator would have not have preformed as well.
Interestingly, selecting the $k$ value that had the lowest error did not result in significant over-fitting and error underestimation, as would typically be expected from increasing hyperparameter optimization.
The tendency of kNN to approach, but not undershoot, the BER turns out to be key trait.
}{(RVR2.1) Added discussion as to reasons why kNN is the best performer.}

\pdfmarkupcomment[color=yellow]{
This method of selecting $k$ (see Section {\ref{sec:knn}}) does have other impacts.
Notably, the estimation of the lower bound for kNN ({$\mbox{kNN}_L$}) does depend directly on the selected value for $k$, thus this variation in the selected $k$ does produce additional noise in {$\mbox{kNN}_L$}.
Practically, it ought to be noted that the implementation in {\cite{renggli_evaluating_2021}} is specifically designed to efficiently test many values of $k$ at once (the distance matrix is only computed once), which is not the case for other kNN implementations.
With other implementations of kNN or much larger datasets, this method may not be computationally feasible within a reasonable time.
}{(RVR2.1) Added further discussion on usability of kNN.}

\subsubsection{Kernel Density Estimation}

\pdfmarkupcomment[color=yellow]{
GKDE preformed poorly in the larger feature spaces, as was expected.
While CLAKDE showed significant improvements in higher dimensions.
Both CLAKDE and GKDE should have had a slight advantage in these scenarios, since the distributions are constructed with Gaussian mixtures and these methods utilize a Gaussian kernel.
}{(RVR2.1) Moved section. Added paragraph on structural advantages by estimator.}
For example, a scenario with {\say{hard edges}} or sudden density {\say{dropoffs}} (such as a uniform cube) should be a much more difficult case for these methods.
Despite this, in these scenarios, neither outperformed other methods.

\pdfmarkupcomment[color=yellow]{
CLAKDE was limited by it's noisy output and nonlinear relationship with the BER.
However, most estimators also suffer from these nonlinear relationships and differing behavior over different test cases.
Based on the results, it can be concluded that KDE-based BER estimation methods may not preform as poorly as originally believed in higher dimensional spaces.
However, this is only an indication that KDE may be able to be adapted for higher dimensions, but further research is required in order to create an estimator that meets all the requirements.
}{(RVR2.2) Added/edited to highlight future applications.}

\subsubsection{Combining Estimators}

\pdfmarkupcomment[color=yellow]{
As to the possibility of multiple estimators improving robustness to large numbers of features, GC does appear as the best estimator in some cases for the smaller sample sizes and large numbers of features (Table {\ref{table:tvt-best-bounds}}).
This may be an indicator utilizing multiple estimators together may be able counter some of the non-linear behavior produced by individual estimators and add the flexibility needed to adapt to different distributions.
}{(RVR2.2) (RVR2.3) Created section. Emphasized connection to ensemble methods.}

\pdfmarkupcomment[color=yellow]{
Outside of GC, there are also indications that other estimators may balance each other out.
For example, for TvT at 20 features and 500 samples per class, the results of $\mbox{GHP}_L$ are (-9.1, 5.7) and $\mbox{kNN}_L$ are (-6.0, 10.5) (from Appendix {\ref{sec:AppB}}).
So while $\mbox{kNN}_L$ tends to underestimate the BER, $\mbox{GHP}_L$ tends to overestimate it.
Cases like this represent possible scenarios where combining estimators may lead to a better overall estimation of the BER.
}{(RVR2.2) (RVR2.3) Added further details for ensemble methods.}

\pdfmarkupcomment[color=yellow]{
It is also worthwhile to remember that the kNN method used here is effectively a selection out of a pool of many NN estimators.
While it is not a combination of estimators, the fact that many estimators are run on the dataset is a key part of it's success.
Therefore, these results suggest that ensemble methods may be needed in order to create a more accurate BER estimator under the conditions seen here, but those types of methods are beyond the scope of this work and a topic for future research.
}{(RVR2.2) Clarified connection to kNN estimators.}

\subsection{Usability}

The primary factor is the size of the dataset.
Effectively, if a minimum number of samples available is too low, no estimators will be accurate enough to be of use.
In cases where that minimum number of samples is reached but the number of features is too high for an accurate BER estimate, the results could be used to predict how much dimensionality reduction would be required to make use of these BER estimators.
The test cases assume that every feature is an important contributor to the overall BER, but that may not be the case in real datasets.
Real data may contain features which are purely noise, which can be removed without influencing the BER.

The test cases included here are better off treated as the minimum dimensionality reduction that may be achieved on a dataset, where removing additional features would reduce the separability of the problem.
In addition, the type of underlying structure in the data is very important to the accuracy of the estimators, so every opportunity should be taken to transform the problem into a simpler one.
Simple does not necessarily mean Gaussian, TvT shows that even a Gaussian mixture is a much easier problem than SvS, so clumps are likely to be preferable over surfaces.

In any case, all of these estimators do require that all of the data follows the same probability distribution.
This is possibly an issue when applied as a black box technique, where there may be underlying distribution shifts in the data.
Domain shifts, where data from different sources that have their own distribution are combined, or any kind of data drift, such as sensor drift, would cause issues.
Unfortunately, such shifts can show up unexpectedly in some engineering applications \cite{wheat_impact_2024}, and would have to be solved before any BER estimator is applied to the data.

\section{Conclusion}

Unfortunately, none of the estimators under consideration met all the objectives.
However, significant improvements were made on the performance on KDE-based techniques in higher dimensional spaces, demonstrating more potential than may have been expected.
On the other hand, kNN, a rather simple technique, unexpectedly outperformed all other techniques on accuracy alone.
It should be noted that it isn't sufficient for an estimator to be the best under a single set of circumstances and there is still a large gap to be crossed to make the \say{best} estimators into \say{usable} estimators.

\subsection{Summary}

\pdfmarkupcomment[color=yellow]{
This section briefly reiterates the key findings.
Taking all scenarios into account, the minimum guidelines for sampling are as follows:
}{Section moved to improve flow.}

\noindent
\begin{minipage}{0.95\linewidth}
\medskip
\begin{itemize}
    \item 2 Features: 1000 samples per class
    \item 3 Features: 2000 samples per class
    \item 4 Features: 2500 samples per class
\end{itemize}
\smallskip
\end{minipage}

\noindent
\begin{minipage}{0.95\linewidth}
\smallskip
\noindent Major takeaways:
\begin{itemize}
    \item An estimator or classifier can report results lower than the BER on a dataset (Section \ref{sec:bayesClassifier}).
    \item The most important factor to accuracy is the number of samples. If there is not a sufficient number of samples, there's no chance of an accurate measure.
    \item The number of features and problem complexity have large effects on accuracy. The data should be reduced to the simplest form possible before applying the estimators.
\end{itemize}
\smallskip
\end{minipage}

\noindent
\begin{minipage}{0.95\linewidth}
\smallskip
\noindent Recommendations for future research:
\begin{itemize}
    \item Use a range of BER values, spanning the range of possible values for each distribution, to test out new estimators.
    \item Use different distributions for multiple test cases, as results may vary by distribution.
    \item The number of Monte Carlo simulations should be large, in thousands, not hundreds, for more consistent results.
\end{itemize}
\smallskip
\end{minipage}

\section{Acknowledgments}
\ifx\templateType\templateArxiv
This work was supported in part by the Canada Research Chairs (CRC) Program under Project CRC-2020-0127 and in part by the Natural Sciences and Engineering Research Council of Canada (NSERC) Ford-Mitacs Alliance under Project ALLRP-590906-23.
 The authors would like to kindly thank 
Winnie Trandinh
for their valuable feedback on drafts of this article.

 \vspace{20mm}
\else
The authors would like to kindly thank 
Winnie Trandinh
for their valuable feedback on drafts of this article.

 \fi

\ifx\templateType\templateArxiv
    \bibliographystyle{IEEEtran.bst}
\else \ifx\templateType\templateIEEE
    \bibliographystyle{IEEEtran.bst}
\else \ifx\templateType\templateElsevier
    \bibliographystyle{elsarticle-num-names}
    \fi
\fi
\fi

\FloatBarrier
\bibliography{bib/pub.bib}

\begin{thebibliography}{10}
\providecommand{\url}[1]{#1}
\csname url@samestyle\endcsname
\providecommand{\newblock}{\relax}
\providecommand{\bibinfo}[2]{#2}
\providecommand{\BIBentrySTDinterwordspacing}{\spaceskip=0pt\relax}
\providecommand{\BIBentryALTinterwordstretchfactor}{4}
\providecommand{\BIBentryALTinterwordspacing}{\spaceskip=\fontdimen2\font plus
\BIBentryALTinterwordstretchfactor\fontdimen3\font minus \fontdimen4\font\relax}
\providecommand{\BIBforeignlanguage}[2]{{%
\expandafter\ifx\csname l@#1\endcsname\relax
\typeout{** WARNING: IEEEtran.bst: No hyphenation pattern has been}%
\typeout{** loaded for the language `#1'. Using the pattern for}%
\typeout{** the default language instead.}%
\else
\language=\csname l@#1\endcsname
\fi
#2}}
\providecommand{\BIBdecl}{\relax}
\BIBdecl

\bibitem{tumer_estimating_1996}
\BIBentryALTinterwordspacing
K.~Tumer and J.~Ghosh, ``Estimating the {Bayes} error rate through classifier combining,'' in \emph{Proceedings of 13th {International} {Conference} on {Pattern} {Recognition}}, vol.~2, Aug. 1996, pp. 695--699 vol.2, iSSN: 1051-4651. [Online]. Available: \url{https://ieeexplore.ieee.org/document/546912}
\BIBentrySTDinterwordspacing

\bibitem{sekeh_learning_2020}
\BIBentryALTinterwordspacing
S.~Y. Sekeh, B.~Oselio, and A.~O. Hero, ``\BIBforeignlanguage{en}{Learning to {Bound} the {Multi}-{Class} {Bayes} {Error}},'' \emph{\BIBforeignlanguage{en}{IEEE Transactions on Signal Processing}}, vol.~68, pp. 3793--3807, 2020. [Online]. Available: \url{https://ieeexplore.ieee.org/document/9093984/}
\BIBentrySTDinterwordspacing

\bibitem{berisha_empirically_2016}
\BIBentryALTinterwordspacing
V.~Berisha, A.~Wisler, A.~O. Hero, and A.~Spanias, ``\BIBforeignlanguage{en}{Empirically {Estimable} {Classification} {Bounds} {Based} on a {Nonparametric} {Divergence} {Measure}},'' \emph{\BIBforeignlanguage{en}{IEEE Transactions on Signal Processing}}, vol.~64, no.~3, pp. 580--591, Feb. 2016. [Online]. Available: \url{http://ieeexplore.ieee.org/document/7254229/}
\BIBentrySTDinterwordspacing

\bibitem{renggli_evaluating_2021}
\BIBentryALTinterwordspacing
C.~Renggli, L.~Rimanic, N.~Hollenstein, and C.~Zhang, ``\BIBforeignlanguage{en}{Evaluating {Bayes} {Error} {Estimators} on {Real}-{World} {Datasets} with {FeeBee}},'' \emph{\BIBforeignlanguage{en}{Proceedings of the Neural Information Processing Systems Track on Datasets and Benchmarks}}, vol.~1, Dec. 2021. [Online]. Available: \url{https://datasets-benchmarks-proceedings.neurips.cc/paper/2021/hash/045117b0e0a11a242b9765e79cbf113f-Abstract-round2.html}
\BIBentrySTDinterwordspacing

\bibitem{fukunaga_bayes_1987}
\BIBentryALTinterwordspacing
K.~Fukunaga and D.~M. Hummels, ``\BIBforeignlanguage{en}{Bayes {Error} {Estimation} {Using} {Parzen} and k-{NN} {Procedures}},'' \emph{\BIBforeignlanguage{en}{IEEE Transactions on Pattern Analysis and Machine Intelligence}}, vol. PAMI-9, no.~5, pp. 634--643, Sep. 1987. [Online]. Available: \url{https://ieeexplore.ieee.org/document/4767958}
\BIBentrySTDinterwordspacing

\bibitem{theisen_evaluating_2021}
\BIBentryALTinterwordspacing
R.~Theisen, H.~Wang, L.~R. Varshney, C.~Xiong, and R.~Socher, ``Evaluating {State}-of-the-{Art} {Classification} {Models} {Against} {Bayes} {Optimality},'' in \emph{Advances in {Neural} {Information} {Processing} {Systems}}, vol.~34.\hskip 1em plus 0.5em minus 0.4em\relax Curran Associates, Inc., 2021, pp. 9367--9377. [Online]. Available: \url{https://proceedings.neurips.cc/paper/2021/hash/4e0ccd2b894f717df5ebc12f4282ee70-Abstract.html}
\BIBentrySTDinterwordspacing

\bibitem{noshad_learning_2019}
\BIBentryALTinterwordspacing
M.~Noshad, L.~Xu, and A.~Hero, ``\BIBforeignlanguage{en}{Learning to {Benchmark}: {Determining} {Best} {Achievable} {Misclassification} {Error} from {Training} {Data}},'' Sep. 2019, arXiv:1909.07192 [cs, stat]. [Online]. Available: \url{http://arxiv.org/abs/1909.07192}
\BIBentrySTDinterwordspacing

\bibitem{chen_evaluating_2023}
\BIBentryALTinterwordspacing
Q.~Chen, F.~Cao, Y.~Xing, and J.~Liang, ``\BIBforeignlanguage{en}{Evaluating {Classification} {Model} {Against} {Bayes} {Error} {Rate}},'' \emph{\BIBforeignlanguage{en}{IEEE Transactions on Pattern Analysis and Machine Intelligence}}, vol.~45, no.~8, pp. 9639--9653, Aug. 2023. [Online]. Available: \url{https://ieeexplore.ieee.org/document/10027467/}
\BIBentrySTDinterwordspacing

\bibitem{jeong_demystifying_2024}
\BIBentryALTinterwordspacing
M.~Jeong, M.~Cardone, and A.~Dytso, ``Demystifying the optimal performance of multi-class classification,'' in \emph{Proceedings of the 37th {International} {Conference} on {Neural} {Information} {Processing} {Systems}}, ser. {NIPS} '23.\hskip 1em plus 0.5em minus 0.4em\relax Red Hook, NY, USA: Curran Associates Inc., May 2024, pp. 31\,638--31\,664. [Online]. Available: \url{https://proceedings.neurips.cc/paper_files/paper/2023/hash/647e122fc406573c51276692f20379b5-Abstract-Conference.html}
\BIBentrySTDinterwordspacing

\bibitem{fukunaga_error_1990}
K.~Fukunaga, ``\BIBforeignlanguage{eng}{Error {Probability} in {Hypothesis} {Testing}},'' in \emph{\BIBforeignlanguage{eng}{Introduction to {Statistical} {Pattern} {Recognition}}}, 2nd~ed., ser. Computer science and scientific computing.\hskip 1em plus 0.5em minus 0.4em\relax San Diego [u.a.]: Acad. Press, 1990, pp. 85--94.

\bibitem{wheat_impact_2024}
\BIBentryALTinterwordspacing
L.~Wheat, M.~V. Mohrenschildt, S.~Habibi, and D.~Al-Ani, ``Impact of {Data} {Leakage} in {Vibration} {Signals} {Used} for {Bearing} {Fault} {Diagnosis},'' \emph{IEEE Access}, vol.~12, pp. 169\,879--169\,895, 2024. [Online]. Available: \url{https://ieeexplore.ieee.org/document/10752530/}
\BIBentrySTDinterwordspacing

\bibitem{devijver_multiclass_1985}
\BIBentryALTinterwordspacing
P.~A. Devijver, ``\BIBforeignlanguage{en}{A multiclass, k-{NN} approach to {Bayes} risk estimation},'' \emph{\BIBforeignlanguage{en}{Pattern Recognition Letters}}, vol.~3, no.~1, pp. 1--6, Jan. 1985. [Online]. Available: \url{https://linkinghub.elsevier.com/retrieve/pii/0167865585900352}
\BIBentrySTDinterwordspacing

\bibitem{cover_nearest_1967}
\BIBentryALTinterwordspacing
T.~Cover and P.~Hart, ``\BIBforeignlanguage{en}{Nearest neighbor pattern classification},'' \emph{\BIBforeignlanguage{en}{IEEE Transactions on Information Theory}}, vol.~13, no.~1, pp. 21--27, Jan. 1967. [Online]. Available: \url{http://ieeexplore.ieee.org/document/1053964/}
\BIBentrySTDinterwordspacing

\bibitem{devroye_asymptotic_1981}
\BIBentryALTinterwordspacing
L.~Devroye, ``On the {Asymptotic} {Probability} of {Error} in {Nonparametric} {Discrimination},'' \emph{The Annals of Statistics}, vol.~9, no.~6, pp. 1320--1327, 1981, publisher: Institute of Mathematical Statistics. [Online]. Available: \url{https://www.jstor.org/stable/2240421}
\BIBentrySTDinterwordspacing

\bibitem{samworth_optimal_2012}
\BIBentryALTinterwordspacing
R.~J. Samworth, ``\BIBforeignlanguage{en}{Optimal weighted nearest neighbour classifiers},'' \emph{\BIBforeignlanguage{en}{The Annals of Statistics}}, vol.~40, no.~5, pp. 2733--2763, Oct. 2012. [Online]. Available: \url{https://projecteuclid.org/journals/annals-of-statistics/volume-40/issue-5/Optimal-weighted-nearest-neighbour-classifiers/10.1214/12-AOS1049.full}
\BIBentrySTDinterwordspacing

\bibitem{wand_kernel_1995}
\BIBentryALTinterwordspacing
M.~P. Wand and M.~C. Jones, \emph{\BIBforeignlanguage{en}{Kernel {Smoothing}}}.\hskip 1em plus 0.5em minus 0.4em\relax FL, USA: Chapman \& Hall/CRC, 1995. [Online]. Available: \url{https://doi.org/10.1201/b14876}
\BIBentrySTDinterwordspacing

\bibitem{pedregosa_scikit-learn_2011}
\BIBentryALTinterwordspacing
F.~Pedregosa, G.~Varoquaux, A.~Gramfort, V.~Michel, B.~Thirion, O.~Grisel, M.~Blondel, P.~Prettenhofer, R.~Weiss, V.~Dubourg, J.~Vanderplas, A.~Passos, D.~Cournapeau, M.~Brucher, M.~Perrot, and {\'E}.~Duchesnay, ``Scikit-learn: {Machine} {Learning} in {Python},'' \emph{Journal of Machine Learning Research}, vol.~12, no.~85, pp. 2825--2830, 2011. [Online]. Available: \url{http://jmlr.org/papers/v12/pedregosa11a.html}
\BIBentrySTDinterwordspacing

\bibitem{Toews_kernel_2007}
\BIBentryALTinterwordspacing
M.~W. Toews, ``Kernel density.svg,'' Jul. 2007. [Online]. Available: \url{https://en.wikipedia.org/wiki/File:Kernel_density.svg}
\BIBentrySTDinterwordspacing

\bibitem{scott_feasibility_1991}
\BIBentryALTinterwordspacing
D.~W. Scott, ``\BIBforeignlanguage{en}{Feasibility of multivariate density estimates},'' \emph{\BIBforeignlanguage{en}{Biometrika}}, vol.~78, no.~1, pp. 197--205, 1991. [Online]. Available: \url{https://academic.oup.com/biomet/article-lookup/doi/10.1093/biomet/78.1.197}
\BIBentrySTDinterwordspacing

\bibitem{olsen_think_2024}
\BIBentryALTinterwordspacing
K.~Olsen, R.~M.~H. Lindrup, and M.~M{\o}rup, ``\BIBforeignlanguage{en}{Think {Global}, {Adapt} {Local}: {Learning} {Locally} {Adaptive} {K}-{Nearest} {Neighbor} {Kernel} {Density} {Estimators}},'' in \emph{\BIBforeignlanguage{en}{Proceedings of {The} 27th {International} {Conference} on {Artificial} {Intelligence} and {Statistics}}}.\hskip 1em plus 0.5em minus 0.4em\relax PMLR, Apr. 2024, pp. 4114--4122, iSSN: 2640-3498. [Online]. Available: \url{https://proceedings.mlr.press/v238/olsen24a.html}
\BIBentrySTDinterwordspacing

\bibitem{virtanen_scipy_2020}
\BIBentryALTinterwordspacing
P.~Virtanen, R.~Gommers, T.~E. Oliphant, M.~Haberland, T.~Reddy, D.~Cournapeau, E.~Burovski, P.~Peterson, W.~Weckesser, J.~Bright, S.~J. van~der Walt, M.~Brett, J.~Wilson, K.~J. Millman, N.~Mayorov, A.~R.~J. Nelson, E.~Jones, R.~Kern, E.~Larson, C.~J. Carey, {\.I}.~Polat, Y.~Feng, E.~W. Moore, J.~VanderPlas, D.~Laxalde, J.~Perktold, R.~Cimrman, I.~Henriksen, E.~A. Quintero, C.~R. Harris, A.~M. Archibald, A.~H. Ribeiro, F.~Pedregosa, and P.~van Mulbregt, ``\BIBforeignlanguage{en}{{SciPy} 1.0: fundamental algorithms for scientific computing in {Python}},'' \emph{\BIBforeignlanguage{en}{Nature Methods}}, vol.~17, no.~3, pp. 261--272, Mar. 2020, publisher: Nature Publishing Group. [Online]. Available: \url{https://www.nature.com/articles/s41592-019-0686-2}
\BIBentrySTDinterwordspacing

\bibitem{harris_array_2020}
\BIBentryALTinterwordspacing
C.~R. Harris, K.~J. Millman, S.~J. Van Der~Walt, R.~Gommers, P.~Virtanen, D.~Cournapeau, E.~Wieser, J.~Taylor, S.~Berg, N.~J. Smith, R.~Kern, M.~Picus, S.~Hoyer, M.~H. Van~Kerkwijk, M.~Brett, A.~Haldane, J.~F. Del~R{\'i}o, M.~Wiebe, P.~Peterson, P.~G{\'e}rard-Marchant, K.~Sheppard, T.~Reddy, W.~Weckesser, H.~Abbasi, C.~Gohlke, and T.~E. Oliphant, ``\BIBforeignlanguage{en}{Array programming with {NumPy}},'' \emph{\BIBforeignlanguage{en}{Nature}}, vol. 585, no. 7825, pp. 357--362, Sep. 2020. [Online]. Available: \url{https://www.nature.com/articles/s41586-020-2649-2}
\BIBentrySTDinterwordspacing

\bibitem{hunter_matplotlib_2007}
\BIBentryALTinterwordspacing
J.~D. Hunter, ``Matplotlib: {A} {2D} {Graphics} {Environment},'' \emph{Computing in Science \& Engineering}, vol.~9, no.~3, pp. 90--95, May 2007, conference Name: Computing in Science \& Engineering. [Online]. Available: \url{https://ieeexplore.ieee.org/document/4160265}
\BIBentrySTDinterwordspacing

\bibitem{paszke_pytorch_2019}
\BIBentryALTinterwordspacing
A.~Paszke, S.~Gross, F.~Massa, A.~Lerer, J.~Bradbury, G.~Chanan, T.~Killeen, Z.~Lin, N.~Gimelshein, L.~Antiga, A.~Desmaison, A.~K{\"o}pf, E.~Yang, Z.~DeVito, M.~Raison, A.~Tejani, S.~Chilamkurthy, B.~Steiner, L.~Fang, J.~Bai, and S.~Chintala, ``{PyTorch}: an imperative style, high-performance deep learning library,'' in \emph{Proceedings of the 33rd {International} {Conference} on {Neural} {Information} {Processing} {Systems}}.\hskip 1em plus 0.5em minus 0.4em\relax Red Hook, NY, USA: Curran Associates Inc., Dec. 2019, no. 721, pp. 8026--8037. [Online]. Available: \url{https://dl.acm.org/doi/10.5555/3454287.3455008}
\BIBentrySTDinterwordspacing

\bibitem{the_mpmath_development_team_mpmath_2023}
\BIBentryALTinterwordspacing
{The mpmath development team}, ``mpmath: a {Python} library for arbitrary-precision floating-point arithmetic (version 1.3.0),'' 2023. [Online]. Available: \url{http://mpmath.org}
\BIBentrySTDinterwordspacing

\bibitem{seabold_statsmodels_2010}
\BIBentryALTinterwordspacing
S.~Seabold and J.~Perktold, ``\BIBforeignlanguage{en}{Statsmodels: {Econometric} and {Statistical} {Modeling} with {Python}},'' \emph{\BIBforeignlanguage{en}{scipy}}, May 2010. [Online]. Available: \url{https://proceedings.scipy.org/articles/Majora-92bf1922-011}
\BIBentrySTDinterwordspacing

\bibitem{mckinney_data_2010}
\BIBentryALTinterwordspacing
W.~McKinney, ``\BIBforeignlanguage{en}{Data {Structures} for {Statistical} {Computing} in {Python}},'' \emph{\BIBforeignlanguage{en}{scipy}}, May 2010. [Online]. Available: \url{https://proceedings.scipy.org/articles/Majora-92bf1922-00a}
\BIBentrySTDinterwordspacing

\bibitem{ishida_is_2023}
\BIBentryALTinterwordspacing
T.~Ishida, I.~Yamane, N.~Charoenphakdee, G.~Niu, and M.~Sugiyama, ``\BIBforeignlanguage{en}{Is {The} {Performance} {Of} {My} {Deep} {Network} {Too} {Good} {To} {Be} {True}? {A} {Direct} {Approach} {To} {Estimating} {The} {Bayes} {Error} {In} {Binary} {Classification}},'' in \emph{\BIBforeignlanguage{en}{The {Eleventh} {International} {Conference} on {Learning} {Representations}}}.\hskip 1em plus 0.5em minus 0.4em\relax Kigali Rwanda: ICLR, Mar. 2023. [Online]. Available: \url{https://arxiv.org/abs/2202.00395}
\BIBentrySTDinterwordspacing

\bibitem{lee_hypersphere_2018}
\BIBentryALTinterwordspacing
C.~Lee, ``Hypersphere volume and surface area graphs.svg,'' Feb. 2018. [Online]. Available: \url{https://en.wikipedia.org/wiki/File:Hypersphere_volume_and_surface_area_graphs.svg}
\BIBentrySTDinterwordspacing

\bibitem{ishida_binary_2018}
\BIBentryALTinterwordspacing
T.~Ishida, G.~Niu, and M.~Sugiyama, ``Binary {Classification} from {Positive}-{Confidence} {Data},'' in \emph{Advances in {Neural} {Information} {Processing} {Systems}}, vol.~31.\hskip 1em plus 0.5em minus 0.4em\relax Curran Associates, Inc., 2018. [Online]. Available: \url{https://proceedings.neurips.cc/paper_files/paper/2018/hash/bd1354624fbae3b2149878941c60df99-Abstract.html}
\BIBentrySTDinterwordspacing

\end{thebibliography}

\ifx\templateType\templateArxiv
    \appendices
\else \ifx\templateType\templateIEEE
    \begin{IEEEbiography}[{\includegraphics[width=1in,height=1.25in,clip,keepaspectratio]{a1.png}}]{Lesley Wheat} received the B. Eng. in mechatronics engineering from McMaster University in 2020. Currently, she is a graduate student pursuing her Ph.D. in software engineering at McMaster University. Her research interests include fault detection and diagnosis, machine learning, data complexity measures and embedded sensor systems.

\end{IEEEbiography}

\begin{IEEEbiography}[{\includegraphics[width=1in,height=1.25in,clip,keepaspectratio]{a2.png}}]{Martin v. Mohrenschildt} received the Ph.D. degree in mathematics from the ETH-Zürich, Zurich, Switzerland, in 1994. He is a Faculty Member with McMaster University, Hamilton, ON, Canada, initially in Electrical and Computer Engineering and then in Computing and Software. From 2005 to 2011, he was the Chair of the Department. During this time, he implemented the Mechatronics Program with McMaster University. While his core expertise is in mathematics, Dr. Mohrenschildt always had a big passion for electrical and mechanical systems. Over the years, Dr. Mohrenschildt and his students built several software/hardware systems for industrial applications that at their core use signal processing-based feature extraction. For the mining industry he continues to develop a system of sensors that are used commercially to analyze vibrating screens. He also designed and built a minivan sized fully immersive 6 DOF flight/driving simulator to study the effect of motion cuing in learning. His research interests include signal processing and control, sensors and data acquisition, and real time data processing.
\end{IEEEbiography}

\begin{IEEEbiography}[{\includegraphics[width=1in,height=1.25in,clip,keepaspectratio]{a3.png}}]{SAEID HABIBI} (Member, IEEE) received the Ph.D. degree in control engineering from the University of Cambridge, U.K., in 1990. He was the Chair of the Department of Mechanical Engineering, from 2008 to 2013. He is currently the Tier I Canada Research Chair and a Full Professor with the Department of Mechanical Engineering, McMaster University. He is the Founder and the Director of the Centre for Mechatronics and Hybrid Technologies (CMHT). CHMT has created one of the most advanced and best equipped automotive research laboratories in Canada and internationally specializing in predictive algorithms, tracking, prognostics, and diagnostics. CMHT currently supports a number of large projects and being a resource to automotive companies and start-ups. CMHT research is supported by industry and funding from Ontario Research Fund Research Excellence Awards and NSERC. He has published over 200 peerreviewed articles. His academic background includes research into artificial intelligence, intelligent control, state and parameter estimation, fault diagnosis, and prediction. He is a fellow of American Society of Mechanical Engineers (ASME) and Canadian Society of Mechanical Engineers (CSME).
\end{IEEEbiography}

  \EOD \appendices
\else \ifx\templateType\templateElsevier
\fi
\fi
\fi

\FloatBarrier
\clearpage
\section{Additional Cases from Literature} \label{sec:AppA}

See Section \ref{sec:lit_dists} for more information.

\renewcommand{\arraystretch}{1.1}
\begin{table}[h]
\centering
\begin{tabular}{ lllSl} 
\hline

Source & \makecell[l]{Distribution\\Type} & \multicolumn{2}{l}{Parameters} \\ \hline

\multirow{5}{1.5cm}{\cite{ishida_is_2023}, \cite{ishida_binary_2018}} &
\multirow{4}{1.5cm}{Binary Gaussian Distributions} &
\multirow{4}{*}{\begin{tabular}{@{}l@{}}
Number of samples: 10000 \\ Number of features: 2
\end{tabular}} &
\begin{tabular}{@{}l@{}}
$\mu_1$: [0, 0], 
$\sum_1$: 
$\begin{bmatrix}
7, -6\\
-6, 7
\end{bmatrix}$, 
$\mu_2$: [-2, 5], 
$\sum_2$: 
$\begin{bmatrix}
2, 0\\
0, 2
\end{bmatrix}$ \end{tabular} \\ \cline{4-4}

& & &
\begin{tabular}{@{}l@{}}
$\mu_1$: [0, 0], 
$\sum_1$: 
$\begin{bmatrix}
5 , 3\\
3, 5
\end{bmatrix}$, 
$\mu_2$: [0, 4], 
$\sum_2$: 
$\begin{bmatrix}
5, -3\\
-3, 5
\end{bmatrix}$ \end{tabular} \\ \cline{4-4}

& & &
\begin{tabular}{@{}l@{}}
$\mu_1$: [0, 0], 
$\sum_1$: 
$\begin{bmatrix}
7,-6\\
-6,7
\end{bmatrix}$, 
$\mu_2$: [0, 8], 
$\sum_2$: 
$\begin{bmatrix}
7,6\\
6,7
\end{bmatrix}$ \end{tabular} \\ \cline{4-4}

& & &
\begin{tabular}{@{}l@{}}
$\mu_1$: [0, 0], 
$\sum_1$: 
$\begin{bmatrix}
4,0\\
0,4
\end{bmatrix}$, 
$\mu_2$: [0, 4], 
$\sum_2$: 
$\begin{bmatrix}
1,0\\
0,1
\end{bmatrix}$ \end{tabular} \\ \hline

\multirow{5}{1.5cm}{\cite{berisha_empirically_2016}} & \multirow{2}{1.5cm}{Binary Gaussian Distributions} &
\multirow{2}{*}{\begin{tabular}{@{}l@{}}
Number of features: 8 \\
Number of samples: 100, 500, 1000 \\
Number of Monte Carlo runs: 50
\end{tabular}}
&
$\mu_1$: $\vec{0}$, $\sigma_1$: $\vec{1}$, $\mu_2$: $[2.56, 0 ... 0]$, $\sigma_2$: $\vec{1}$
\\ \cline{4-4}

& & &
 \begin{tabular}{@{}l@{}}
$\mu_1$: $\vec{0}$, $\sigma_1$: $\vec{1}$\\
$\mu_2$: $[3.86, 3.10, 0.84, 0.84, 1.64, 1.08, 0.26, 0.01]$\\
$\sigma_2$: $[8.41, 12.06, 0.12, 0.22, 1.49, 1.77, 0.35, 2.73]$
\end{tabular}
\\ \cline{3-4}

& & 
\begin{tabular}{@{}l@{}}
Number of samples: 1000\\
Number of Monte Carlo runs: 150
\end{tabular}
&
\begin{tabular}{@{}l@{}}
$\sigma_1$: $\vec{1}$, $\sigma_2$: $\vec{1}$\\
Spacing: [0, 3]
\end{tabular}
\\ \hline

\multirow{5}{1.5cm}{\cite{tumer_estimating_1996}} &
\multirow{2}{1.5cm}{Binary Gaussian Distributions} &
\multirow{2}{*}{\begin{tabular}{@{}l@{}}
Number of features: 8 \\
Number of samples: 1000
\end{tabular}}
&
$\mu_1$: $\vec{0}$, $\sigma_1$: $\vec{1}$, $\mu_2$: $[2.56, 0 ... 0]$, $\sigma_2$: $\vec{1}$
\\ \cline{4-4}

& & &
 \begin{tabular}{@{}l@{}}
$\mu_1$: $\vec{0}$, $\sigma_1$: $\vec{1}$\\
$\mu_2$: $[3.86, 3.10, 0.84, 0.84, 1.64, 1.08, 0.26, 0.01]$\\
$\sigma_2$: $[8.41, 12.06, 0.12, 0.22, 1.49, 1.77, 0.35, 2.73]$
\end{tabular}
\\ \hline

\multirow{5}{1.5cm}{\cite{fukunaga_bayes_1987}} &
\multirow{3}{1.5cm}{Binary Gaussian Distributions} &
Number of features: 8
&
\begin{tabular}{@{}l@{}}
$\mu_1$: $\vec{0}$, $\sigma_1$: $\vec{1}$, $\mu_2$: $[2.56, 0 ... 0]$, $\sigma_2$: $\vec{1}$\\
Number of samples per class: [25, 400]
\end{tabular}
\\ \cline{3-4}

& & 
\multirow{3}{*}{\begin{tabular}{@{}l@{}}
Number of features: 8 \\
Number of samples per class: 100
\end{tabular}}
&
$\mu_1$: $\vec{0}$, $\sigma_1$: $\vec{4}$, $\mu_2$: $\vec{0}$, $\sigma_2$: $\vec{1}$
\\ \cline{4-4}

& & &
 \begin{tabular}{@{}l@{}}
$\mu_1$: $\vec{0}$, $\sigma_1$: $\vec{1}$\\
$\mu_2$: $[3.86, 3.10, 0.84, 0.84, 1.64, 1.08, 0.26, 0.01]$\\
$\sigma_2$: $[8.41, 12.06, 0.12, 0.22, 1.49, 1.77, 0.35, 2.73]$
\end{tabular}
\\ \cline{2-4}

& \makecell[l]{Binary\\Gaussian\\Mixture}& 
 \begin{tabular}{@{}l@{}}
Number of features: 8\\
Number of samples per class: 100
\end{tabular}
&
 \begin{tabular}{@{}l@{}}
$M_{11}$: $\vec{0}$, $M_{12}$: [6.58, 0, ... 0] \\
$M_{21}$: [3.29, 0, ..., 0] , $M_{22}$: [9.87, 0, ..., 0]
\end{tabular}
\\ \hline

\end{tabular}

\caption{Expanded dataset details from literature.}
\label{table:litRev_ex}

\end{table}

\FloatBarrier
\clearpage
\section{Additional Error Bounds} \label{sec:AppB}

See Section \ref{sec:testResults} for more information.

\ifx\templateType\templateArxiv\renewcommand{\arraystretch}{1}\fi
\ifx\templateType\templateIEEE\renewcommand{\arraystretch}{1.1}\fi
\begin{table}[h]
\centering \small
\ifx\templateType\templateArxiv
    \vspace{-6mm}
\else 
    \vspace{-6mm}
\fi

\begin{tabular*}{\linewidth}{@{\extracolsep{\fill}}cccccccc}\\
&&\multicolumn{6}{c}{Number of Features}\\\cline{3-8}
&&2&3&4&10&20&30\\ \cline{2-8}
\parbox[t]{3mm}{\multirow{10}{*}{\rotatebox[origin=c]{90}{Number of Samples Per Class}}}&\multirow{2}{*}{500}&\cellcolor{gray!25}$\mbox{GHP}_{M}$&$\mbox{GHP}_{M}$&\cellcolor{gray!25}$\mbox{GHP}_{L}$&$\mbox{GHP}_{L}$&\cellcolor{gray!25}$\mbox{GHP}_{L}$&$\mbox{GHP}_{L}$\\
&&\cellcolor{gray!25}(-6.8, 2.2)&(-6.8, 2.1)&\cellcolor{gray!25}(-4.9, 6.6)&(-6.8, 6.4)&\cellcolor{gray!25}(-8.7, 5.5)&(-11.8, 4.9)\\ \cline{2-8}
&\multirow{2}{*}{1000}&$\mbox{GHP}_{M}$&\cellcolor{gray!25}$\mbox{GHP}_{M}$&$\mbox{GHP}_{L}$&\cellcolor{gray!25}$\mbox{GHP}_{L}$&$\mbox{GHP}_{L}$&\cellcolor{gray!25}$\mbox{GHP}_{L}$\\
&&(-5.6, 1.4)&\cellcolor{gray!25}(-5.6, 1.2)&(-4.2, 5.5)&\cellcolor{gray!25}(-5.1, 4.4)&(-6.7, 4.0)&\cellcolor{gray!25}(-9.9, 4.1)\\ \cline{2-8}
&\multirow{2}{*}{1500}&\cellcolor{gray!25}$\mbox{GHP}_{M}$&$\mbox{GHP}_{M}$&\cellcolor{gray!25}$\mbox{GHP}_{L}$&$\mbox{GHP}_{L}$&\cellcolor{gray!25}$\mbox{GHP}_{L}$&$\mbox{GHP}_{L}$\\
&&\cellcolor{gray!25}(-4.7, 0.8)&(-5.1, 0.9)&\cellcolor{gray!25}(-3.2, 5.0)&(-3.9, 3.9)&\cellcolor{gray!25}(-5.9, 3.4)&(-9.3, 3.3)\\ \cline{2-8}
&\multirow{2}{*}{2000}&$\mbox{GHP}_{M}$&\cellcolor{gray!25}$\mbox{GHP}_{M}$&$\mbox{GHP}_{L}$&\cellcolor{gray!25}$\mbox{GHP}_{L}$&$\mbox{GHP}_{L}$&\cellcolor{gray!25}$\mbox{GHP}_{L}$\\
&&(-4.5, 0.7)&\cellcolor{gray!25}(-4.8, 0.5)&(-2.5, 4.7)&\cellcolor{gray!25}(-3.6, 3.6)&(-5.4, 2.9)&\cellcolor{gray!25}(-8.7, 2.8)\\ \cline{2-8}
&\multirow{2}{*}{2500}&\cellcolor{gray!25}$\bm{\mbox{GHP}_{M}}$&$\bm{\mbox{GHP}_{M}}$&\cellcolor{gray!25}$\mbox{GHP}_{L}$&$\mbox{GHP}_{L}$&\cellcolor{gray!25}$\mbox{GHP}_{L}$&$\mbox{GHP}_{L}$\\
&&\cellcolor{gray!25}\textbf{(-4.2, 0.6)}&\textbf{(-4.4, 0.5)}&\cellcolor{gray!25}(-2.7, 4.5)&(-2.8, 3.4)&\cellcolor{gray!25}(-5.0, 3.0)&(-8.7, 2.8)\\ \cline{2-8}
\end{tabular*} \caption{Type: GvG. Best GHP Estimator Bounds.}

\ifx\templateType\templateArxiv
    \vspace{-12mm}
\else 
    \vspace{-10mm}
\fi

\end{table}

\begin{table}[h]
\centering \small
\begin{tabular*}{\linewidth}{@{\extracolsep{\fill}}ccccccccc}\\
&&\multicolumn{7}{c}{Number of Features}\\\cline{3-9}
&&2&4&6&8&10&12&20\\ \cline{2-9}
\parbox[t]{3mm}{\multirow{10}{*}{\rotatebox[origin=c]{90}{Number of Samples Per Class}}}&\multirow{2}{*}{500}&\cellcolor{gray!25}$\mbox{GHP}_{M}$&$\mbox{GHP}_{L}$&\cellcolor{gray!25}$\mbox{GHP}_{L}$&$\mbox{GHP}_{L}$&\cellcolor{gray!25}$\mbox{GHP}_{L}$&$\mbox{GHP}_{L}$&\cellcolor{gray!25}$\mbox{GHP}_{L}$\\
&&\cellcolor{gray!25}(-6.6, 2.1)&(-4.4, 6.4)&\cellcolor{gray!25}(-5.4, 6.6)&(-6.2, 6.0)&\cellcolor{gray!25}(-6.9, 6.2)&(-7.0, 6.0)&\cellcolor{gray!25}(-9.1, 5.7)\\ \cline{2-9}
&\multirow{2}{*}{1000}&$\mbox{GHP}_{M}$&\cellcolor{gray!25}$\mbox{GHP}_{L}$&$\mbox{GHP}_{L}$&\cellcolor{gray!25}$\mbox{GHP}_{L}$&$\mbox{GHP}_{L}$&\cellcolor{gray!25}$\mbox{GHP}_{L}$&$\mbox{GHP}_{L}$\\
&&(-5.3, 1.3)&\cellcolor{gray!25}(-3.6, 5.6)&(-3.9, 5.1)&\cellcolor{gray!25}(-4.4, 4.6)&(-4.6, 4.5)&\cellcolor{gray!25}(-5.5, 4.8)&(-6.7, 4.2)\\ \cline{2-9}
&\multirow{2}{*}{1500}&\cellcolor{gray!25}$\mbox{GHP}_{M}$&$\mbox{GHP}_{L}$&\cellcolor{gray!25}$\mbox{GHP}_{L}$&$\mbox{GHP}_{L}$&\cellcolor{gray!25}$\mbox{GHP}_{L}$&$\mbox{GHP}_{L}$&\cellcolor{gray!25}$\mbox{GHP}_{L}$\\
&&\cellcolor{gray!25}(-4.7, 1.0)&(-3.2, 5.1)&\cellcolor{gray!25}(-3.4, 4.8)&(-3.7, 4.1)&\cellcolor{gray!25}(-3.8, 4.2)&(-4.8, 3.8)&\cellcolor{gray!25}(-5.9, 3.4)\\ \cline{2-9}
&\multirow{2}{*}{2000}&$\mbox{GHP}_{M}$&\cellcolor{gray!25}$\mbox{GHP}_{L}$&$\mbox{GHP}_{L}$&\cellcolor{gray!25}$\mbox{GHP}_{L}$&$\mbox{GHP}_{L}$&\cellcolor{gray!25}$\mbox{GHP}_{L}$&$\mbox{GHP}_{L}$\\
&&(-4.4, 0.7)&\cellcolor{gray!25}(-3.1, 4.8)&(-2.7, 4.2)&\cellcolor{gray!25}(-3.1, 4.0)&(-3.6, 3.5)&\cellcolor{gray!25}(-4.1, 3.7)&(-5.3, 3.1)\\ \cline{2-9}
&\multirow{2}{*}{2500}&\cellcolor{gray!25}$\bm{\mbox{GHP}_{M}}$&$\mbox{GHP}_{M}$&\cellcolor{gray!25}$\mbox{GHP}_{L}$&$\mbox{GHP}_{L}$&\cellcolor{gray!25}$\mbox{GHP}_{L}$&$\mbox{GHP}_{L}$&\cellcolor{gray!25}$\mbox{GHP}_{L}$\\
&&\cellcolor{gray!25}\textbf{(-4.2, 0.6)}&(-4.6, 0.6)&\cellcolor{gray!25}(-2.6, 4.2)&(-2.4, 3.7)&\cellcolor{gray!25}(-3.2, 3.7)&(-3.8, 3.4)&\cellcolor{gray!25}(-5.1, 3.3)\\ \cline{2-9}
\end{tabular*} \caption{Type: TvT. Best GHP Estimator Bounds.}
\ifx\templateType\templateArxiv
    \vspace{-12mm}
\else 
    \vspace{-10mm}
\fi
\end{table}

\begin{table}[h]
\centering \small
\begin{tabular*}{\linewidth}{@{\extracolsep{\fill}}ccccccccc}\\
&&\multicolumn{7}{c}{Number of Features}\\\cline{3-9}
&&2&4&6&8&10&12&14\\ \cline{2-9}
\parbox[t]{3mm}{\multirow{10}{*}{\rotatebox[origin=c]{90}{Number of Samples Per Class}}}&\multirow{2}{*}{500}&\cellcolor{gray!25}$\mbox{GHP}_{M}$&$\mbox{GHP}_{L}$&\cellcolor{gray!25}$\mbox{GHP}_{L}$&$\mbox{GHP}_{L}$&\cellcolor{gray!25}$\mbox{GHP}_{L}$&$\mbox{GHP}_{L}$&\cellcolor{gray!25}$\mbox{GHP}_{L}$\\
&&\cellcolor{gray!25}(-6.1, 2.4)&(-4.9, 6.6)&\cellcolor{gray!25}(-5.4, 6.4)&(-6.9, 6.2)&\cellcolor{gray!25}(-7.0, 6.0)&(-7.7, 6.2)&\cellcolor{gray!25}(-8.5, 5.2)\\ \cline{2-9}
&\multirow{2}{*}{1000}&$\mbox{GHP}_{M}$&\cellcolor{gray!25}$\mbox{GHP}_{M}$&$\mbox{GHP}_{L}$&\cellcolor{gray!25}$\mbox{GHP}_{L}$&$\mbox{GHP}_{L}$&\cellcolor{gray!25}$\mbox{GHP}_{L}$&$\mbox{GHP}_{L}$\\
&&(-4.9, 1.5)&\cellcolor{gray!25}(-5.8, 1.2)&(-4.1, 5.1)&\cellcolor{gray!25}(-4.8, 5.0)&(-5.3, 4.5)&\cellcolor{gray!25}(-5.5, 4.6)&(-6.6, 4.1)\\ \cline{2-9}
&\multirow{2}{*}{1500}&\cellcolor{gray!25}$\mbox{GHP}_{M}$&$\mbox{GHP}_{M}$&\cellcolor{gray!25}$\mbox{GHP}_{L}$&$\mbox{GHP}_{L}$&\cellcolor{gray!25}$\mbox{GHP}_{L}$&$\mbox{GHP}_{L}$&\cellcolor{gray!25}$\mbox{GHP}_{L}$\\
&&\cellcolor{gray!25}(-4.2, 1.1)&(-5.0, 1.1)&\cellcolor{gray!25}(-3.7, 4.6)&(-4.1, 4.0)&\cellcolor{gray!25}(-3.9, 3.7)&(-4.9, 3.6)&\cellcolor{gray!25}(-5.9, 3.8)\\ \cline{2-9}
&\multirow{2}{*}{2000}&$\mbox{GHP}_{M}$&\cellcolor{gray!25}$\mbox{GHP}_{M}$&$\mbox{GHP}_{L}$&\cellcolor{gray!25}$\mbox{GHP}_{L}$&$\mbox{GHP}_{L}$&\cellcolor{gray!25}$\mbox{GHP}_{L}$&$\mbox{GHP}_{L}$\\
&&(-4.1, 1.0)&\cellcolor{gray!25}(-4.7, 0.6)&(-2.8, 4.3)&\cellcolor{gray!25}(-3.3, 3.9)&(-3.8, 3.4)&\cellcolor{gray!25}(-4.4, 3.4)&(-5.1, 3.2)\\ \cline{2-9}
&\multirow{2}{*}{2500}&\cellcolor{gray!25}$\bm{\mbox{GHP}_{M}}$&$\mbox{GHP}_{M}$&\cellcolor{gray!25}$\mbox{GHP}_{L}$&$\mbox{GHP}_{L}$&\cellcolor{gray!25}$\mbox{GHP}_{L}$&$\mbox{GHP}_{L}$&\cellcolor{gray!25}$\mbox{GHP}_{L}$\\
&&\cellcolor{gray!25}\textbf{(-3.8, 0.6)}&(-4.5, 0.6)&\cellcolor{gray!25}(-2.7, 4.2)&(-2.9, 3.7)&\cellcolor{gray!25}(-3.0, 3.3)&(-4.1, 3.1)&\cellcolor{gray!25}(-4.9, 2.8)\\ \cline{2-9}
\end{tabular*} \caption{Type: TvS. Best GHP Estimator Bounds.}
\ifx\templateType\templateArxiv
    \vspace{-12mm}
\else 
    \vspace{-12mm}
\fi
\end{table}

\begin{table}[h]
\centering \small
\begin{tabular*}{\linewidth}{@{\extracolsep{\fill}}ccccccccc}\\
&&\multicolumn{7}{c}{Number of Features}\\\cline{3-9}
&&2&3&4&6&8&10&12\\ \cline{2-9}
\parbox[t]{3mm}{\multirow{10}{*}{\rotatebox[origin=c]{90}{Number of Samples Per Class}}}&\multirow{2}{*}{500}&\cellcolor{gray!25}$\mbox{GHP}_{L}$&$\mbox{GHP}_{L}$&\cellcolor{gray!25}$\mbox{GHP}_{L}$&$\mbox{GHP}_{L}$&\cellcolor{gray!25}$\mbox{GHP}_{L}$&$\mbox{GHP}_{L}$&\cellcolor{gray!25}$\mbox{GHP}_{L}$\\
&&\cellcolor{gray!25}(-4.0, 7.1)&(-5.3, 6.3)&\cellcolor{gray!25}(-6.4, 6.5)&(-9.6, 5.1)&\cellcolor{gray!25}(-12.6, 4.9)&(-14.8, 4.7)&\cellcolor{gray!25}(-16.6, 6.0)\\ \cline{2-9}
&\multirow{2}{*}{1000}&$\mbox{GHP}_{M}$&\cellcolor{gray!25}$\mbox{GHP}_{L}$&$\mbox{GHP}_{L}$&\cellcolor{gray!25}$\mbox{GHP}_{L}$&$\mbox{GHP}_{L}$&\cellcolor{gray!25}$\mbox{GHP}_{L}$&$\mbox{GHP}_{L}$\\
&&(-5.7, 1.5)&\cellcolor{gray!25}(-4.1, 5.4)&(-4.3, 4.4)&\cellcolor{gray!25}(-6.2, 4.1)&(-8.7, 4.0)&\cellcolor{gray!25}(-10.4, 3.7)&(-12.5, 4.2)\\ \cline{2-9}
&\multirow{2}{*}{1500}&\cellcolor{gray!25}$\mbox{GHP}_{M}$&$\mbox{GHP}_{L}$&\cellcolor{gray!25}$\mbox{GHP}_{L}$&$\mbox{GHP}_{L}$&\cellcolor{gray!25}$\mbox{GHP}_{L}$&$\mbox{GHP}_{L}$&\cellcolor{gray!25}$\mbox{GHP}_{L}$\\
&&\cellcolor{gray!25}(-5.0, 1.3)&(-2.8, 5.1)&\cellcolor{gray!25}(-3.6, 4.5)&(-5.0, 3.8)&\cellcolor{gray!25}(-6.8, 4.0)&(-8.5, 3.1)&\cellcolor{gray!25}(-10.5, 3.4)\\ \cline{2-9}
&\multirow{2}{*}{2000}&$\mbox{GHP}_{M}$&\cellcolor{gray!25}$\mbox{GHP}_{L}$&$\mbox{GHP}_{L}$&\cellcolor{gray!25}$\mbox{GHP}_{L}$&$\mbox{GHP}_{L}$&\cellcolor{gray!25}$\mbox{GHP}_{L}$&$\mbox{GHP}_{L}$\\
&&(-4.7, 1.1)&\cellcolor{gray!25}(-2.8, 4.8)&(-2.6, 4.0)&\cellcolor{gray!25}(-4.2, 3.8)&(-6.0, 3.1)&\cellcolor{gray!25}(-7.5, 2.9)&(-9.1, 2.8)\\ \cline{2-9}
&\multirow{2}{*}{2500}&\cellcolor{gray!25}$\mbox{GHP}_{M}$&$\mbox{GHP}_{L}$&\cellcolor{gray!25}$\mbox{GHP}_{L}$&$\mbox{GHP}_{L}$&\cellcolor{gray!25}$\mbox{GHP}_{L}$&$\mbox{GHP}_{L}$&\cellcolor{gray!25}$\mbox{GHP}_{L}$\\
&&\cellcolor{gray!25}(-4.4, 0.7)&(-2.1, 4.4)&\cellcolor{gray!25}(-3.1, 4.0)&(-3.6, 3.3)&\cellcolor{gray!25}(-5.1, 2.9)&(-6.5, 2.5)&\cellcolor{gray!25}(-8.4, 2.6)\\ \cline{2-9}
\end{tabular*} \caption{Type: SvS. Best GHP Estimator Bounds.}
\ifx\templateType\templateArxiv
    \vspace{-300mm}
\else 
    \vspace{-16mm}
\fi
\end{table}

\FloatBarrier
\clearpage
\ifx\templateType\templateArxiv\renewcommand{\arraystretch}{1}\fi
\ifx\templateType\templateIEEE\renewcommand{\arraystretch}{1.1}\fi
\begin{table}[ht]
\centering \small

\begin{tabular*}{\linewidth}{@{\extracolsep{\fill}}cccccccc}\\
&&\multicolumn{6}{c}{Number of Features}\\\cline{3-8}
&&2&3&4&10&20&30\\ \cline{2-8}
\parbox[t]{3mm}{\multirow{10}{*}{\rotatebox[origin=c]{90}{Number of Samples Per Class}}}&\multirow{2}{*}{500}&\cellcolor{gray!25}$\mbox{kNN}_{H}$&$\mbox{kNN}_{H}$&\cellcolor{gray!25}$\mbox{kNN}_{H}$&$\mbox{kNN}_{M}$&\cellcolor{gray!25}$\mbox{kNN}_{M}$&$\mbox{kNN}_{L}$\\
&&\cellcolor{gray!25}(-2.1, 3.2)&(-2.2, 3.0)&\cellcolor{gray!25}(-2.6, 2.7)&(-3.1, 5.0)&\cellcolor{gray!25}(-8.3, 5.2)&(-8.3, 10.8)\\ \cline{2-8}
&\multirow{2}{*}{1000}&$\bm{\mbox{kNN}_{H}}$&\cellcolor{gray!25}$\bm{\mbox{kNN}_{H}}$&$\bm{\mbox{kNN}_{H}}$&\cellcolor{gray!25}$\mbox{kNN}_{M}$&$\mbox{kNN}_{M}$&\cellcolor{gray!25}$\mbox{kNN}_{L}$\\
&&\textbf{(-1.6, 2.2)}&\cellcolor{gray!25}\textbf{(-1.8, 1.9)}&\textbf{(-2.0, 1.9)}&\cellcolor{gray!25}(-2.8, 3.7)&(-7.7, 3.9)&\cellcolor{gray!25}(-7.9, 9.0)\\ \cline{2-8}
&\multirow{2}{*}{1500}&\cellcolor{gray!25}$\bm{\mbox{kNN}_{H}}$&$\bm{\mbox{kNN}_{H}}$&\cellcolor{gray!25}$\bm{\mbox{kNN}_{H}}$&$\mbox{kNN}_{M}$&\cellcolor{gray!25}$\mbox{kNN}_{L}$&$\mbox{kNN}_{L}$\\
&&\cellcolor{gray!25}\textbf{(-1.6, 1.6)}&\textbf{(-1.5, 1.6)}&\cellcolor{gray!25}\textbf{(-1.8, 1.5)}&(-2.6, 3.2)&\cellcolor{gray!25}(-4.0, 6.8)&(-7.4, 7.8)\\ \cline{2-8}
&\multirow{2}{*}{2000}&$\bm{\mbox{kNN}_{H}}$&\cellcolor{gray!25}$\bm{\mbox{kNN}_{H}}$&$\bm{\mbox{kNN}_{H}}$&\cellcolor{gray!25}$\mbox{kNN}_{M}$&$\mbox{kNN}_{L}$&\cellcolor{gray!25}$\mbox{kNN}_{L}$\\
&&\textbf{(-1.4, 1.4)}&\cellcolor{gray!25}\textbf{(-1.6, 1.2)}&\textbf{(-1.6, 1.2)}&\cellcolor{gray!25}(-2.3, 2.8)&(-3.4, 5.9)&\cellcolor{gray!25}(-7.1, 7.2)\\ \cline{2-8}
&\multirow{2}{*}{2500}&\cellcolor{gray!25}$\bm{\mbox{kNN}_{H}}$&$\bm{\mbox{kNN}_{H}}$&\cellcolor{gray!25}$\bm{\mbox{kNN}_{H}}$&$\bm{\mbox{kNN}_{M}}$&\cellcolor{gray!25}$\mbox{kNN}_{M}$&$\mbox{kNN}_{L}$\\
&&\cellcolor{gray!25}\textbf{(-1.3, 1.2)}&\textbf{(-1.4, 1.1)}&\cellcolor{gray!25}\textbf{(-1.6, 1.1)}&\textbf{(-2.3, 2.6)}&\cellcolor{gray!25}(-6.1, 2.6)&(-7.2, 7.3)\\ \cline{2-8}
\end{tabular*} 
\caption{Type: GvG. Best kNN Estimator Bounds.}
\label{table:gvg-knn}
\vspace{-2mm}

\begin{tabular*}{\linewidth}{@{\extracolsep{\fill}}ccccccccc}\\
&&\multicolumn{7}{c}{Number of Features}\\\cline{3-9}
&&2&4&6&8&10&12&20\\ \cline{2-9}
\parbox[t]{3mm}{\multirow{10}{*}{\rotatebox[origin=c]{90}{Number of Samples Per Class}}}&\multirow{2}{*}{500}&\cellcolor{gray!25}$\mbox{kNN}_{H}$&$\mbox{kNN}_{H}$&\cellcolor{gray!25}$\mbox{kNN}_{M}$&$\mbox{kNN}_{M}$&\cellcolor{gray!25}$\mbox{kNN}_{M}$&$\mbox{kNN}_{M}$&\cellcolor{gray!25}$\mbox{kNN}_{L}$\\
&&\cellcolor{gray!25}(-2.5, 2.8)&(-3.6, 2.3)&\cellcolor{gray!25}(-2.6, 5.4)&(-3.2, 4.9)&\cellcolor{gray!25}(-4.9, 5.6)&(-6.7, 5.4)&\cellcolor{gray!25}(-6.0, 10.5)\\ \cline{2-9}
&\multirow{2}{*}{1000}&$\bm{\mbox{kNN}_{H}}$&\cellcolor{gray!25}$\bm{\mbox{kNN}_{H}}$&$\mbox{kNN}_{M}$&\cellcolor{gray!25}$\mbox{kNN}_{M}$&$\mbox{kNN}_{M}$&\cellcolor{gray!25}$\mbox{kNN}_{M}$&$\mbox{kNN}_{L}$\\
&&\textbf{(-1.8, 1.9)}&\cellcolor{gray!25}\textbf{(-2.3, 1.5)}&(-1.6, 3.8)&\cellcolor{gray!25}(-2.3, 4.0)&(-3.9, 4.1)&\cellcolor{gray!25}(-6.0, 3.9)&(-5.0, 9.6)\\ \cline{2-9}
&\multirow{2}{*}{1500}&\cellcolor{gray!25}$\bm{\mbox{kNN}_{H}}$&$\bm{\mbox{kNN}_{H}}$&\cellcolor{gray!25}$\bm{\mbox{kNN}_{H}}$&$\bm{\mbox{kNN}_{M}}$&\cellcolor{gray!25}$\mbox{kNN}_{M}$&$\mbox{kNN}_{M}$&\cellcolor{gray!25}$\mbox{kNN}_{L}$\\
&&\cellcolor{gray!25}\textbf{(-1.5, 1.6)}&\textbf{(-2.1, 1.2)}&\cellcolor{gray!25}\textbf{(-3.1, 0.8)}&\textbf{(-1.9, 2.9)}&\cellcolor{gray!25}(-3.3, 3.2)&(-5.8, 3.3)&\cellcolor{gray!25}(-4.4, 7.9)\\ \cline{2-9}
&\multirow{2}{*}{2000}&$\bm{\mbox{kNN}_{H}}$&\cellcolor{gray!25}$\bm{\mbox{kNN}_{H}}$&$\bm{\mbox{kNN}_{M}}$&\cellcolor{gray!25}$\bm{\mbox{kNN}_{M}}$&$\mbox{kNN}_{M}$&\cellcolor{gray!25}$\mbox{kNN}_{M}$&$\mbox{kNN}_{L}$\\
&&\textbf{(-1.4, 1.3)}&\cellcolor{gray!25}\textbf{(-1.9, 1.1)}&\textbf{(-1.0, 2.8)}&\cellcolor{gray!25}\textbf{(-1.6, 2.5)}&(-2.9, 3.0)&\cellcolor{gray!25}(-5.3, 2.9)&(-4.0, 8.1)\\ \cline{2-9}
&\multirow{2}{*}{2500}&\cellcolor{gray!25}$\bm{\mbox{kNN}_{H}}$&$\bm{\mbox{kNN}_{H}}$&\cellcolor{gray!25}$\bm{\mbox{kNN}_{H}}$&$\bm{\mbox{kNN}_{M}}$&\cellcolor{gray!25}$\mbox{kNN}_{M}$&$\mbox{kNN}_{M}$&\cellcolor{gray!25}$\mbox{kNN}_{L}$\\
&&\cellcolor{gray!25}\textbf{(-1.3, 1.2)}&\textbf{(-1.8, 0.9)}&\cellcolor{gray!25}\textbf{(-2.5, 0.7)}&\textbf{(-1.4, 2.4)}&\cellcolor{gray!25}(-2.7, 2.6)&(-4.7, 2.5)&\cellcolor{gray!25}(-3.9, 7.1)\\ \cline{2-9}
\end{tabular*} 
\caption{Type: TvT. Best kNN Estimator Bounds.}
\label{table:tvt-knn}
\vspace{-2mm}

\begin{tabular*}{\linewidth}{@{\extracolsep{\fill}}ccccccccc}\\
&&\multicolumn{7}{c}{Number of Features}\\\cline{3-9}
&&2&4&6&8&10&12&14\\ \cline{2-9}
\parbox[t]{3mm}{\multirow{10}{*}{\rotatebox[origin=c]{90}{Number of Samples Per Class}}}&\multirow{2}{*}{500}&\cellcolor{gray!25}$\mbox{kNN}_{H}$&$\mbox{kNN}_{H}$&\cellcolor{gray!25}$\mbox{kNN}_{H}$&$\mbox{kNN}_{M}$&\cellcolor{gray!25}$\mbox{kNN}_{M}$&$\mbox{kNN}_{M}$&\cellcolor{gray!25}$\mbox{kNN}_{M}$\\
&&\cellcolor{gray!25}(-2.3, 3.1)&(-3.3, 2.4)&\cellcolor{gray!25}(-4.3, 2.1)&(-5.2, 5.6)&\cellcolor{gray!25}(-7.4, 5.2)&(-8.9, 5.4)&\cellcolor{gray!25}(-9.7, 4.9)\\ \cline{2-9}
&\multirow{2}{*}{1000}&$\bm{\mbox{kNN}_{H}}$&\cellcolor{gray!25}$\bm{\mbox{kNN}_{H}}$&$\bm{\mbox{kNN}_{H}}$&\cellcolor{gray!25}$\mbox{kNN}_{M}$&$\mbox{kNN}_{M}$&\cellcolor{gray!25}$\mbox{kNN}_{M}$&$\mbox{kNN}_{M}$\\
&&\textbf{(-1.8, 2.1)}&\cellcolor{gray!25}\textbf{(-2.4, 1.6)}&\textbf{(-3.0, 1.4)}&\cellcolor{gray!25}(-3.8, 4.6)&(-5.6, 4.0)&\cellcolor{gray!25}(-7.3, 4.3)&(-8.5, 4.2)\\ \cline{2-9}
&\multirow{2}{*}{1500}&\cellcolor{gray!25}$\bm{\mbox{kNN}_{H}}$&$\bm{\mbox{kNN}_{H}}$&\cellcolor{gray!25}$\bm{\mbox{kNN}_{H}}$&$\mbox{kNN}_{M}$&\cellcolor{gray!25}$\mbox{kNN}_{M}$&$\mbox{kNN}_{M}$&\cellcolor{gray!25}$\mbox{kNN}_{M}$\\
&&\cellcolor{gray!25}\textbf{(-1.5, 1.7)}&\textbf{(-2.0, 1.3)}&\cellcolor{gray!25}\textbf{(-2.7, 1.0)}&(-3.1, 3.5)&\cellcolor{gray!25}(-4.7, 3.6)&(-6.4, 3.6)&\cellcolor{gray!25}(-7.6, 4.0)\\ \cline{2-9}
&\multirow{2}{*}{2000}&$\bm{\mbox{kNN}_{H}}$&\cellcolor{gray!25}$\bm{\mbox{kNN}_{H}}$&$\bm{\mbox{kNN}_{H}}$&\cellcolor{gray!25}$\mbox{kNN}_{M}$&$\mbox{kNN}_{M}$&\cellcolor{gray!25}$\mbox{kNN}_{M}$&$\mbox{kNN}_{L}$\\
&&\textbf{(-1.4, 1.4)}&\cellcolor{gray!25}\textbf{(-1.9, 1.1)}&\textbf{(-2.4, 0.9)}&\cellcolor{gray!25}(-2.9, 3.1)&(-4.2, 2.7)&\cellcolor{gray!25}(-5.9, 3.3)&(-5.5, 7.9)\\ \cline{2-9}
&\multirow{2}{*}{2500}&\cellcolor{gray!25}$\bm{\mbox{kNN}_{H}}$&$\bm{\mbox{kNN}_{H}}$&\cellcolor{gray!25}$\bm{\mbox{kNN}_{H}}$&$\mbox{kNN}_{M}$&\cellcolor{gray!25}$\mbox{kNN}_{M}$&$\mbox{kNN}_{M}$&\cellcolor{gray!25}$\mbox{kNN}_{M}$\\
&&\cellcolor{gray!25}\textbf{(-1.3, 1.2)}&\textbf{(-1.7, 1.0)}&\cellcolor{gray!25}\textbf{(-2.2, 0.8)}&(-2.5, 2.7)&\cellcolor{gray!25}(-3.8, 2.5)&(-5.3, 2.8)&\cellcolor{gray!25}(-6.6, 2.9)\\ \cline{2-9}
\end{tabular*} 
\caption{Type: TvS. Best kNN Estimator Bounds.}
\label{table:tvs-knn}
\vspace{-2mm}

\begin{tabular*}{\linewidth}{@{\extracolsep{\fill}}ccccccccc}\\
&&\multicolumn{7}{c}{Number of Features}\\\cline{3-9}
&&2&3&4&6&8&10&12\\ \cline{2-9}
\parbox[t]{3mm}{\multirow{10}{*}{\rotatebox[origin=c]{90}{Number of Samples Per Class}}}&\multirow{2}{*}{500}&\cellcolor{gray!25}$\mbox{kNN}_{H}$&$\mbox{kNN}_{M}$&\cellcolor{gray!25}$\mbox{kNN}_{M}$&$\mbox{kNN}_{L}$&\cellcolor{gray!25}$\mbox{kNN}_{L}$&$\mbox{kNN}_{L}$&\cellcolor{gray!25}$\mbox{kNN}_{L}$\\
&&\cellcolor{gray!25}(-3.7, 2.3)&(-3.6, 5.9)&\cellcolor{gray!25}(-6.0, 5.6)&(-6.4, 11.7)&\cellcolor{gray!25}(-9.1, 10.8)&(-10.9, 10.2)&\cellcolor{gray!25}(-12.8, 10.8)\\ \cline{2-9}
&\multirow{2}{*}{1000}&$\bm{\mbox{kNN}_{H}}$&\cellcolor{gray!25}$\mbox{kNN}_{M}$&$\mbox{kNN}_{M}$&\cellcolor{gray!25}$\mbox{kNN}_{L}$&$\mbox{kNN}_{L}$&\cellcolor{gray!25}$\mbox{kNN}_{L}$&$\mbox{kNN}_{L}$\\
&&\textbf{(-2.4, 1.6)}&\cellcolor{gray!25}(-1.9, 4.5)&(-4.0, 4.4)&\cellcolor{gray!25}(-4.4, 9.7)&(-6.5, 8.9)&\cellcolor{gray!25}(-8.2, 8.5)&(-9.6, 9.2)\\ \cline{2-9}
&\multirow{2}{*}{1500}&\cellcolor{gray!25}$\bm{\mbox{kNN}_{H}}$&$\mbox{kNN}_{M}$&\cellcolor{gray!25}$\mbox{kNN}_{M}$&$\mbox{kNN}_{L}$&\cellcolor{gray!25}$\mbox{kNN}_{L}$&$\mbox{kNN}_{L}$&\cellcolor{gray!25}$\mbox{kNN}_{L}$\\
&&\cellcolor{gray!25}\textbf{(-1.9, 1.3)}&(-1.3, 3.8)&\cellcolor{gray!25}(-2.8, 3.9)&(-3.3, 8.7)&\cellcolor{gray!25}(-5.4, 9.5)&(-7.1, 8.0)&\cellcolor{gray!25}(-8.8, 7.8)\\ \cline{2-9}
&\multirow{2}{*}{2000}&$\bm{\mbox{kNN}_{H}}$&\cellcolor{gray!25}$\bm{\mbox{kNN}_{M}}$&$\mbox{kNN}_{M}$&\cellcolor{gray!25}$\mbox{kNN}_{L}$&$\mbox{kNN}_{L}$&\cellcolor{gray!25}$\mbox{kNN}_{L}$&$\mbox{kNN}_{L}$\\
&&\textbf{(-1.7, 1.2)}&\cellcolor{gray!25}\textbf{(-0.9, 3.2)}&(-2.3, 3.6)&\cellcolor{gray!25}(-2.4, 8.2)&(-4.8, 8.2)&\cellcolor{gray!25}(-6.3, 7.0)&(-7.8, 7.7)\\ \cline{2-9}
&\multirow{2}{*}{2500}&\cellcolor{gray!25}$\bm{\mbox{kNN}_{H}}$&$\bm{\mbox{kNN}_{M}}$&\cellcolor{gray!25}$\bm{\mbox{kNN}_{M}}$&$\mbox{kNN}_{L}$&\cellcolor{gray!25}$\mbox{kNN}_{L}$&$\mbox{kNN}_{L}$&\cellcolor{gray!25}$\mbox{kNN}_{L}$\\
&&\cellcolor{gray!25}\textbf{(-1.6, 1.0)}&\textbf{(-0.7, 3.0)}&\cellcolor{gray!25}\textbf{(-1.8, 3.0)}&(-2.0, 7.1)&\cellcolor{gray!25}(-4.2, 7.0)&(-5.8, 6.3)&\cellcolor{gray!25}(-7.4, 7.3)\\ \cline{2-9}
\end{tabular*} 
\caption{Type: SvS. Best kNN Estimator Bounds.}
\label{table:svs-knn}

\end{table}

\end{document}